\documentclass{article}

\usepackage{microtype}
\usepackage{graphicx}
\usepackage{subcaption}
\usepackage{booktabs} 

\usepackage{hyperref}

\usepackage[accepted]{icml2024}

\usepackage{amsmath}
\usepackage{amssymb}
\usepackage{mathtools}
\usepackage{amsthm}

\usepackage[capitalize,noabbrev]{cleveref}

\theoremstyle{plain}

\theoremstyle{definition}

\theoremstyle{remark}

\usepackage[textsize=tiny]{todonotes}
\usepackage[utf8]{inputenc} 
\usepackage[T1]{fontenc}    

\usepackage{url}            
\usepackage{booktabs}       
\usepackage{amsfonts}       
\usepackage{bbm} 
\usepackage{amsmath}       
\usepackage{nicefrac}       
\usepackage{microtype}      
\usepackage{xcolor}        
\usepackage{verbatim}        

\usepackage{algorithm}
\usepackage{algorithmic}

\usepackage{tikz}

\definecolor{olivegreen}{rgb}{0, 0.6, 0}
\definecolor{black}{HTML}{000000}
\definecolor{white}{HTML}{ffffff}
\definecolor{color1}{HTML}{ACE5EE}
\definecolor{color2}{HTML}{0093AF}
\definecolor{color3}{HTML}{CC0000}
\definecolor{color4}{HTML}{0087BD}
\definecolor{color5}{HTML}{333399}
\definecolor{color6}{HTML}{20B2AA}
\definecolor{color7}{HTML}{87CEEB}

\usepackage{xspace}         
\usepackage{graphicx}

\usepackage{lipsum}
\usepackage{pgfplots}
\usepackage{pgfplotstable}
\usepgfplotslibrary{groupplots}

\usepackage{subcaption}
\usepackage{enumitem}
\usepackage{array, makecell} %
\newcolumntype{x}[1]{>{\centering\arraybackslash\hspace{0pt}}p{#1}}

\usepackage{multirow}
\usepackage{wrapfig}

\usepackage{pifont}
\newcommand{\cmark}{\color{olivegreen}\ding{51}}%
\newcommand{\xmark}{\color{red}\ding{55}}%

\newcommand{\name}{DataFreeShield\xspace}%
\newcommand{\nameabb}{Ours\xspace}%
\newcommand{\mix}{diversified sample synthesis\xspace}%
\newcommand{\Mix}{Diversified sample synthesis\xspace}%
\newcommand{\MIX}{Diversified Sample Synthesis\xspace}%
\newcommand{\mixabb}{DSS\xspace}
\newcommand{\lossdfs}{\mathcal{L}_{DFShield}}
\newcommand{\gradmask}{GradRefine\xspace}
\newcommand{\medmnist}{medical datasets\xspace}

\newcommand{\loss}{\mathcal{L}}

\newcommand{\eps}{\epsilon} 

\newcommand\norm[1]{\lVert#1\rVert}

\newcommand{\JL}[1]{{\color{olivegreen}[\textbf{\sc JLee}: \textit{#1}]}}
\newcommand{\KH}[1]{{\color{purple}[\textbf{\sc KH}: \textit{#1}]}}
\newcommand{\HY}[1]{{\color{blue}[\textbf{\sc HY}: \textit{#1}]}}

 \def\final{}   
 \ifdefined\final
  \renewcommand{\JL}[1]{}
  \renewcommand{\KH}[1]{}
  \renewcommand{\HY}[1]{}
\fi

\def\blfootnote{\xdef\@thefnmark{}\@footnotetext}

\date{}

\usepackage{etoolbox}

\icmltitlerunning{DataFreeShield: Defending Adversarial Attacks without Training Data}

\begin{document}

\twocolumn[

\icmltitle{DataFreeShield: Defending Adversarial Attacks without Training Data}



\icmlsetsymbol{equal}{*}

\begin{icmlauthorlist}
\icmlauthor{Hyeyoon Lee}{snu}
\icmlauthor{Kanghyun Choi}{snu}
\icmlauthor{Dain Kwon}{snu}
\icmlauthor{Sunjong Park}{snu}
\icmlauthor{Mayoore Selvarasa Jaiswal}{nvidia}
\icmlauthor{Noseong Park}{kaist}
\icmlauthor{Jonghyun Choi}{snu}
\icmlauthor{Jinho Lee}{snu}
\end{icmlauthorlist}

\icmlaffiliation{snu}{Department of Electrical and Computer Engineering, Seoul National University, Seoul, South Korea}
\icmlaffiliation{kaist}{School of Computing, KAIST, Daejeon, South Korea}
\icmlaffiliation{nvidia}{NVIDIA, Work done while at IBM}

\icmlcorrespondingauthor{Jinho Lee}{leejinho@snu.ac.kr}

\icmlkeywords{Adversarial Robustness, Adversarial Defense, Deep Learning, Data-Free, Adversarial Attack}

\vskip 0.3in
]



\printAffiliationsAndNotice{}  

\begin{abstract}
Recent advances in adversarial robustness rely on an abundant set of training data, where using external or additional datasets has become a common setting.
However, in real life, the training data is often kept private for security and privacy issues, while only the pretrained weight is available to the public.
In such scenarios, existing methods that assume accessibility to the original data become inapplicable.
Thus we investigate the pivotal problem of \emph{data-free adversarial robustness}, where we try to achieve adversarial robustness without accessing any real data.
Through a preliminary study, we highlight the severity of the problem by showing that robustness without the original dataset is difficult to achieve, even with similar domain datasets.
To address this issue, we propose \name, which tackles the problem from two perspectives: surrogate dataset generation and adversarial training using the generated data.
Through extensive validation, we show that \name outperforms baselines, demonstrating that the proposed method sets the first entirely data-free solution for the adversarial robustness problem.

\end{abstract}

\section{Introduction}
\label{sec:intro}
Since the discovery of the adversarial examples~\citep{goodfellow2014explaining, szegedy2013intriguing} and their ability to successfully fool well-trained classifiers, training a robust classifier has become an important topic of research~\citep{schmidt2018adversarially, obfuscated}.  
If not properly circumvented, adversarial attacks can be a great threat to real-life applications such as self-driving automobiles and face recognition when intentionally abused.

Among many efforts made over the past few years, adversarial training (AT)~\citep{madry} has become the de facto standard approach to training a robust model. 
AT uses adversarially perturbed examples as part of training data so that models can learn to correctly classify them.
Due to its success, many variants of AT have been proposed to further improve its effectiveness~\citep{trades, mart, iad}. 
For AT and its variants, it is commonly assumed that the original data is available for training.
Going a step further, many approaches import external data from the same or similar domains to add diversity to the training samples (e.g., adding Tiny-ImageNet to CIFAR-10), such that the trained model can have better generalization ability~\citep{rebuffi2021data, unlabelddata}. 

Unfortunately, the original training dataset is often not available in many real-world scenarios~\citep{liu2021machine, hathaliya2020exhaustive}.
While there are some public datasets available for certain domains (e.g., image classification),
many real-world data are publicly unavailable due to privacy, security, or proprietary issues, with only the pretrained models available~\citep{clip, imagen, dalle}.
The problem becomes even more severe when it comes to specific domains that are privacy-sensitive (e.g., biometric, medical, etc.) where alternative datasets for training are difficult to find in the public.
Therefore, if a user wants a pretrained model to become robust against adversarial attacks, there is currently no clear method to do so without the original training data. 

In such circumstances, we study the problem of adversarial robustness under a more realistic and practical setting of \emph{data-free adversarial robustness},  where a \textit{non-robustly} pretrained model is given and its robust version should be learned without access to the original training data.
To address the problem, we propose \name, a novel method thoroughly designed to achieve robustness without any real data.
Specifically, we propose a synthetic sample diversification method with dynamic synthetic loss modulation to maximize the diversity of the synthetic dataset.
 Moreover,  we propose a gradient refinement method \gradmask to obtain a smoother loss surface, to minimize the impact of the distribution gap between synthetic and real data.
 Along with a soft guided training loss designed to maximize the transferability of robustness, \name achieves significantly better robustness over prior art. 
To the best of our knowledge, this is the first work to faithfully address the definition of data-free adversarial robustness, and suggest an effective
solution without relying on any real data. 

Overall, our contributions are summarized as follows:
\begin{itemize}[topsep=0.em, itemsep=0.em, leftmargin=1.6em]
    \item For the first time, we properly address the robustness problem in an entirely data-free manner, which gives adversarial robustness to non-robustly pretrained models without the original datasets. 
    \item To tackle the challenge of limited diversity in synthetic datasets, we devise \mix, a novel technique for generating synthetic samples.
    \item We propose a gradient refinement method with a soft-guidance based training loss
    to minimize the impact of distribution shift incurred from synthetic data training. 
    \item We propose \name, a completely data-free approach that can effectively convert a pretrained model to an adversarially robust one and show that \name achieves significantly better robustness on various datasets over the baselines. 
\end{itemize}

\section{Background}

\subsection{Adversarial Robustness}
Among many defense techniques for making DNN models robust against adversarial attacks, adversarial training~\citep{madry} (AT) has been the most successful method, formulated as:
\begin{equation}
\begin{split}
    \min_{\theta}\frac{1}{n}\sum^{n}_{i}\max_{{x}'_i \in \mathcal{X} }\loss(f_{\theta}(x_i'),y_i),\\
    ~~\text{where}~~\mathcal{X} = \{{x}'_i \vert \ \norm{{x}'_i-{x}_i}_p \leq\epsilon \},
\label{eq:adv}
\end{split}
\end{equation}
where $\loss$ is the loss function for classification (e.g., cross-entropy), $n$ is the number of training samples, and $\epsilon$ is the maximum perturbation limit.
$x'$ is an arbitrary adversarial sample that is generated based on $x$ to deceive the original decision, where $p=\infty$ is a popular choice. 
In practice, finding the optimal solution for the inner maximization is intractable, such that known adversarial attack methods are often used.
For example, PGD~\citep{madry} is a widely-used method, such that
\begin{equation}
    x^t = \Pi_{\epsilon}(x^{t-1} + \alpha \cdot sign \left( \nabla_{x} \loss(f_{\theta}(x^{t-1}),y) ) \right), 
\label{eq:pgd}
\end{equation}
where $t$ is the number of iteration steps.
For each step, the image is updated to maximize the target loss, then projected onto the epsilon ball, denoted by $\Pi_{\epsilon}$.

\subsection{Similar Approaches to Data-Free Robustness} 
One related approach to data-free robustness is test-time defense techniques~\citep{dad,diffpure,TTE} that do not train the target model but use separate modules for attack detection or purification.
However, we find that those methods often rely on the availability of other data, or have limited applicability when used on their own.
For instance, DAD~\citep{dad} uses the test set data for calibration, which could be regarded as a data leak.
DiffPure~\citep{diffpure} relies on diffusion models that are pretrained on a superset of the target dataset domain, restricting its use against unseen domains.
TTE~\citep{TTE} provides defense from augmentations agnostic to datasets. 
However, it is mainly used to enhance defense on models already adversarially trained, and its effect on non-robust models is marginal.
Thus existing methods do not truthfully address the problem of data-free adversarial robustness, leaving vulnerability in data-absent situations.

\subsection{Dataset Generation for Data-free Learning} 
When a model needs to be trained without the training data (i.e., data-free learning), a dominant approach is to generate a surrogate dataset, commonly adopted in data-free adaptations of knowledge distillation~\citep{dfkd, dfad}, quantization~\citep{gdfq, qimera}, model extraction~\citep{dfme}, or domain adaptation~\citep{kurmi2021domain}.
Given only a pretrained model, the common choice for synthesis loss in the literature~\citep{imagine, deepinversion} are as follows:
\begin{align}
\mathit{\loss_{class}}&= - \sum^C_c \hat y_c log(f_\theta(\hat x)_c), \\
\mathit{\loss_{feature}}&= \sum_{l=1}^{L}\|\mu_{l}^{T}-\mu_{l} \|_{2}^{2} + \|\sigma_{l}^{T} - \sigma_{l}\|_{2}^{2}, \\
\mathit{\loss_{prior}}&= \sum_{i,j}\|\hat{x}_{i,j+1}-\hat{x}_{i,j}\|_{2}^{2} + \|\hat{x}_{i+1,j}-\hat{x}_{i,j}\|_{2}^{2}, 
\label{eq:synth_each}
\end{align}
where $\loss_{class}$ is the cross-entropy loss with artificial label $\hat y$ among $C$ classes with the output of the model $f_\theta$, $\loss_{feature}$ regularizes the samples' layer-wise distributions ($\mu$, $\sigma$) to follow the saved statistics in the batch normalization ($\mu^T$,$\sigma^T$) over $L$ layers, and $\loss_{prior}$ penalizes the total variance of the samples in pixel level. 
These losses are jointly used to train a generator~\citep{zaq, gdfq, qimera} or to directly optimize samples from noise~\citep{imagine, deepinversion, plugin}, with fixed hyperparameters $\alpha_i$:
\begin{equation}
    \mathit{\loss_{Synth}}= \alpha_{1}\mathit{\loss_{class}}+\alpha_{2}\mathit{\loss_{feature}}+\alpha_{3}\mathit{\loss_{prior}}. 
\label{eq:synth_all}
\end{equation}

\section{Data-free Adversarial Robustness Problem}
\subsection{Problem Definition}
\label{sec:problem}
In the problem of learning data-free adversarial robustness, the objective is to obtain a robust model $S(\cdot)$ from a pretrained original model $T(\cdot)$, without access to its original training data $(x,y)$. 
Hereafter, we will denote $T(\cdot)$ and $S(\cdot)$ as teacher and student, respectively. 

In the problem, the typical AT formulation of \cref{eq:adv} cannot be directly applied because none of $x$ or $y$ is available for training or fine-tuning.
Instead, following the common choice of data-free learning~\cite{gdfq,ait, dfkd}, we choose to use a surrogate training dataset $(\hat{x},\hat{y})$ to train $S(\cdot)$, 
which allows us to use the de facto standard method for adversarial robustness: adversarial training.
With the given notations, we can reformulate the objective in \cref{eq:adv} as:
\begin{equation}
\begin{split}
\min_{\theta}\frac{1}{n}\sum^{n}_{i}\max_{\hat{x}'_i \in \mathcal{\hat{X}}} \loss(S_{\theta}(\hat{x}_i'),\hat{y}_i), \\
~~\text{where}~~ \mathcal{\hat{X}} = \{\hat{x}'_i \vert \ \norm{\hat{x}'_i-\hat{x}_i}_p \leq\epsilon \}.
\label{eq:dfar_obj}
\end{split}
\end{equation}
However, it remains to be answered how to create good surrogate training samples $(\hat{x},\hat{y})$, and what loss function $\loss$ can best generalize the learned robustness to defend against attacks on real data.

\subsection{Motivational Study}
Here, we demonstrate the difficulty of the problem by answering one naturally arising question:  
\textbf{Can we just use another real dataset?} 
A relevant prior art is test-time defense such as DiffPure~\citep{diffpure}, or DAD~\citep{dad} which uses auxiliary models trained on large datasets (e.g., ImageNet) to improve defense on CIFAR-10.
However, they strongly rely on the datasets from the same domain. 
In practice, there is no guarantee on such similarity, 
especially on tasks with specific domains (e.g., biomedical). 

\begin{figure}[t]
    \centering
    \begin{subfigure}{.77\columnwidth}
    \includegraphics[width=\textwidth]{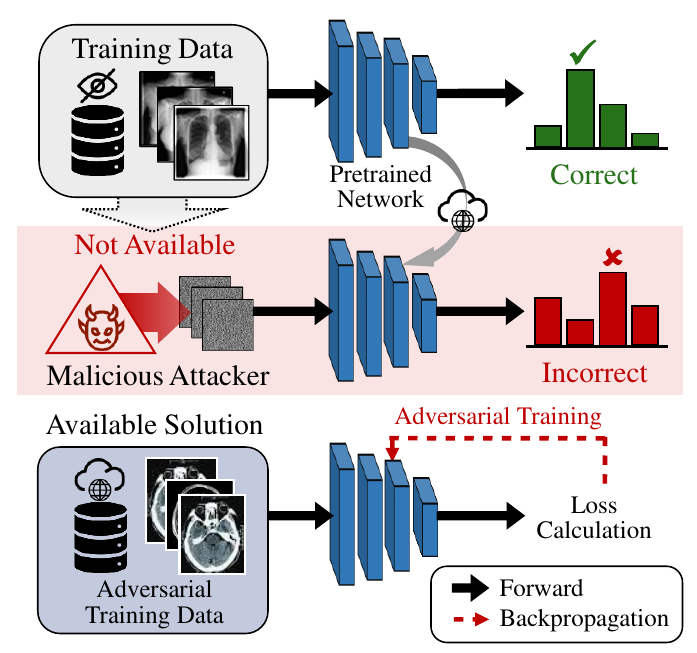}
    \caption{Motivational study scenario.}
    \label{fig:moti_medmnist1}
    \end{subfigure} 
    \vspace{0mm}
    \begin{subfigure}{.75\columnwidth}
    \includegraphics[width=\textwidth,trim={5mm 0 7mm 0},clip]{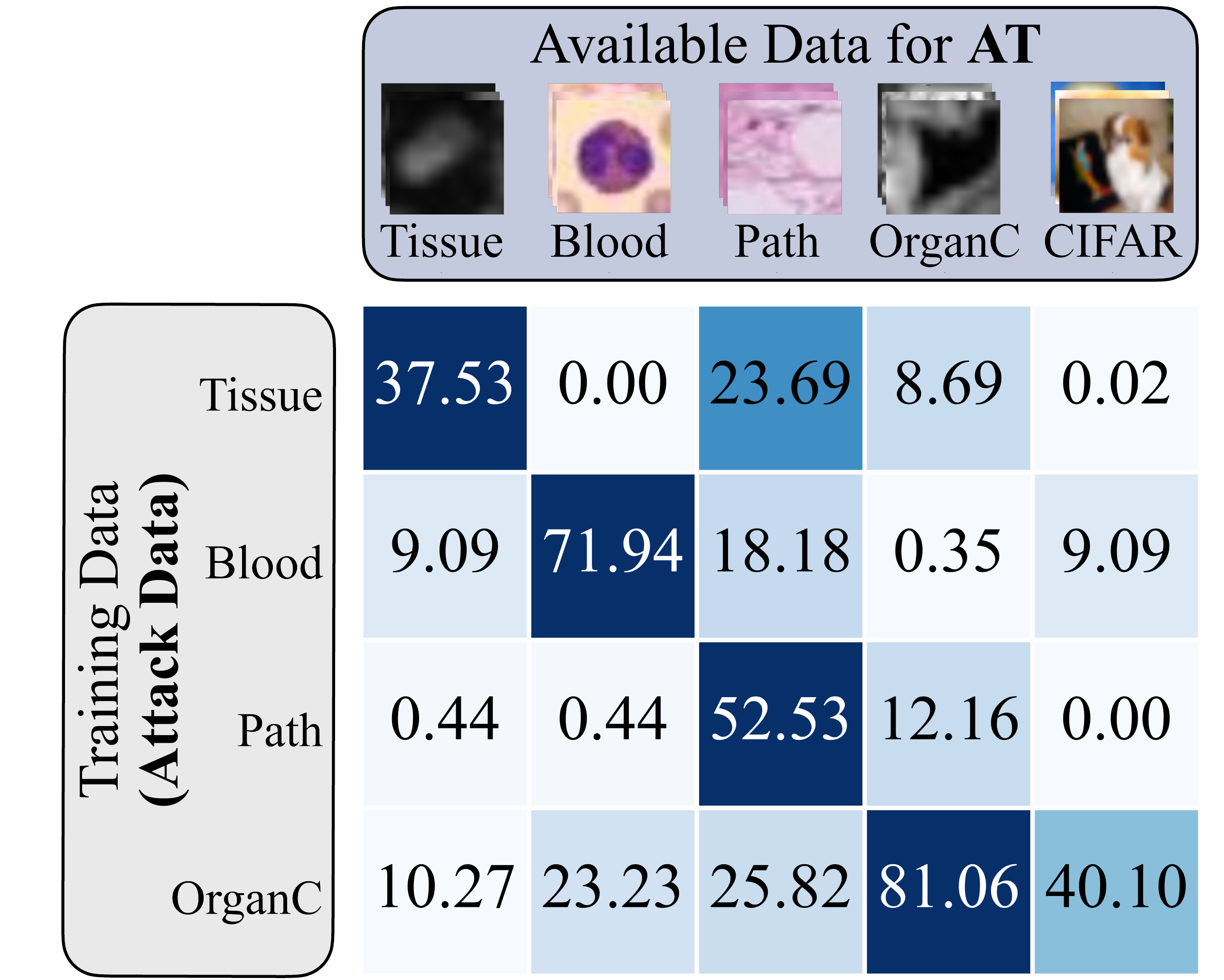} 
    \caption{Robust accuracy trained on similar and general domain datasets.}
    \label{fig:moti_medmnist2}
    \end{subfigure}
    \vspace{-2mm}
\caption{Motivational experiment using biomedical datasets \citep{medmnistv2}. (a) demonstrates the problem scenario where adversarial threat prevails for models pretrained with private datasets. 
(b) plots the results when adversarial training is done with a similar or public dataset.
}
\label{fig:moti_medmnist}
\end{figure}

\cref{fig:moti_medmnist1} denotes the overall design of the motivational experiment,
using biomedical image datasets from \citet{medmnistv2}. 
Assuming the absence of the dataset used for pretraining a given model, we use another dataset in the collection for additional adversarial training steps~\citep{madry}. 
Due to the different label spaces, we use teacher outputs as soft labels (i.e., $KL(S(x')\Vert T(x))$).
\cref{fig:moti_medmnist2} shows the PGD-10 ($l_{\infty}$, $\epsilon=8/255$) evaluation results using ResNet-18.
Each row represents the dataset used for (non-robustly) pretraining a model, and each column represents the dataset used for additional adversarial training steps.

It is clear that models adversarially trained using alternative datasets show poor robustness compared to those trained using the original dataset (the diagonal cells).
Although there exist a few combinations that obtain minor robustness from other datasets (e.g., Path \textrightarrow \ Tissue), they still suffer from large degradation compared to that of AT using the original dataset.
Moreover, using a publicly available general domain dataset (CIFAR-10) also performs poorly, indicating that adversarial robustness is difficult to obtain from other datasets without access to the data in the same domain.

\begin{figure*}[t]
    \centering
    \includegraphics[width=.82\textwidth]
    {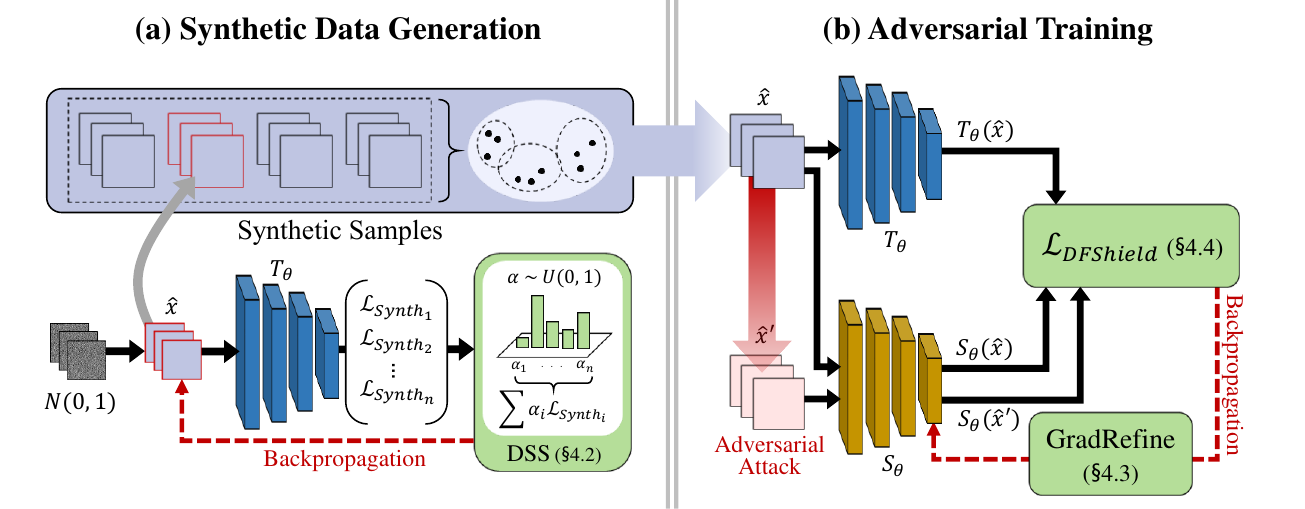}
    \vspace{-4mm}
\caption{Procedure of the proposed method. (a) denotes synthetic data generation using the proposed \mixabb. (b) shows adversarial training of target model $S_{\theta}$ using $\lossdfs$ and \gradmask. The pseudo-code is provided in \cref{sec:supp:pseudocode}.} 
\label{fig:dfar} 
\end{figure*}

\section{\name: Learning Data-free Adversarial Robustness}
To tackle the data-free adversarial robustness problem,
we propose \emph{\name}, an effective solution to improve the robustness of the target model without any real data. 
First, we generate a synthetic surrogate dataset using the information of the pretrained model $T_{\theta}$ (\cref{fig:dfar}(a)). 
Then, we use the synthetic dataset to adversarially train $S_{\theta}$ initialized with $T_{\theta}$ (\cref{fig:dfar}(b)). 
For generation, we propose \mix for dataset diversity (\S\ref{sec:mix}).
For training, we propose a gradient refinement (\S\ref{sec:gradmask}) method and a soft-label guided objective function (\S\ref{sec:Ltrain}).

\subsection{Key Challenges}
\label{sec:challenges}
Although AT using a synthetic dataset seems like a promising approach, naively conducting the plan yields poor performance.
We identify the following challenges to achieve high adversarial robustness.

\textbf{Challenge 1: Limited Diversity of Synthetic Samples.} 
Robust training of DNNs is known to require higher sample complexity~\citep{khim2018adversarial, yin2019rademacher} and substantially more data~\citep{schmidt2018adversarially} than standard training.
Recent findings~\citep{rebuffi2021data, aug_alone} appoint diversity as the key contributing factor to adversarial robustness, and \citep{proxy} showed that enlarging train set is helpful only if it adds to the diversity of the data.
Unfortunately, diversity is particularly hard to achieve in data-free methods, as they do not have direct access to training data distributions.
While there exists a few data-free diversification techniques~\citep{qimera, intraq, rdskd}, they show negligible improvement in diversity, and often come at the price of low fidelity and quality.
More importantly, none of these works contribute to enhancing adversarial robustness (See \cref{sec:analysis_diversity}).

\begin{figure*}
\captionsetup[subfigure]{justification=centering}
\centering
     \begin{subfigure}{.25\textwidth}
    \includegraphics[width=\textwidth]{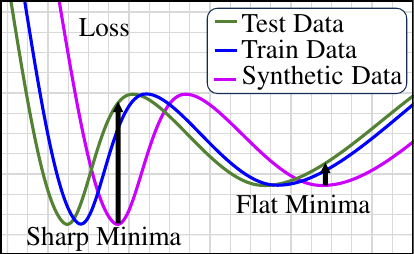}
    \caption{A conceptual diagram of generalization gap. } 
    \label{fig:lossvis:concept}
    \end{subfigure} 
\hspace{0.5mm}
    \begin{subfigure}{.17\textwidth}
    \includegraphics[width=\textwidth]{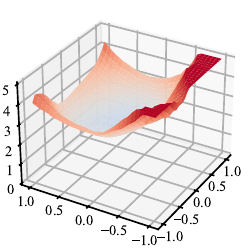}
    \caption{TRADES \\ w/o \gradmask}
    \label{fig:lossvis:trades_nomask}
    \end{subfigure} 
    \begin{subfigure}{.17\textwidth}
    \includegraphics[width=\textwidth]{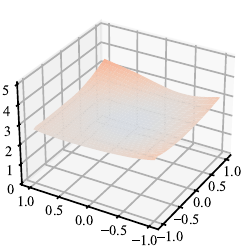}
    \caption{TRADES \\ w/ \gradmask}
    \label{fig:lossvis:trades_mask}
    \end{subfigure} 
    \begin{subfigure}{.17\textwidth}
    \includegraphics[width=\textwidth]{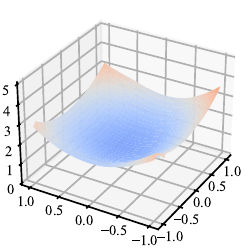}
    \caption{$\lossdfs$ \\ w/o \gradmask}
    \label{fig:lossvis:dfs_nomask}
    \end{subfigure} 
    \begin{subfigure}{.17\textwidth}
    \includegraphics[width=\textwidth]{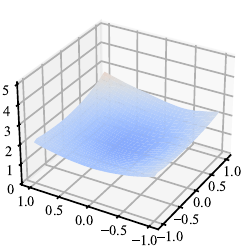}
    \caption{$\lossdfs$ \\ w/ \gradmask}
    \label{fig:lossvis:dfs_mask}
    \end{subfigure} 
    \caption{(a) demonstrates a conceptual image of generalization gap between synthetic and real data. (b)-(e) shows loss surface visualization on ResNet-20 with CIFAR-10 showing that \gradmask achieves flatter loss surfaces. Each figure represents different training losses with or without \gradmask. We use normalized random direction for $x$,$y$ axis, following \citet{li2018visualizing}.  }
    \label{fig:loss_surf_vis}
\end{figure*}

\textbf{Challenge 2: Poor Generalization to Real Adversarial Samples.}
The ultimate goal of AT is to learn robustness that can be generalized to unseen adversaries.
However, AT is known to be highly prone to robust overfitting~\cite{robustoverfitting, flatminima, weightperturb, liu2020loss}, where the model often fails to generalize the learned robustness to unseen data.
Unfortunately, such difficulty is more severe in our problem. 
While the conventional overfitting problem is caused by the distributional gap between the training and test data, synthetic data will undergo a more drastic shift from the training data, further widening the gap. 
As illustrated in \cref{fig:lossvis:concept}, this double gap causes a large degradation in the test robustness.
Many data-free learning methods study this distributional gap between synthetic and real data~\cite{deepinversion, PSAQ-VIT, ait}, but none of them have addressed the problem under the light of adversarial robustness.

\subsection{\MIX}
\label{sec:mix}
To address the first key challenge of generating diverse samples, we propose 
a novel diversifying technique called \emph{\mix} (\mixabb).
We choose to directly optimize each sample one by one from a normal distribution $N(0,1)$ through backpropagation using an objective function ($\loss_{Synth}$)~\citep{deepinversion, zeroq, intraq}.
In \mixabb, we leverage its characteristic to enhance the diversity of the samples, where we dynamically modulate the synthesis loss $\loss_{Synth}$.
We first formulate $\loss_{Synth}$ as a weighted sum of multiple losses. 
Then the weights are randomly set for every batch, giving each batch a distinct distribution.
Given a set $\mathit{\mathbbm{S}} = \{\mathit{\loss_{Synth_1}}, \mathit{\loss_{Synth_2}}, ..., \mathit{\loss_{Synth_n}}\}$, 
the conventional approaches use their weighted sum with fixed hyperparameters as in \cref{eq:synth_all}.
On the other hand, we use coefficients $\alpha_i$ differently sampled for every batch from a continuous space:
\begin{align}
\mathit{\loss_{Synth}}= \sum_{i=1}^{|\mathbbm{S}|}\alpha_i \mathit{\loss_{Synth_i}}, \qquad \alpha_i \sim U(0,1). 
\label{eq:mix}
\end{align}
For the set $\mathbbm{S}$, we use the three terms from \cref{eq:synth_all}.
The sampling of coefficients can follow any arbitrary distribution, where we choose a uniform distribution.

\begin{figure*}[t]
    \centering
    \begin{subfigure}{.26\textwidth}
    \includegraphics[width=\textwidth]{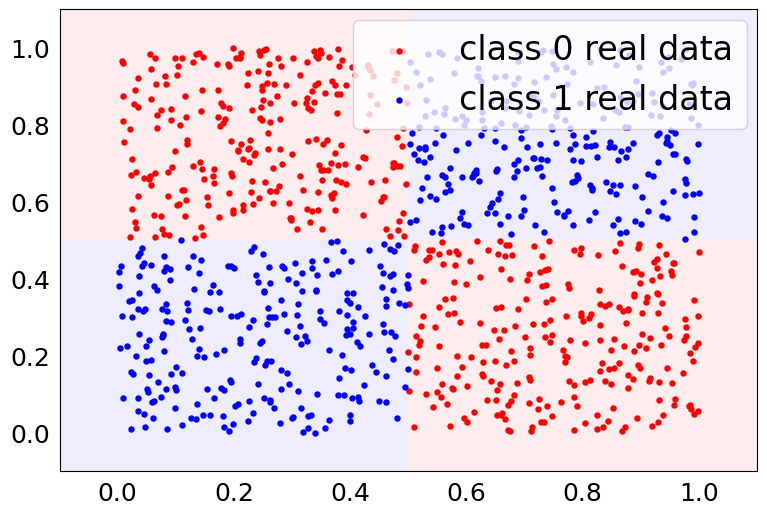}
    \caption{Real data}
    \label{fig:toy_real}
    \end{subfigure} 
    \hspace{0.3mm}
    \begin{subfigure}{.26\textwidth}
    \includegraphics[width=\textwidth]{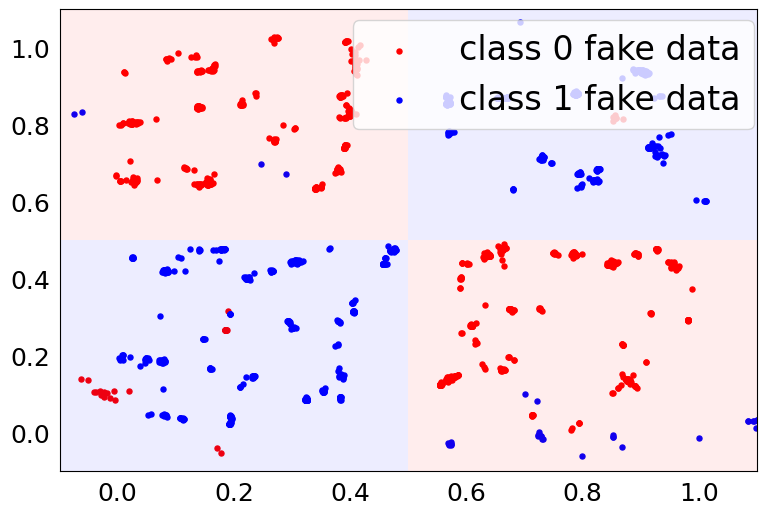}
    \caption{Fake data (Fixed coefficient)}
    \label{fig:toy_fake1}
    \end{subfigure} 
    \hspace{0.3mm}
    \begin{subfigure}{.26\textwidth}
    \includegraphics[width=\textwidth]{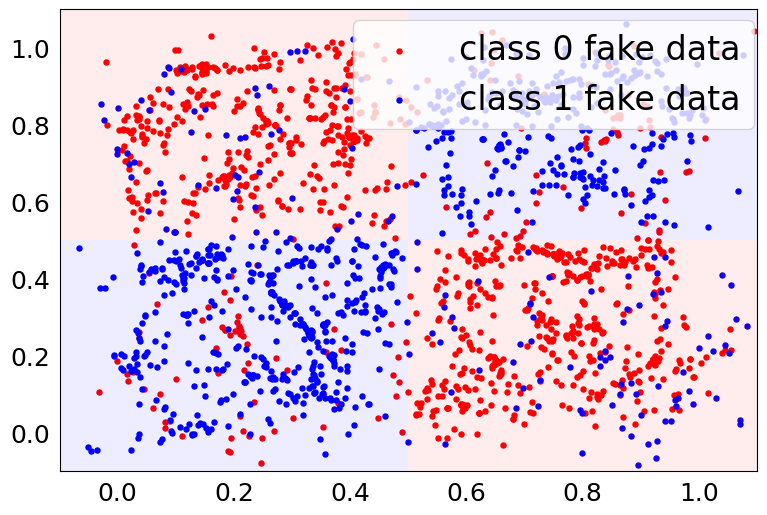}
    \caption{Fake data (Proposed)}
    \label{fig:toy_fake2}
    \end{subfigure} 
    \caption{Comparison of synthesis methods using the same number of 2-d data. The conventional fixed coefficient setting leads to limited diversity, while \mixabb generates diverse samples.}
    \label{fig:toy_example}
\end{figure*}

\textbf{A Toy Experiment.}
 To demonstrate the effectiveness \mixabb has on sample diversity, we conducted an empirical study on 
 a toy experiment. 
\cref{fig:toy_example} displays the simplified experiment using 2-d data.
The real data distribution is depicted in \cref{fig:toy_real}.
Using the real data, we train a 4-layer network with batch normalization. 
\cref{fig:toy_fake1} demonstrates the results from conventional approaches (fixed coefficients following \citep{deepinversion}).
Although the data generally follows class information, 
they are highly clustered with small variance. 
On the other hand, \cref{fig:toy_fake2} shows the data generated using \mixabb, 
which are highly diverse and exhibit coverage much closer to that of the real data distribution.
In addition, we observe noisy samples in both \cref{fig:toy_fake1} and \cref{fig:toy_fake2}. 
This is due to the nature of the synthetic data generation process, which uses artificial labels to guide synthetic samples towards given arbitrary classes.
As these noisy samples may harm the training, we address this problem in  \cref{sec:Ltrain}.

\subsection{Gradient Refinement for Smoother Loss Surface} 
\label{sec:gradmask}
As discussed in \cref{sec:challenges}, the large distributional gap between synthetic and test data causes performance degradation.
In such a case, searching for a flatter minima is a better strategy than searching for a low, but sharp minima as illustrated in \cref{fig:lossvis:concept}.
For this, we devise a novel gradient refinement technique \emph{\gradmask}. 
Inspired by a few techniques from domain generalization and federated learning~\citep{tenison2022gradient, mansilla2021domain}, \gradmask regularizes the influence of rapidly changing gradients during training.
After computing gradients $g$ from $\mathcal{B}$ mini-batches, 
we calculate the agreement score ${A}_{k}$ for each parameter $k$ as:
\begin{equation}
{A}_{k} = \frac{1}{\mathcal{B}}  \sum_{b=1}^{\mathcal{B}} sign(g_k^{(b)}).
\label{eq:gradmask}
\end{equation}
Intuitively, ${A}$ denotes the amount which one sign dominates the other. 
${A}$ is bounded by [-1,1], where ${A}=0$ means equal distribution in both signs (maximum disagreement), and ${A}=\pm 1$ means one sign completely dominates the other (maximum agreement).
Using ${A}$, we compute the final gradient $g^*_{k}$ that will be used for parameter $k$ update:
\begin{equation}
\begin{split}
    g^*_{k}&=
        \Phi({A}_k) \sum^{\mathcal{B}}_{b=1} \mathbbm{1}_{\{{A}_{k} \cdot g^{(b)}_{k}>0\}} \cdot g^{(b)}_{k},\\
     \Phi({A}_k)&=
    \begin{cases}
        1, & \text{if }\left|{A_k}\right| \geq \tau,\\
        0, & \text{otherwise},
    \label{eq:gradmask3}
    \end{cases}
\end{split}
\end{equation}
where $\mathbbm{1}(\cdot)$ is the indicator function, and $\Phi$ is a masking function. 
We use $\tau$ value of 0.5, 
which indicates that one should dominate the other for more than half its entirety.
This allows high-fluctuating parameters to be ignored by $\Phi$, and we further pursue alignment via selective use of agreeing gradient elements.
\cref{fig:loss_surf_vis} (b)-(e) visualizes the effect of \gradmask on the loss surface. 
In both TRADES~\citep{trades} and $\lossdfs$ (\S\ref{sec:Ltrain}), \gradmask yields a flatter loss surface, contributing towards better performance.
Please refer to \cref{sec:supp:trainloss} for the full set of visualization.

\setlength{\tabcolsep}{4pt}
\begin{table*}[]
\centering
    \caption{Performance on \medmnist with $l_\infty$ perturbation budget.}
            \vspace{1mm}
    \def\arraystretch{0.88}%
\resizebox{.82\textwidth}{!}{

\begin{tabular}{clcccccccccccc}
\midrule
\multirow{2}{*}{Model} &
\multirow{2}{*}{\makecell{Method}}                                                                                                       
                         & \multicolumn{3}{c}{Tissue} & \multicolumn{3}{c}{Blood} & \multicolumn{3}{c}{Path} & \multicolumn{3}{c}{OrganC}  \\
                         \cmidrule(lr){3-5}  \cmidrule(lr){6-8}  \cmidrule(lr){9-11}  \cmidrule(lr){12-14} 
                         & & $\mathcal{A}_{Clean}$ & $\bf\mathcal{A}_{PGD}$ & $\bf\mathcal{A}_{AA}$      &      $\mathcal{A}_{Clean}$ & $\bf\mathcal{A}_{PGD}$ & $\bf\mathcal{A}_{AA}$ &
                         $\mathcal{A}_{Clean}$ & $\bf\mathcal{A}_{PGD}$ & $\bf\mathcal{A}_{AA}$  &
                         $\mathcal{A}_{Clean}$ & $\bf\mathcal{A}_{PGD}$ & $\bf\mathcal{A}_{AA}$  \\ 
\midrule
 \multirowcell{6}{RN-18} 

& Public &		22.04	&	{\textcolor{white}0}0.02	&	{\textcolor{white}0}0.00	&	{\textcolor{white}0}9.09{$^\dagger$}	&	{\textcolor{white}0}9.09	&	{\textcolor{white}0}0.00	&	13.30{$^\dagger$}	&	{\textcolor{white}0}0.00	&	{\textcolor{white}0}0.00	&	79.41	&	40.10	&	36.53	\\						
\cmidrule{2-14}	
																						
& DaST  &		23.27	&	{\textcolor{white}0}7.01	&	{\textcolor{white}0}5.98	&	16.92	&	{\textcolor{white}0}6.75	&	{\textcolor{white}0}4.82	&	{\textcolor{white}0}7.49	&    {\textcolor{white}0}3.36	& {\textcolor{white}0}1.20	&	83.13	&	27.91	&	24.49	\\
&  DFME&		{\textcolor{white}0}7.01	&	{\textcolor{white}0}4.33	&	{\textcolor{white}0}4.17	&	46.59	&	{\textcolor{white}0}0.20	&	{\textcolor{white}0}0.03	&	76.43	&   {\textcolor{white}0}0.50	& {\textcolor{white}0}0.38	&	79.73	&	19.27	&	17.19	\\
&  AIT &		15.62	&	11.64	&	{\textcolor{white}0}9.72	&	18.24	&	10.55	&	{\textcolor{white}0}1.64	&	16.66	&  10.24   	& {\textcolor{white}0}3.89	&	56.85	&	18.02	&	16.67	\\
& DFARD &		{\textcolor{white}0}9.31	&	{\textcolor{white}0}8.48	&	{\textcolor{white}0}1.87	&	22.60	&	10.17	&	{\textcolor{white}0}9.70	&	11.59	&  {\textcolor{white}0}4.93    &   {\textcolor{white}0}3.18	&	81.97	&	21.71	&	19.50	\\
\cmidrule{2-14}																									
& \textbf{\nameabb}   &		32.07	&	\textbf{31.63}	&	\textbf{31.57}	&	59.89	&	\textbf{21.72}	&	\textbf{19.29}	&	33.06	&   \textbf{29.78}   &   \textbf{25.38}	&	83.35	&	\textbf{47.01}	&	\textbf{42.56}	\\
\midrule																									
\multirowcell{6}{RN-50}

& Public &		27.84	&	10.11	&	{\textcolor{white}0}8.64	&	{\textcolor{white}0}9.09{$^\dagger$}	&	{\textcolor{white}0}9.09	&	{\textcolor{white}0}0.00	&	{\textcolor{white}0}7.54	&{\textcolor{white}0}1.21&	{\textcolor{white}0}0.37	&	84.41	&	46.12	&	43.44	\\
\cmidrule{2-14}	
																	
& DaST  &		{\textcolor{white}0}4.73	&	{\textcolor{white}0}1.36	&	{\textcolor{white}0}0.05	&	{\textcolor{white}0}9.12	&	{\textcolor{white}0}8.77	&	{\textcolor{white}0}8.16	&	{\textcolor{white}0}8.25	&  {\textcolor{white}0}6.92    &   {\textcolor{white}0}2.12	&	21.03	&	{\textcolor{white}0}9.18	&	{\textcolor{white}0}8.36	\\
&  DFME&		{\textcolor{white}0}7.13	&	{\textcolor{white}0}6.55	&	{\textcolor{white}0}4.76	&	{\textcolor{white}0}7.16	&	{\textcolor{white}0}3.36	&	{\textcolor{white}0}3.19	&	80.10	&  {\textcolor{white}0}2.28	  &  {\textcolor{white}0}2.01	&	27.76	&	22.00	&	21.78	\\
&  AIT &		32.08	&	{\textcolor{white}0}4.75	&	{\textcolor{white}0}0.74	&	19.47	&	12.48	&	{\textcolor{white}0}9.94	&	14.29	&  10.00	& {\textcolor{white}0}2.21	&	15.34	&	{\textcolor{white}0}8.90	&	{\textcolor{white}0}6.02	\\
& DFARD &		23.69	&	12.99	&	{\textcolor{white}0}7.01	&	26.63	&	{\textcolor{white}0}9.21	&	{\textcolor{white}0}0.00	&	14.04	&   {\textcolor{white}0}2.44 	&  {\textcolor{white}0}0.77	&	80.99	&	11.93	&	{\textcolor{white}0}8.13	\\
\cmidrule{2-14}																									
& \textbf{\nameabb}   &		31.91	&	\textbf{27.15}	&	\textbf{26.68}	&	74.63	&	\textbf{36.07}	&	\textbf{30.17}	&	41.63	&   \textbf{15.35}	& \textbf{12.28}   	&	86.56	&	\textbf{62.60}	&	\textbf{59.86}	\\
\bottomrule

 \multicolumn{14}{r}{$^\dagger$Did not converge}\\
\end{tabular}
}
\vspace{-5mm}
    \label{tab:medmnist}

\end{table*}

\subsection{Training Objective Function} 
\vspace{-1mm}
\label{sec:Ltrain} 
There exist several objective functions for adversarial training~\citep{trades, mart, ard, rslad}.
However, those objective functions mostly rely on the hard label $y$ of the dataset.
For synthetic data, the assigned artificial labels are not ground truths, but simply target for optimization using cross-entropy loss.
In such circumstances, relying on these artificial labels 
could convey incorrect guidance.
As such, we devise a new objective function $\mathit{\lossdfs}$ that does not rely on the hard label, but only utilizes the soft guidance from $T(\hat{x})$ using KL-divergence.
\vspace{-2mm}
\begin{multline}
\mathit{\loss_{Train}}=
\mathit{\lossdfs}=
  \overbrace{KL(S(\hat{x}),T(\hat{x}))}^\textrm{ \hypertarget{loss:first}{(a) clean accuracy} } \\
+ \lambda_1 \underbrace{KL(S(\hat{x}'),T(\hat{x}))}_\textrm{\hypertarget{loss:second}{(b) robustness training }} 
+ \lambda_2 \underbrace{KL(S(\hat{x}'),S(\hat{x}))}_\textrm{\hypertarget{loss:third}{(c) smoothness term} }.
\label{eq:triple}
\end{multline}
The first term \hyperlink{loss:first}{(a)} optimizes the accuracy on clean samples, and can be thought as a replacement for the common cross-entropy loss. 
The second term \hyperlink{loss:second}{(b)} serves the purpose of learning adversarial robustness similar to the cross-entropy loss in standard AT (\cref{eq:adv}).
Adversarial samples exist because classifiers tend to rapidly change their decisions~\citep{yang2020closer}, due to failing to learn general features and instead relying on trivial features.
To mitigate this, we add a smoothness term \hyperlink{loss:third}{(c)} which penalizes rapid changes in the model's output.
This regularizes the model's sensitivity to small variations in the input,
helping to train the target model to be stable under small perturbations.

\begin{table}[bp]
    \centering
    \vspace{-2mm}
    \caption{Loss functions of baseline approaches.}
            \vspace{1mm}
     \resizebox{.93\columnwidth}{!}
     {
    \begin{tabular}{lcc}
    \toprule
    \multirow{1}{*}{Baselines} & $\loss_{\mathit{Synth}}$ & $\loss_{\mathit{train}}$  \\
    \midrule
\multirow{1}{*}{DaST} & $-\loss_{CE}(S(x), y)$ & $\loss_{CE}(S(x^\prime), y)$\\ 
\multirow{1}{*}{DFME} & $-\sum \lvert T(x) - S(x) \lvert$ & $\sum \lvert T(x) - S(x^\prime) \lvert$ \\
\multirow{1}{*}{AIT} & $\loss_{feature} + \loss_{CE}(T(x),y)$ & $\mathit{\loss_{TRADES}}$ \\
\multirow{1}{*}{DFARD} & $-\loss_{KL}(S(x),T(x),\Tilde{\tau})$ & $\loss_{KL}(S(x'),T(x),\Tilde{\tau})$ \\
      \bottomrule
    \end{tabular} 
     } 
    \label{tab:baseline_losses}
\end{table}

\section{Evaluation}
\subsection{Experimental Setup}
\label{sec:eval:setup}
We use total of four datasets: MedMNIST-v2 as \medmnist~\citep{medmnistv2}, SVHN~\citep{svhn}, CIFAR-10, and CIFAR-100~\citep{cifar}.
For \medmnist{}, we use ResNet-18 and ResNet-50 as pretrained networks, and $l_{\infty}, \epsilon=8/255$ for perturbation budget.
For CIFAR and SVHN datasets, we chose three pretrained models from PyTorchCV~\citep{pytorchcv} library: ResNet-20, ResNet-56~\citep{resnet}, and WRN-28-10~\citep{wrn}.
We use $l_{\infty}, \epsilon=4/255$ perturbation budget for SVHN, CIFAR-10, and CIFAR-100, and additionally examine $l_{2}, \epsilon=128/255$ setting. 
For results on extensive perturbation settings, please refer to \cref{sec:supp:eps}.
For evaluation, AutoAttack~\citep{autoattack} accuracy (denoted $\bf\mathcal{A}_{AA}$) is generally perceived as the standard metric~\citep{robustbench}. 
While we regard $\bf\mathcal{A}_{AA}$ as the primary interest, we also report the clean accuracy ($\mathcal{A}_{Clean}$) and PGD-10 accuracy ($\bf\mathcal{A}_{PGD}$) for interested readers.
Further details of experimental settings can be found in \cref{sec:supp:expsetting}.

\subsection{Baselines}
\label{sec:baseline}

Since our work tackles a less-studied problem of data-free adversarial robustness with no known clear solution, it is important to set an adequate baseline for comparison.
We choose four of the most relevant works of other data-free learning tasks that generate synthetic samples to replace the original: DaST~\citep{dast} from a black-box attack method, DFME~\citep{dfme} from data-free model extraction, AIT~\citep{ait} from data-free quantization, and DFARD~\citep{dfard} from data-free robust distillation.
Since these four methods are not designed specifically for the data-free adversarial robustness problem, we adapt the training objective, summarized in \cref{tab:baseline_losses}.
We also compare our method against test-time defense methods, including DAD~\citep{dad}, TTE~\citep{TTE}, and DiffPure~\citep{diffpure} on \medmnist{}.
For details on the implementations, please refer to the experimental settings in \cref{sec:supp:expsetting:baseline}.

\begin{table}[]
    \centering
    
    \caption{Performance on \medmnist{} with $l_\infty$ perturbation budget using test-time defense methods.}
            \vspace{1mm}
    \label{tab:main_medmnist_post_train}
    
    \def\arraystretch{0.85}%
    \resizebox{\columnwidth}{!}
    {
    \begin{tabular}{clcccccc}
    \toprule
     & \multicolumn{1}{r}{} &  \multicolumn{3}{c}{ResNet-18} & \multicolumn{3}{c}{ResNet-50} \\
    \cmidrule(lr){3-5}\cmidrule(lr){6-8}
      Dataset  & Method &   $\mathcal{A}_{Clean}$& $\bf\mathcal{A}_{PGD}$ & $\bf\mathcal{A}_{AA}$ & $\mathcal{A}_{Clean}$& $\bf\mathcal{A}_{PGD}$ & $\bf\mathcal{A}_{AA}$ \\
      \midrule

\multirowcell{4}{Tissue} 

&   DAD &   55.86	&	22.90	&	{\textcolor{white}0}4.38  & 		59.72	&	\textbf{31.59}	&	{\textcolor{white}0}3.49	 \\
& DiffPure &		26.17	&	22.85	&	{\textcolor{white}0}9.06	&		27.73	&	27.54	&	{\textcolor{white}0}1.81	\\
& TTE &		56.60	&	{\textcolor{white}0}0.00	&	{\textcolor{white}0}0.00	&		62.01	&	{\textcolor{white}0}0.00	&	{\textcolor{white}0}0.00	\\

\cmidrule{2-8}
&  \textbf{\nameabb} &		32.07	&	\textbf{31.63}	&	\textbf{31.57} &		31.91	&	27.15	&	\textbf{26.68}	\\
\midrule

\multirowcell{4}{Blood} 

&   DAD &	91.96	&	17.25	&	{\textcolor{white}0}0.00	&83.46	&	34.43	&	{\textcolor{white}0}0.00	\\
& DiffPure  &   49.02	&	\textbf{29.10}	&	{\textcolor{white}0}8.71	&	51.17	&	\textbf{36.91}	&	13.77	\\
& TTE &	{\textcolor{white}0}9.09{$^{\dagger}$}	&	{\textcolor{white}0}9.09	&	{\textcolor{white}0}8.92	&	16.84	&	{\textcolor{white}0}0.03	&	{\textcolor{white}0}0.00	\\

\cmidrule{2-8}
&  \textbf{\nameabb} &	59.89	&	21.72	&	\textbf{19.29}	&	74.63	&	36.07	&	\textbf{30.17}	\\
\midrule  																	
\multirowcell{4}{Path} 

&   DAD &	91.28	&   15.54	& {\textcolor{white}0}0.21	&	81.50	&    12.79	& {\textcolor{white}0}1.38\\
& DiffPure &	19.73	&  18.95	&    {\textcolor{white}0}8.91	&	14.65	&    14.26	& \textbf{13.79}	\\
& TTE &	76.56	& {\textcolor{white}0}0.64   &	{\textcolor{white}0}0.36	&	75.08	&   {\textcolor{white}0}4.23	&  {\textcolor{white}0}1.88	\\

\cmidrule{2-8}
&  \textbf{\nameabb} &	33.06	&   \textbf{29.78}   &   \textbf{25.38}	&	41.63	&   \textbf{15.35}	& 12.28   	\\
\midrule 

\multirowcell{4}{OrganC} 

&   DAD & 80.19	&	31.22	&	12.57 &	87.54	&	25.46	&	{\textcolor{white}0}7.84	\\
& DiffPure &	69.73	&	\textbf{57.03}	&	19.00&	58.20	&	51.76	&	34.38\\
& TTE &	61.03	&	22.90	&	15.98	&	56.54	&	25.82	&	18.63	\\

\cmidrule{2-8}
&  \textbf{\nameabb} &	83.35	&	47.01	&	\textbf{42.56}	&	86.56	&	\textbf{62.60}	&	\textbf{59.86}	\\
\bottomrule
     \multicolumn{8}{r}{$^\dagger$Did not converge}\\
    \end{tabular}
    }
   \vspace{-5mm}
\end{table}

\subsection{Performance Comparison}

\textbf{Privacy Sensitive Dataset.} \cref{tab:medmnist} shows experimental results for \medmnist{}, compared against the baseline methods (\S\ref{sec:baseline}).
The experiments represent a scenario close to real life where classification models are used for specific domains in the absence of public datasets from the same/similar domains.
In all cases, \name achieves the best results under $\mathcal{A}_{AA}$ evaluation.
The baselines often perform worse than simply using public datasets of different domains.
For example, in OrganC, using CIFAR-10 leads to some meaningful robustness. 
This could be because those datasets share similar features with CIFAR-10. 
Nonetheless, \name performs significantly better in all cases.

We also show the limited applicability of existing test-time defense techniques in \cref{tab:main_medmnist_post_train}. 
Although they work relatively well on general domain data, they perform poorly on privacy-sensitive datasets with a large distributional gap to general ones.
For example, DiffPure, which is known to show superior performance to AT methods, fails to show practical performance in most cases.
Similarly, DAD performs poorly against AutoAttack, and TTE shows close-to-zero robustness in ResNet-50 in most settings.

\textbf{General Domain Datasets.} In \cref{tab:main}, the performance of \name is compared against the baselines on more general domain datasets: SVHN, CIFAR-10, and CIFAR-100. 
\name outperforms the baselines by a huge margin. 
The improvements reach up to 23.19\%p in $\mathcal{A}_{AA}$, revealing the effectiveness of \name and that the result is not from gradient obfuscation~\citep{autoattack}.
Aligned with previous findings~\citep{schmidt2018adversarially, archview}, models with larger capacity (ResNet-20 \textrightarrow \ ResNet-56 \textrightarrow \ WRN-28-10) tend to have significantly better robust accuracy of up to 21.08\%p difference under AutoAttack. 
However, the baselines were often unable to exploit the model capacity (e.g., 19.65\% \textrightarrow \ 14.57\% in ResNet-56 \textrightarrow \ WRN-28-10 with DaST on SVHN), we believe this is due to the limited diversity of their synthetic samples. 
A similar trend can be found from the experiments with $l_2$ perturbation budgets shown in \cref{tab:main_l2}, where we compare with the best-performing baseline AIT from \cref{tab:main}.
Extended results using different budgets are presented in \cref{sec:supp:eps}.

\setlength{\tabcolsep}{2pt}
\begin{table}[t]
    
    \centering
    
    \caption{Performance on SVHN, CIFAR-10, and CIFAR-100 with $l_{\infty}$ perturbation budget.}
        \vspace{1mm}
    \label{tab:main}
    
    \def\arraystretch{0.95}%
    \resizebox{\columnwidth}{!}
    {
    \begin{tabular}{lccccccccc}
    \toprule
      & \multicolumn{3}{c}{ResNet-20} & \multicolumn{3}{c}{ResNet-56} & \multicolumn{3}{c}{WRN-28-10} \\
    \cmidrule(lr){2-4}\cmidrule(lr){5-7}\cmidrule(lr){8-10}
      \it SVHN &  $\mathcal{A}_{Clean}$ & $\bf\mathcal{A}_{PGD}$&$\bf\mathcal{A}_{AA}$ & $\mathcal{A}_{Clean}$& $\bf\mathcal{A}_{PGD}$& $\bf\mathcal{A}_{AA}$ & $\mathcal{A}_{Clean}$& $\bf\mathcal{A}_{PGD}$ & $\bf\mathcal{A}_{AA}$ \\
      \midrule

DaST & 20.66 & 13.90 &{\textcolor{white}0}7.06 & 10.55 & {\textcolor{white}0}0.25  & {\textcolor{white}0}0.00 & 20.15 & 19.17  &14.57	\\													
DFME & 11.32& {\textcolor{white}0}2.59 & {\textcolor{white}0}0.84& 20.20& 19.22 & {\textcolor{white}0}4.27 & {\textcolor{white}0}6.94& {\textcolor{white}0}5.31 &	{\textcolor{white}0}0.28\\
AIT &	91.45	& 37.87	&	24.74 &	86.65	&	45.45	&	38.96 & 83.89	&	40.45	&	33.06	\\
DFARD	&	25.62	&	18.65	&	{\textcolor{white}0}0.19	&	19.58	&	15.43	&	{\textcolor{white}0}0.00	&	92.32	&	13.08	&	{\textcolor{white}0}0.01	\\
\cmidrule(lr){1-10}																			
\textbf{\nameabb} 	&	91.83	&	\textbf{54.82}	&	\textbf{47.55}	&	88.66	&	\textbf{62.05}	&	\textbf{57.54}	&	94.14	&	\textbf{69.60}	&	\textbf{62.66}	\\

\midrule	
      \it CIFAR-10 &  $\mathcal{A}_{Clean}$ & $\bf\mathcal{A}_{PGD}$ & $\bf\mathcal{A}_{AA}$ & $\mathcal{A}_{Clean}$& $\bf\mathcal{A}_{PGD}$ & $\bf\mathcal{A}_{AA}$ & $\mathcal{A}_{Clean}$& $\bf\mathcal{A}_{PGD}$ & $\bf\mathcal{A}_{AA}$ \\
      \midrule
DaST & 10.00$^{\dagger}$ & {\textcolor{white}0}9.89 & {\textcolor{white}0}8.62 & 12.06 &{\textcolor{white}0}7.68 & {\textcolor{white}0}5.32 & 10.00$^{\dagger}$ & {\textcolor{white}0}9.65 & {\textcolor{white}0}2.85	\\
DFME & 14.36& {\textcolor{white}0}5.23 & {\textcolor{white}0}0.08& 13.81& {\textcolor{white}0}3.92 & {\textcolor{white}0}0.03& 10.00$^{\dagger}$& {\textcolor{white}0}9.98 &	{\textcolor{white}0}0.05\\
AIT &	32.89	&	11.93	&	10.67 &	38.47	&	12.29	&	11.36 &	34.92	&	10.90	&	{\textcolor{white}0}9.47	\\
DFARD	&	12.28	&	{\textcolor{white}0}5.33	&	{\textcolor{white}0}0.00	&	10.84	&	{\textcolor{white}0}8.93	&	{\textcolor{white}0}0.00	&	{\textcolor{white}0}9.82	&	12.01	&	{\textcolor{white}0}0.02	\\
\cmidrule(lr){1-10}																			
\textbf{\nameabb} 	&	74.79	&	\textbf{29.29}	&	\textbf{22.65}	&	81.30	&	\textbf{35.55}	&	\textbf{30.51}	&	86.74	&	\textbf{51.13}	&	\textbf{43.73}	\\

\midrule	
      \it CIFAR-100 &  $\mathcal{A}_{Clean}$ & $\bf\mathcal{A}_{PGD}$ & $\bf\mathcal{A}_{AA}$ & $\mathcal{A}_{Clean}$& $\bf\mathcal{A}_{PGD}$ & $\bf\mathcal{A}_{AA}$ & $\mathcal{A}_{Clean}$& $\bf\mathcal{A}_{PGD}$ & $\bf\mathcal{A}_{AA}$ \\
      \midrule
DaST &  {\textcolor{white}0}1.01$^{\dagger}$ &   {\textcolor{white}0}0.99 &  {\textcolor{white}0}0.95 &  {\textcolor{white}0}1.13  &  {\textcolor{white}0}0.72 &  {\textcolor{white}0}0.34 &   {\textcolor{white}0}1.39 &   {\textcolor{white}0}0.66  &  {\textcolor{white}0}0.18\\
DFME &   {\textcolor{white}0}1.86 &   {\textcolor{white}0}0.53 &  {\textcolor{white}0}0.24 & 24.16  &  {\textcolor{white}0}0.98 &   {\textcolor{white}0}0.25 &  66.30 &   {\textcolor{white}0}0.67  &  {\textcolor{white}0}0.00	\\
AIT &  {\textcolor{white}0}7.92 &  {\textcolor{white}0}2.51 &  {\textcolor{white}0}1.39  &  {\textcolor{white}0}9.68 &  {\textcolor{white}0}2.97 &  {\textcolor{white}0}2.04 &  22.21 &  {\textcolor{white}0}3.11  &  {\textcolor{white}0}1.28 \\
DFARD	& 66.59 &  {\textcolor{white}0}0.02 &  {\textcolor{white}0}0.00 & 69.20 &  {\textcolor{white}0}0.26 &  {\textcolor{white}0}0.00 & 82.03 &  {\textcolor{white}0}1.10  &  {\textcolor{white}0}0.00\\
\cmidrule(lr){1-10}																			
\textbf{\nameabb} 	&	41.67	&	\textbf{10.41}	&	 \textbf{{\textcolor{white}0}5.97}	&	39.29	&	\textbf{13.23}	&	 \textbf{{\textcolor{white}0}9.49}	&	61.35	&	\textbf{23.22}	&	\textbf{16.44}	\\

     \bottomrule
     \multicolumn{10}{r}{$^\dagger$Did not converge}\\
    \end{tabular}
    }
\end{table}

\begin{table}[]
    \centering
    \setlength{\tabcolsep}{3pt}
    \caption{Performance on SVHN, CIFAR-10, and CIFAR-100 with $l_{2}$ perturbation budget.}
            \vspace{1mm}
    \label{tab:main_l2}
    
    \def\arraystretch{0.8}%
    \resizebox{\columnwidth}{!}
    {
    \begin{tabular}{lccccccccc}
    \toprule
      & \multicolumn{3}{c}{ResNet-20} & \multicolumn{3}{c}{ResNet-56} & \multicolumn{3}{c}{WRN-28-10} \\
    \cmidrule(lr){2-4}\cmidrule(lr){5-7}\cmidrule(lr){8-10}
      \it SVHN &  $\mathcal{A}_{Clean}$ & $\bf\mathcal{A}_{PGD}$&$\bf\mathcal{A}_{AA}$ & $\mathcal{A}_{Clean}$& $\bf\mathcal{A}_{PGD}$& $\bf\mathcal{A}_{AA}$ & $\mathcal{A}_{Clean}$& $\bf\mathcal{A}_{PGD}$ & $\bf\mathcal{A}_{AA}$ \\
      \midrule

	AIT	&	92.34	&	40.19	&	26.63	& 86.83	&	36.44	&	28.31	&	82.56	&	20.17	&	11.59	\\
  \textbf{\nameabb} 	&	92.15	&	\textbf{51.86}	&	\textbf{42.67}	&		89.06	&	\textbf{58.98}	&	\textbf{53.45}	&	94.20	&	\textbf{66.28}	&	\textbf{56.94}	\\
\midrule																	
      \it CIFAR-10 &  $\mathcal{A}_{Clean}$ & $\bf\mathcal{A}_{PGD}$ & $\bf\mathcal{A}_{AA}$ & $\mathcal{A}_{Clean}$& $\bf\mathcal{A}_{PGD}$ & $\bf\mathcal{A}_{AA}$ & $\mathcal{A}_{Clean}$& $\bf\mathcal{A}_{PGD}$ & $\bf\mathcal{A}_{AA}$ \\
      \midrule

	AIT	&	24.49	&	{\textcolor{white}0}7.85	&	{\textcolor{white}0}2.68	&	47.98	&	12.69	&	{\textcolor{white}0}0.49	&	57.85	&	13.78	&	10.66	\\
  \textbf{\nameabb} 	&	74.27	&	\textbf{31.68}	&	\textbf{25.46}	&	83.33	&	\textbf{38.15}	&	\textbf{32.34}	&	88.54	&	\textbf{50.53}	&	\textbf{42.09}	\\
\midrule	
      \it CIFAR-100 &  $\mathcal{A}_{Clean}$ & $\bf\mathcal{A}_{PGD}$ & $\bf\mathcal{A}_{AA}$ & $\mathcal{A}_{Clean}$& $\bf\mathcal{A}_{PGD}$ & $\bf\mathcal{A}_{AA}$ & $\mathcal{A}_{Clean}$& $\bf\mathcal{A}_{PGD}$ & $\bf\mathcal{A}_{AA}$ \\
      \midrule

 AIT		&35.63 & {\textcolor{white}0}0.33& {\textcolor{white}0}0.01	&42.89	& {\textcolor{white}0}1.05	& {\textcolor{white}0}0.19	&31.84	&{\textcolor{white}0}0.79	&{\textcolor{white}0}0.00\\
  \textbf{\nameabb} 	&	43.57	&	\textbf{12.11}	&	\textbf{{\textcolor{white}0}7.60}	&	43.28	&	\textbf{15.42}	&	\textbf{11.32}	&	64.34	&	\textbf{24.92}	&	\textbf{17.14}	\\
     \bottomrule
    \end{tabular}
    }
   \vspace{-3mm}
\end{table}

\setlength{\tabcolsep}{3pt}
    \begin{table}[]    
    \centering
    \caption{Comparison of dataset diversification methods.}
        \vspace{1mm}
    \label{tab:ablation_diversification}
    
    \def\arraystretch{0.9}%
    \resizebox{\columnwidth}{!}
    {
    \begin{tabular}{lccccccc}
    \toprule
   \multirowcell{2}[-4pt][l]{Method} & \multicolumn{3}{c}{CIFAR-10 Accuracy}& \multicolumn{4}{c}{Diversity Metric} \\
   \cmidrule(lr){2-4} \cmidrule(lr){5-8}
        & $\mathcal{A}_{Clean}$& $\bf\mathcal{A}_{PGD}$ & $\bf\mathcal{A}_{AA}$  &\multicolumn{1}{c}{Recall $\uparrow$} & \multicolumn{1}{c}{Coverage $\uparrow$} & \multicolumn{1}{c}{NDB $\downarrow$} & \multicolumn{1}{c}{JSD $\downarrow$}\\
      \midrule

 Qimera	&	76.88	&	18.90	&	10.68	&	0.000	&	0.002	&	99	&	0.514	\\
RDSKD	&	10.00$^{\dagger}$	&	10.00	&	10.00	&	0.000	&	0.001	&	98	&	0.658	\\
IntraQ	&	13.77	&	36.13	&	12.46	&	0.308	&	0.087	&	\textbf{88}	&	0.275	\\
\midrule
 $\mathcal{L}_{Synth}$ &	91.46	&	43.66	&	36.34	& 0.535	&	0.101	&	91	&	0.253	\\
+ Mixup&  	90.61	&	48.16	&	36.43 	& 0.641	&	0.084	&	94	&	0.322	\\
+ Cutout  &	92.59	&	39.84	&	34.39 &0.535	&	0.034	&	95	&	0.443	\\
+ CutMix &	91.90	&	42.79	&	34.79 & \textbf{0.845}	&	0.084	&	93	&	0.328	\\

 \midrule
 \textbf{+ \mixabb} & 88.16	&	\textbf{50.13}	&	\textbf{41.40}	&0.830	&	\textbf{0.163}	&	\textbf{88}	&	\textbf{0.211}	\\

     \bottomrule
    \multicolumn{8}{r}{$^\dagger$Did not converge}\\
    \end{tabular}
    }
    \vspace{-3mm}
\end{table}

\subsection{In-depth Study on \name}
\label{sec:study}

We perform an in-depth study on \name, and analyze the efficacy of each component.
WRN-28-10 is mainly used, and more experiments can be found in the Appendix.

\begin{table}[]
  \centering
    \vspace{-3mm}
    \caption{Comparison of $\loss_{Train}$ on WRN-28-10. }
        \vspace{1mm}
    \label{tab:ablation_trainloss}
    \def\arraystretch{0.85}%
    \resizebox{.99\columnwidth}{!}
    {
    \begin{tabular}{lcccccc}
    \toprule
     &  \multicolumn{3}{c}{SVHN} & \multicolumn{3}{c}{CIFAR-10} \\
    \cmidrule(lr){2-4}\cmidrule(lr){5-7}
         $\loss_{Train}$ & $\mathcal{A}_{Clean}$ & $\bf\mathcal{A}_{PGD}$ & $\bf\mathcal{A}_{AA}$ & $\mathcal{A}_{Clean}$& $\bf\mathcal{A}_{PGD}$ & $\bf\mathcal{A}_{AA}$ \\
      \midrule
													
 STD &		93.71	&	69.32	&	62.58	&	81.63	&	48.03	&	38.94	\\
 TRADES&		94.12	&	69.10	&	61.75 &	79.61 &	45.86	&37.08	\\
 MART &		35.94	&	{\textcolor{white}0}2.55	&	{\textcolor{white}0}1.09	&	13.69	&	{\textcolor{white}0}6.74	&	{\textcolor{white}0}0.09	\\
 ARD &96.29 &	61.11	&52.56  &90.95	&36.61	& 31.16 \\
 RSLAD&96.03&	64.59&	57.04 &90.25 &	39.30	&31.16 \\
\midrule
\makecell[l]{$\bf{\lossdfs}$} &	94.87	&	\textbf{69.67}	&	\textbf{65.66} 	&	88.16	&	\textbf{50.13}	&	\textbf{41.40} 	\\
 
     \bottomrule
     \vspace{-6mm}
    \end{tabular}
    }
    \end{table}

\textbf{Dataset Diversification.} 
\label{sec:analysis_diversity}
\cref{tab:ablation_diversification} compares \mixabb with other existing methods for dataset diversification. 
On the one hand, we choose three data-free synthesis baselines for comparison: Qimera~\citep{qimera}, IntraQ~\citep{intraq}, and RDSKD~\citep{rdskd}.
We additionally test three image augmentation methods, Mixup~\citep{mixup}, Cutout~\citep{cutout}, and CutMix~\citep{cutmix} on top of direct sample optimization~\citep{deepinversion}, and for training we used $\lossdfs$ for all cases.
In terms of robustness, it is clear that \mixabb outperforms all other diversification methods in terms of $\mathcal{A}_{AA}$.

For further investigation, we measure several well-known diversity metrics often used in evaluating generative models: recall, coverage~\citep{gandiversitymetric}, number of statistically-different bins (NDB)~\citep{ndb}, and Jensen-Shannon divergence (JSD).
In almost all metrics, \mixabb shows the highest diversity, explaining its performance benefits.
Although CutMix~\citep{cutmix} shows slightly better recall than \mixabb, the difference is negligible and the coverage metric is generally perceived as a more exact measure of distributional diversity~\citep{gandiversitymetric}.
Measures on other datasets and models are in \cref{sec:supp:detailed}.

\textbf{Training Loss.} 
~\cref{tab:ablation_trainloss} compares our proposed train loss against state-of-the-art ones used in adversarial training.
STD~\citep{madry}, TRADES~\citep{trades}, and MART~\citep{mart} are from general adversarial training literature, while ARD~\citep{ard} and RSLAD~\citep{rslad} are from robust distillation methods.
\Mix was used for all cases for fair comparison.
Interestingly, MART provides almost no robustness in our problem. 
MART encourages learning from misclassified samples, 
which may lead the model to overfit on synthetic samples. 
On the other hand, $\lossdfs$ achieves the best results under both PGD-10 and AutoAttack in both datasets. 
The trend is consistent across different datasets and models, which we include in \cref{sec:supp:detailed}.

\setlength{\tabcolsep}{3pt}
\begin{table}[]
    \centering
    
    \caption{Ablation study of \name on CIFAR-10 dataset. \HY{I removed PGD acc for the sake of space}}
        \vspace{1mm}
    \label{tab:ablation}
    
    \def\arraystretch{0.8}%
    \resizebox{\columnwidth}{!}
    {
    \begin{tabular}{cccccl}
    \toprule
      Model  & $\lossdfs$ & \mixabb & \gradmask  & $\mathcal{A}_{Clean}$ & $\mathcal{A}_{AA}$ \\
      \midrule

\multirow{4}{*}{ResNet-20}& \xmark & \xmark & \xmark & 86.42  & {\color{white}0}2.03 \\
& \cmark & \xmark & \xmark & 82.58  & 14.61 (+12.58)\\
& \cmark & \cmark & \xmark & 77.83	&	19.09 (+17.06)	\\
\cmidrule(lr){2-6}
& \cmark & \cmark & \cmark & 74.79	&	\textbf{22.65 (+20.62)}  \\
\midrule

\multirow{4}{*}{ResNet-56}& \xmark & \xmark & \xmark &  78.22  & 24.34\\
& \cmark & \xmark & \xmark &  83.72  & 27.42  (+3.08)\\
& \cmark & \cmark & \xmark & 	83.67 &	27.69 (+3.35)\\
\cmidrule(lr){2-6}
& \cmark & \cmark & \cmark & 81.30	&	\textbf{30.51 (+6.17)}\\
\midrule

\multirow{4}{*}{WRN-28-10}& \xmark & \xmark & \xmark & 80.29 & 37.96\\
& \cmark & \xmark & \xmark & 91.46 & 36.34  (-1.62)\\
& \cmark & \cmark & \xmark & 88.16		&	41.40 (+3.44)	\\
\cmidrule(lr){2-6}
& \cmark & \cmark & \cmark & 86.74	&	\textbf{43.73 (+5.77)} \\

     \bottomrule
     \vspace{-5mm}
    \end{tabular}
    }
\end{table}

\textbf{Ablation Study.} \cref{tab:ablation} shows an ablation study of \name. 
The baseline where none of our methods are applied denotes using the exact same set of synthesis loss functions without \mixabb, and adversarial training is done via TRADES. 
Across all models, there is a consistent gain on $\mathcal{A}_{AA}$.
Applying $\lossdfs$, seems to slightly degrade $\mathcal{A}_{AA}$ on WRN-28-10, but when combined with the other techniques, it results in better performance as shown in \cref{tab:ablation_trainloss}.
This is due to $\lossdfs$ effectively reducing the gap between relatively weaker and stronger attacks.
\gradmask adds a similar improvement, 
resulting in 6.17\%p to 20.62\%p gain altogether under AutoAttack.

\setlength{\tabcolsep}{3pt}
\begin{table}[]
  \centering
    \vspace{-2mm}
    \caption{Evaluation on gradient-free and adaptive attacks.}
            \vspace{1mm}
    \label{tab:gradient_free}
    \def\arraystretch{0.95}%
    \resizebox{\columnwidth}{!}
    {
    \begin{tabular}{lccccccc}
    \toprule
     & & \multicolumn{3}{c}{Gradient-free} & \multicolumn{3}{c}{Adaptive} \\
    \cmidrule(lr){3-5}\cmidrule(lr){6-8}
                \it SVHN & Clean &  GenAtt. & BDRY & SPSA & AA & $A^3$ & Automated \\

      \midrule

RN20 	&	91.83	&	90.28	&	91.77	&	60.99	&	47.55	&	47.25	&	47.52	\\
RN56 	&	88.66	&	87.05	&	88.60	&	67.65	&	57.54	&	57.28	&	57.54	\\
WRN28 	&	94.14	&	93.44	&	91.77	&	70.01	&	62.66	&	62.97	&	62.75	\\

\midrule
        \it CIFAR-10 & Clean &  GenAtt. & BDRY & SPSA & AA & $A^3$ & Automated \\
      \midrule

RN20 	&	74.79	&	72.92	&	74.73	&	34.58	&	22.65	&	22.62	&	22.65	\\
RN56 	&	81.30	&	79.10	&	81.27	&	43.65	&	30.51	&	30.48	&	30.50	\\
WRN28 	&	86.74	&	85.15	&	86.66	&	55.12	&	43.73	&	43.66	&	43.72	\\
 
     \bottomrule
     \vspace{-2mm}
    \end{tabular}
    }
    \end{table}

\setlength{\tabcolsep}{3pt}
\begin{table}[]
  \centering
            \vspace{-4mm}
    \caption{Performance comparison using different iterations and unbounded attack.}
        \vspace{1mm}
    \label{tab:obfuscated}
    \def\arraystretch{0.99}%
    \resizebox{\columnwidth}{!}
    {
    \begin{tabular}{lccccccc}
    \toprule
     & & Single & \multicolumn{4}{c}{Iterative Attack} & Unbound\\
    \cmidrule(lr){3-3} \cmidrule(lr){4-7} \cmidrule(lr){8-8}
                 \it SVHN &Clean & 
                $PGD_1$ & $PGD_2$ & $PGD_5$ & $PGD_{10}$ & $PGD_{100}$ & $PGD_{1000}^\infty$ \\

      \midrule
													
RN20	&	91.83	&	85.86	&	78.02	&	57.57	&	54.82	&	53.81	&	0.00	\\
RN56	&	88.66	&	83.69	&	77.99	&	64.01	&	62.05	&	61.06	&	0.00	\\
WRN28	&	94.14	&	90.74	&	85.77	&	71.37	&	69.60	&	68.76	&	0.00	\\

\midrule
                         \it CIFAR-10 &Clean & 
                        $PGD_1$ & $PGD_2$ & $PGD_5$ & $PGD_{10}$ & $PGD_{100}$ & $PGD_{1000}^\infty$ \\

      \midrule

RN20	&	74.79	&	63.34	&	52.38	&	31.05	&	29.29	&	28.62	&	0.00	\\
RN56	&	81.30	&	71.47	&	59.98	&	38.02	&	35.55	&	34.30	&	0.00	\\
WRN28	&	86.74	&	80.16	&	72.20	&	53.26	&	51.13	&	50.32	&	0.00	\\
 
     \bottomrule
    \end{tabular}
    }
    \end{table}

\subsection{Obfuscated Gradients}

Gradient obfuscation~\cite{obfuscated} refers to a case where a model either intentionally or unintentionally masks the gradient path that is necessary for optimization-based attacks.
While these models seem robust against optimization attacks, they are easily attacked by gradient-free methods, demonstrating a false sense of robustness. 
Thus we conduct a series of experiments to validate that models trained using \name do not fall into such case.

First, we use 3 gradient-free attacks: GenAttack~\cite{genattack}, Boundary attack~\cite{boundary}, SPSA~\cite{spsa}, and 2 adaptive attacks: $A^3$~\cite{adaptiveauto} and Automated~\cite{automated} for evaluation.
In \cref{tab:gradient_free}, gradient-free attacks all fail to show higher attack success rate than optimization-based attacks (PGD and AutoAttack), meaning our method does not leverage gradient masking to circumvent optimization-based attacks. 
Also, in adaptive attacks, both $A^3$ and Automated show less than 1\% degradation to the AutoAttack accuracy.
Such consistent robustness across stronger adaptive attacks ensures that the evaluated robustness of \name is not overestimated.

To further eliminate the possibility of gradient obfuscation, we observe \name under increasing number of PGD iterations, including an unbounded attack ($\eps=\infty$), shown in \cref{tab:obfuscated}.
According to \citet{obfuscated}, common signs of obfuscated gradients include single-step attack performing better than iterative attacks, unbounded attacks not reaching 100\% attack success rate, and black-box attacks performing better than white-box attacks. 
In \cref{tab:gradient_free} and \cref{tab:obfuscated}, we observe that \name does not fall into any of these cases.
Rather, we observe that increasing the number of iterations leads to better attack performance, and unbounded attacks reach 100\% success rate in all cases.
Note that GenAttack and Boundary attack are black-box attacks and their performance do not exceed other white-box attacks.

\section{Related Work}
\textbf{Adversarial Defense.} 
Existing defense methods train robust models by finetuning with adversarially perturbed data. 
Popular approaches include designing loss functions as variants of STD~\citep{madry}, such as TRADES~\citep{trades} or MART~\citep{mart}. 
While there exists other techniques for achieving robustness such as random smoothing~\citep{smoothing}, adversarial purification~\cite{diffpure}, or distillation~\cite{rslad, ard}, adversarial training and its variants are shown to be effective in most cases. 
A recent trend is to enhance the performance of adverarial training by importing extra data from other datasets~\citep{unlabelddata, rebuffi2021data}, or generated under the supervision of real data~\citep{rebuffi2021data, proxy}. 
However, such rich datasets are not easy to obtain in practice, sometimes none available in our setting.

\textbf{Data-free Learning.} 
Training or fine-tuning an existing model in absence of data has been studied to some degree. 
However, most are related to, or confined to only compression tasks, some of which are knowledge distillation~\citep{dfad, dfkd}, pruning~\citep{dfpruning}, and quantization~\citep{dfq, zeroq, gdfq, zaq, qimera, ait, autorecon}.
A concurrent work DFARD~\citep{dfard} sets a similar but different problem where a robust model already exists, and the objective is to distill it to a lighter network.
Without the existence of a robust model, the effectiveness of DFARD is significantly reduced.

\textbf{Gradient Refining Techniques.} 
Adjusting gradients is a popular approach for diverse objectives.
\citet{yu2020gradient} directly projects gradients with opposing directionality to dominant task gradients before model update. 
\citet{liu2021conflict} selectively uses gradients that can best aid the worst performing task, and \citet{fernando2022mitigating} estimates unbiased approximations of gradients to ensure convergence. 
\citet{eshratifar2018gradient} also utilizes gradients to maximize generalization ability to unseen data. 
Similar to ours, \citet{mansilla2021domain} and \citet{tenison2022gradient} update the model based on sign agreement of gradients across domains or clients. 
\citet{shi2022gradient}, \citet{wang2023sharpness}, and \citet{dandi2022implicit} maximize gradient inner product between different domains or loss terms. 
While being effective, none target adversarial robustness, especially in data-free settings.

\textbf{Loss Surface Smoothness and Generalization.}
Smoothness of loss surface is often associated with the model's generalization ability.
\citet{keskar2016large}, \citet{jiang2019fantastic}, and \citet{izmailov2018averaging} have shown correlation between smoothness and generalization, using quantitative measures such as eigenvalues of the loss curvature. 
Sharpness-aware minimization methods~\cite{sam, wang2023sharpness} leverage this knowledge to reduce the train-test generalization gap by explicitly regularizing the sharpness of the loss surface. 
Other methods \citep{terjek2019adversarial, qin2019adversarial, moosavi2019robustness} also implicitly regularize the model towards smoother minima.
Notably, in the field where the generalization gap is more severe (e.g. domain generalization or federated learning), methods penalize gradients that hinder the smoothness of the landscape~\cite{zhao2022penalizing}, or regularize the model to have flatter landscapes across different domains of the dataset \citep{cha2021swad, caldarola2022improving, phan2022improving}. 

\section{Conclusion}
In this work, we study the problem of learning data-free adversarial robustness under the absence of real data. 
We propose \name, an effective method for instilling robustness to a given model using synthetic data. 
We approach the problem from perspectives of generating diverse synthetic datasets and training with flatter loss surfaces.
Further, we propose a new training loss function most suitable for our problem, and provide analysis on its effectiveness. 
Experimental results show that \name significantly outperforms baseline approaches, demonstrating that it successfully achieves robustness without the original datasets.

 \section*{Acknowledgements}
 This work was supported by 
the National Research Foundation of Korea (NRF) grant funded by the Korea government (MSIT) (2022R1C1C1011307),
and Institute of Information \& communications Technology Planning \& Evaluation (IITP) (
RS-2023-00256081, 
RS-2024-00347394, 
RS-2022-II220959 (10\%), 
RS-2021-II211343 (10\%, SNU AI), 
RS-2021-II212068 (10\%, AI Innov. Hub)) 
grant funded by the Korea government (MSIT).

\section*{Impact Statement}
\label{sec:supp:privacy}
 As shown in \cref{sec:supp:samples}, the generated samples are not very human-recognizable, and being so does not necessarily lead to better performance of the models.
From these facts, we believe our synthetic input generation does not cause privacy invasion that might have existed from the original training dataset.
However, there is still a possibility where the generated samples could affect privacy concerns, such as membership inference attacks~\citep{mia} or model stealing~\citep{steal}. 
For example, an attacker might compare the image-level or feature-level similarity of some test samples with the synthetically generated samples to find out whether the test sample is part of the training set or not. 
We believe further investigation is needed on such side-effects, which we leave as a future work.

\bibliography{icml2024}

\begin{thebibliography}{95}
\providecommand{\natexlab}[1]{#1}
\providecommand{\url}[1]{\texttt{#1}}
\expandafter\ifx\csname urlstyle\endcsname\relax
  \providecommand{\doi}[1]{doi: #1}\else
  \providecommand{\doi}{doi: \begingroup \urlstyle{rm}\Url}\fi

\bibitem[pyt()]{pytorchcv}
{Computer vision models on PyTorch}.
\newblock URL \url{https://pypi.org/project/pytorchcv/}.

\bibitem[Alzantot et~al.(2019)Alzantot, Sharma, Chakraborty, Zhang, Hsieh, and Srivastava]{genattack}
Alzantot, M., Sharma, Y., Chakraborty, S., Zhang, H., Hsieh, C.-J., and Srivastava, M.~B.
\newblock Genattack: Practical black-box attacks with gradient-free optimization.
\newblock In \emph{Proceedings of the genetic and evolutionary computation conference}, 2019.

\bibitem[Athalye et~al.(2018)Athalye, Carlini, and Wagner]{obfuscated}
Athalye, A., Carlini, N., and Wagner, D.
\newblock Obfuscated gradients give a false sense of security: Circumventing defenses to adversarial examples.
\newblock In \emph{International Conference on Machine Learning}, 2018.

\bibitem[Brendel et~al.(2018)Brendel, Rauber, and Bethge]{boundary}
Brendel, W., Rauber, J., and Bethge, M.
\newblock Decision-based adversarial attacks: Reliable attacks against black-box machine learning models.
\newblock In \emph{Proceedings of International Conference on Learning Representations}, 2018.

\bibitem[Cai et~al.(2020)Cai, Yao, Dong, Gholami, Mahoney, and Keutzer]{zeroq}
Cai, Y., Yao, Z., Dong, Z., Gholami, A., Mahoney, M.~W., and Keutzer, K.
\newblock {ZeroQ: A novel zero shot quantization framework}.
\newblock In \emph{Proceedings of the IEEE/CVF Conference on Computer Vision and Pattern Recognition}, 2020.

\bibitem[Caldarola et~al.(2022)Caldarola, Caputo, and Ciccone]{caldarola2022improving}
Caldarola, D., Caputo, B., and Ciccone, M.
\newblock Improving generalization in federated learning by seeking flat minima.
\newblock In \emph{European Conference on Computer Vision}, 2022.

\bibitem[Carmon et~al.(2019)Carmon, Raghunathan, Schmidt, Duchi, and Liang]{unlabelddata}
Carmon, Y., Raghunathan, A., Schmidt, L., Duchi, J.~C., and Liang, P.~S.
\newblock Unlabeled data improves adversarial robustness.
\newblock In \emph{Advances in Neural Information Processing Systems}, 2019.

\bibitem[Cha et~al.(2021)Cha, Chun, Lee, Cho, Park, Lee, and Park]{cha2021swad}
Cha, J., Chun, S., Lee, K., Cho, H.-C., Park, S., Lee, Y., and Park, S.
\newblock Swad: Domain generalization by seeking flat minima.
\newblock \emph{Advances in Neural Information Processing Systems}, 2021.

\bibitem[Choi et~al.(2021)Choi, Hong, Park, Kim, and Lee]{qimera}
Choi, K., Hong, D., Park, N., Kim, Y., and Lee, J.
\newblock {Qimera: Data-free quantization with synthetic boundary supporting samples}.
\newblock In \emph{Advances in Neural Information Processing Systems}, 2021.

\bibitem[Choi et~al.(2022)Choi, Lee, Hong, Yu, Park, Kim, and Lee]{ait}
Choi, K., Lee, H., Hong, D., Yu, J., Park, N., Kim, Y., and Lee, J.
\newblock It's all in the teacher: Zero-shot quantization brought closer to the teacher.
\newblock In \emph{Proceedings of the IEEE/CVF Conference on Computer Vision and Pattern Recognition}, 2022.

\bibitem[Croce \& Hein(2020)Croce and Hein]{autoattack}
Croce, F. and Hein, M.
\newblock Reliable evaluation of adversarial robustness with an ensemble of diverse parameter-free attacks.
\newblock In \emph{International Conference on Machine Learning}, 2020.

\bibitem[Croce et~al.(2020)Croce, Andriushchenko, Sehwag, Debenedetti, Flammarion, Chiang, Mittal, and Hein]{robustbench}
Croce, F., Andriushchenko, M., Sehwag, V., Debenedetti, E., Flammarion, N., Chiang, M., Mittal, P., and Hein, M.
\newblock Robustbench: a standardized adversarial robustness benchmark.
\newblock \emph{arXiv preprint arXiv:2010.09670}, 2020.

\bibitem[Dandi et~al.(2022)Dandi, Barba, and Jaggi]{dandi2022implicit}
Dandi, Y., Barba, L., and Jaggi, M.
\newblock Implicit gradient alignment in distributed and federated learning.
\newblock In \emph{Proceedings of the AAAI Conference on Artificial Intelligence}, 2022.

\bibitem[DeVries \& Taylor(2017)DeVries and Taylor]{cutout}
DeVries, T. and Taylor, G.~W.
\newblock Improved regularization of convolutional neural networks with cutout.
\newblock \emph{arXiv preprint arXiv:1708.04552}, 2017.

\bibitem[Eshratifar et~al.(2018)Eshratifar, Eigen, and Pedram]{eshratifar2018gradient}
Eshratifar, A.~E., Eigen, D., and Pedram, M.
\newblock Gradient agreement as an optimization objective for meta-learning.
\newblock \emph{arXiv preprint arXiv:1810.08178}, 2018.

\bibitem[Fang et~al.(2019)Fang, Song, Shen, Wang, Chen, and Song]{dfad}
Fang, G., Song, J., Shen, C., Wang, X., Chen, D., and Song, M.
\newblock Data-free adversarial distillation.
\newblock \emph{arXiv preprint arXiv:1912.11006}, 2019.

\bibitem[Fernando et~al.(2022)Fernando, Shen, Liu, Chaudhury, Murugesan, and Chen]{fernando2022mitigating}
Fernando, H.~D., Shen, H., Liu, M., Chaudhury, S., Murugesan, K., and Chen, T.
\newblock Mitigating gradient bias in multi-objective learning: A provably convergent approach.
\newblock In \emph{International Conference on Learning Representations}, 2022.

\bibitem[Foret et~al.(2020)Foret, Kleiner, Mobahi, and Neyshabur]{sam}
Foret, P., Kleiner, A., Mobahi, H., and Neyshabur, B.
\newblock Sharpness-aware minimization for efficiently improving generalization.
\newblock In \emph{International Conference on Learning Representations}, 2020.

\bibitem[Ghiasi et~al.(2022)Ghiasi, Kazemi, Reich, Zhu, Goldblum, and Goldstein]{plugin}
Ghiasi, A., Kazemi, H., Reich, S., Zhu, C., Goldblum, M., and Goldstein, T.
\newblock Plug-in inversion: Model-agnostic inversion for vision with data augmentations.
\newblock In \emph{International Conference on Machine Learning}, 2022.

\bibitem[Goldblum et~al.(2020)Goldblum, Fowl, Feizi, and Goldstein]{ard}
Goldblum, M., Fowl, L., Feizi, S., and Goldstein, T.
\newblock Adversarially robust distillation.
\newblock In \emph{Proceedings of the AAAI Conference on Artificial Intelligence}, 2020.

\bibitem[Goodfellow et~al.(2014)Goodfellow, Pouget-Abadie, Mirza, Xu, Warde-Farley, Ozair, Courville, and Bengio]{gan}
Goodfellow, I., Pouget-Abadie, J., Mirza, M., Xu, B., Warde-Farley, D., Ozair, S., Courville, A., and Bengio, Y.
\newblock Generative adversarial nets.
\newblock In \emph{Advances in Neural Information Processing Systems}, 2014.

\bibitem[Goodfellow et~al.(2015)Goodfellow, Shlens, and Szegedy]{goodfellow2014explaining}
Goodfellow, I.~J., Shlens, J., and Szegedy, C.
\newblock Explaining and harnessing adversarial examples.
\newblock In \emph{International Conference on Learning Representations}, 2015.

\bibitem[Han et~al.(2021)Han, Park, Wang, and Liu]{rdskd}
Han, P., Park, J., Wang, S., and Liu, Y.
\newblock Robustness and diversity seeking data-free knowledge distillation.
\newblock In \emph{IEEE International Conference on Acoustics, Speech and Signal Processing}, 2021.

\bibitem[Hathaliya \& Tanwar(2020)Hathaliya and Tanwar]{hathaliya2020exhaustive}
Hathaliya, J.~J. and Tanwar, S.
\newblock An exhaustive survey on security and privacy issues in healthcare 4.0.
\newblock \emph{Computer Communications}, 2020.

\bibitem[He et~al.(2016)He, Zhang, Ren, and Sun]{resnet}
He, K., Zhang, X., Ren, S., and Sun, J.
\newblock Deep residual learning for image recognition.
\newblock In \emph{Proceedings of the IEEE/CVF Conference on Computer Vision and Pattern Recognition}, 2016.

\bibitem[Huang et~al.(2020)Huang, Zhang, and Zhang]{mixup}
Huang, L., Zhang, C., and Zhang, H.
\newblock Self-adaptive training: Beyond empirical risk minimization.
\newblock In \emph{Advances in Neural Information Processing Systems}, 2020.

\bibitem[Huang et~al.(2022)Huang, Lu, Deb, and Boddeti]{archview}
Huang, S., Lu, Z., Deb, K., and Boddeti, V.~N.
\newblock Revisiting residual networks for adversarial robustness: An architectural perspective.
\newblock \emph{arXiv preprint arXiv:2212.11005}, 2022.

\bibitem[Izmailov et~al.(2018)Izmailov, Podoprikhin, Garipov, Vetrov, and Wilson]{izmailov2018averaging}
Izmailov, P., Podoprikhin, D., Garipov, T., Vetrov, D., and Wilson, A.~G.
\newblock Averaging weights leads to wider optima and better generalization.
\newblock In \emph{34th Conference on Uncertainty in Artificial Intelligence}, 2018.

\bibitem[Jiang et~al.(2019)Jiang, Neyshabur, Mobahi, Krishnan, and Bengio]{jiang2019fantastic}
Jiang, Y., Neyshabur, B., Mobahi, H., Krishnan, D., and Bengio, S.
\newblock Fantastic generalization measures and where to find them.
\newblock In \emph{International Conference on Learning Representations}, 2019.

\bibitem[Keskar et~al.(2017)Keskar, Mudigere, Nocedal, Smelyanskiy, and Tang]{keskar2016large}
Keskar, N.~S., Mudigere, D., Nocedal, J., Smelyanskiy, M., and Tang, P. T.~P.
\newblock On large-batch training for deep learning: Generalization gap and sharp minima.
\newblock In \emph{International Conference on Learning Representations}, 2017.

\bibitem[Khim \& Loh(2018)Khim and Loh]{khim2018adversarial}
Khim, J. and Loh, P.-L.
\newblock Adversarial risk bounds via function transformation.
\newblock \emph{arXiv preprint arXiv:1810.09519}, 2018.

\bibitem[Krizhevsky et~al.(2009)Krizhevsky, Hinton, et~al.]{cifar}
Krizhevsky, A., Hinton, G., et~al.
\newblock Learning multiple layers of features from tiny images, 2009.
\newblock URL \url{http://www.cs.utoronto.ca/~kriz/learning-features-2009-TR.pdf}.

\bibitem[Kurmi et~al.(2021)Kurmi, Subramanian, and Namboodiri]{kurmi2021domain}
Kurmi, V.~K., Subramanian, V.~K., and Namboodiri, V.~P.
\newblock Domain impression: A source data free domain adaptation method.
\newblock In \emph{Proceedings of the IEEE/CVF winter conference on applications of computer vision}, 2021.

\bibitem[Lee et~al.(2019)Lee, Edwards, Molloy, and Su]{steal}
Lee, T., Edwards, B., Molloy, I., and Su, D.
\newblock Defending against neural network model stealing attacks using deceptive perturbations.
\newblock In \emph{IEEE Security and Privacy Workshops}, 2019.

\bibitem[Li et~al.(2018)Li, Xu, Taylor, Studer, and Goldstein]{li2018visualizing}
Li, H., Xu, Z., Taylor, G., Studer, C., and Goldstein, T.
\newblock Visualizing the loss landscape of neural nets.
\newblock In \emph{Advances in Neural Information Processing Systems}, 2018.

\bibitem[Li \& Spratling(2022)Li and Spratling]{aug_alone}
Li, L. and Spratling, M.~W.
\newblock Data augmentation alone can improve adversarial training.
\newblock In \emph{The International Conference on Learning Representations}, 2022.

\bibitem[Li et~al.(2022)Li, Ma, Chen, Xiao, and Gu]{PSAQ-VIT}
Li, Z., Ma, L., Chen, M., Xiao, J., and Gu, Q.
\newblock Patch similarity aware data-free quantization for vision transformers.
\newblock In \emph{European Conference on Computer Vision}, 2022.

\bibitem[Liu et~al.(2021{\natexlab{a}})Liu, Ding, Shaham, Rahayu, Farokhi, and Lin]{liu2021machine}
Liu, B., Ding, M., Shaham, S., Rahayu, W., Farokhi, F., and Lin, Z.
\newblock When machine learning meets privacy: A survey and outlook.
\newblock \emph{ACM Computing Surveys}, 2021{\natexlab{a}}.

\bibitem[Liu et~al.(2021{\natexlab{b}})Liu, Liu, Jin, Stone, and Liu]{liu2021conflict}
Liu, B., Liu, X., Jin, X., Stone, P., and Liu, Q.
\newblock Conflict-averse gradient descent for multi-task learning.
\newblock In \emph{Advances in Neural Information Processing Systems}, 2021{\natexlab{b}}.

\bibitem[Liu et~al.(2020)Liu, Salzmann, Lin, Tomioka, and S{\"u}sstrunk]{liu2020loss}
Liu, C., Salzmann, M., Lin, T., Tomioka, R., and S{\"u}sstrunk, S.
\newblock On the loss landscape of adversarial training: Identifying challenges and how to overcome them.
\newblock In \emph{Advances in Neural Information Processing Systems}, 2020.

\bibitem[Liu et~al.(2021{\natexlab{c}})Liu, Zhang, and Wang]{zaq}
Liu, Y., Zhang, W., and Wang, J.
\newblock Zero-shot adversarial quantization.
\newblock In \emph{Proceedings of the IEEE/CVF Conference on Computer Vision and Pattern Recognition}, 2021{\natexlab{c}}.

\bibitem[Liu et~al.(2022)Liu, Cheng, Gao, Liu, Zhang, and Song]{adaptiveauto}
Liu, Y., Cheng, Y., Gao, L., Liu, X., Zhang, Q., and Song, J.
\newblock Practical evaluation of adversarial robustness via adaptive auto attack.
\newblock In \emph{Proceedings of the IEEE/CVF Conference on Computer Vision and Pattern Recognition}, 2022.

\bibitem[Lopes et~al.(2017)Lopes, Fenu, and Starner]{dfkd}
Lopes, R.~G., Fenu, S., and Starner, T.
\newblock Data-free knowledge distillation for deep neural networks.
\newblock In \emph{Advances in Neural Information Processing Systems Workshops}, 2017.

\bibitem[Madry et~al.(2018)Madry, Makelov, Schmidt, Tsipras, and Vladu]{madry}
Madry, A., Makelov, A., Schmidt, L., Tsipras, D., and Vladu, A.
\newblock Towards deep learning models resistant to adversarial attacks.
\newblock In \emph{International Conference on Learning Representations}, 2018.

\bibitem[Mansilla et~al.(2021)Mansilla, Echeveste, Milone, and Ferrante]{mansilla2021domain}
Mansilla, L., Echeveste, R., Milone, D.~H., and Ferrante, E.
\newblock Domain generalization via gradient surgery.
\newblock In \emph{Proceedings of the IEEE/CVF International Conference on Computer Vision}, 2021.

\bibitem[Moosavi-Dezfooli et~al.(2019)Moosavi-Dezfooli, Fawzi, Uesato, and Frossard]{moosavi2019robustness}
Moosavi-Dezfooli, S.-M., Fawzi, A., Uesato, J., and Frossard, P.
\newblock Robustness via curvature regularization, and vice versa.
\newblock In \emph{Proceedings of the IEEE/CVF Conference on Computer Vision and Pattern Recognition}, 2019.

\bibitem[Naeem et~al.(2020)Naeem, Oh, Uh, Choi, and Yoo]{gandiversitymetric}
Naeem, M.~F., Oh, S.~J., Uh, Y., Choi, Y., and Yoo, J.
\newblock Reliable fidelity and diversity metrics for generative models.
\newblock In \emph{International Conference on Machine Learning}, 2020.

\bibitem[Nagel et~al.(2019)Nagel, Baalen, Blankevoort, and Welling]{dfq}
Nagel, M., Baalen, M.~v., Blankevoort, T., and Welling, M.
\newblock Data-free quantization through weight equalization and bias correction.
\newblock In \emph{Proceedings of the IEEE International Conference on Computer Vision}, 2019.

\bibitem[Nayak et~al.(2022)Nayak, Rawal, and Chakraborty]{dad}
Nayak, G.~K., Rawal, R., and Chakraborty, A.
\newblock Dad: Data-free adversarial defense at test time.
\newblock In \emph{Proceedings of the IEEE/CVF Winter Conference on Applications of Computer Vision}, 2022.

\bibitem[Netzer et~al.(2011)Netzer, Wang, Coates, Bissacco, Wu, and Ng]{svhn}
Netzer, Y., Wang, T., Coates, A., Bissacco, A., Wu, B., and Ng, A.~Y.
\newblock Reading digits in natural images with unsupervised feature learning.
\newblock In \emph{Advances in Neural Information Processing Systems Workshops}, 2011.

\bibitem[Nie et~al.(2022)Nie, Guo, Huang, Xiao, Vahdat, and Anandkumar]{diffpure}
Nie, W., Guo, B., Huang, Y., Xiao, C., Vahdat, A., and Anandkumar, A.
\newblock Diffusion models for adversarial purification.
\newblock In \emph{International Conference on Machine Learning}, 2022.

\bibitem[Odena et~al.(2017)Odena, Olah, and Shlens]{acgan}
Odena, A., Olah, C., and Shlens, J.
\newblock {Conditional image synthesis with auxiliary classifier GANs}.
\newblock In \emph{International Conference on Machine Learning}, 2017.

\bibitem[Patashnik et~al.(2021)Patashnik, Wu, Shechtman, Cohen-Or, and Lischinski]{clip}
Patashnik, O., Wu, Z., Shechtman, E., Cohen-Or, D., and Lischinski, D.
\newblock Styleclip: Text-driven manipulation of stylegan imagery.
\newblock In \emph{Proceedings of the IEEE International Conference on Computer Vision}, 2021.

\bibitem[P{\'e}rez et~al.(2021)P{\'e}rez, Alfarra, Jeanneret, Rueda, Thabet, Ghanem, and Arbel{\'a}ez]{TTE}
P{\'e}rez, J.~C., Alfarra, M., Jeanneret, G., Rueda, L., Thabet, A., Ghanem, B., and Arbel{\'a}ez, P.
\newblock Enhancing adversarial robustness via test-time transformation ensembling.
\newblock In \emph{Proceedings of International Conference on Computer Vision}, 2021.

\bibitem[Phan et~al.(2022)Phan, Tran, Tran, Ho, Phung, and Le]{phan2022improving}
Phan, H., Tran, L., Tran, N.~N., Ho, N., Phung, D., and Le, T.
\newblock Improving multi-task learning via seeking task-based flat regions.
\newblock \emph{arXiv preprint arXiv:2211.13723}, 2022.

\bibitem[Qin et~al.(2019)Qin, Martens, Gowal, Krishnan, Dvijotham, Fawzi, De, Stanforth, and Kohli]{qin2019adversarial}
Qin, C., Martens, J., Gowal, S., Krishnan, D., Dvijotham, K., Fawzi, A., De, S., Stanforth, R., and Kohli, P.
\newblock Adversarial robustness through local linearization.
\newblock \emph{Advances in Neural Information Processing Systems}, 2019.

\bibitem[Ramesh et~al.(2021)Ramesh, Pavlov, Goh, Gray, Voss, Radford, Chen, and Sutskever]{dalle}
Ramesh, A., Pavlov, M., Goh, G., Gray, S., Voss, C., Radford, A., Chen, M., and Sutskever, I.
\newblock Zero-shot text-to-image generation.
\newblock In \emph{International Conference on Machine Learning}, 2021.

\bibitem[Rebuffi et~al.(2021)Rebuffi, Gowal, Calian, Stimberg, Wiles, and Mann]{rebuffi2021data}
Rebuffi, S.-A., Gowal, S., Calian, D.~A., Stimberg, F., Wiles, O., and Mann, T.~A.
\newblock Data augmentation can improve robustness.
\newblock In \emph{Advances in Neural Information Processing Systems}, 2021.

\bibitem[Rice et~al.(2020)Rice, Wong, and Kolter]{robustoverfitting}
Rice, L., Wong, E., and Kolter, Z.
\newblock Overfitting in adversarially robust deep learning.
\newblock In \emph{Proceedings of the International Conference on Machine Learning}, 2020.

\bibitem[Richardson \& Weiss(2018)Richardson and Weiss]{ndb}
Richardson, E. and Weiss, Y.
\newblock On gans and gmms.
\newblock In \emph{Advances in Neural Information Processing Systems}, 2018.

\bibitem[Sabour et~al.(2016)Sabour, Cao, Faghri, and Fleet]{latentattack}
Sabour, S., Cao, Y., Faghri, F., and Fleet, D.~J.
\newblock Adversarial manipulation of deep representations.
\newblock \emph{International Conference on Learning Representations}, 2016.

\bibitem[Saharia et~al.(2022)Saharia, Chan, Saxena, Li, Whang, Denton, Ghasemipour, Gontijo~Lopes, Karagol~Ayan, Salimans, et~al.]{imagen}
Saharia, C., Chan, W., Saxena, S., Li, L., Whang, J., Denton, E.~L., Ghasemipour, K., Gontijo~Lopes, R., Karagol~Ayan, B., Salimans, T., et~al.
\newblock Photorealistic text-to-image diffusion models with deep language understanding.
\newblock In \emph{Advances in Neural Information Processing Systems}, 2022.

\bibitem[Schmidt et~al.(2018)Schmidt, Santurkar, Tsipras, Talwar, and Madry]{schmidt2018adversarially}
Schmidt, L., Santurkar, S., Tsipras, D., Talwar, K., and Madry, A.
\newblock Adversarially robust generalization requires more data.
\newblock In \emph{Advances in Neural Information Processing Systems}, 2018.

\bibitem[Sehwag et~al.(2021)Sehwag, Mahloujifar, Handina, Dai, Xiang, Chiang, and Mittal]{proxy}
Sehwag, V., Mahloujifar, S., Handina, T., Dai, S., Xiang, C., Chiang, M., and Mittal, P.
\newblock Robust learning meets generative models: Can proxy distributions improve adversarial robustness?
\newblock In \emph{International Conference on Learning Representations}, 2021.

\bibitem[Shi et~al.(2022)Shi, Seely, Torr, N, Hannun, Usunier, and Synnaeve]{shi2022gradient}
Shi, Y., Seely, J., Torr, P., N, S., Hannun, A., Usunier, N., and Synnaeve, G.
\newblock Gradient matching for domain generalization.
\newblock In \emph{International Conference on Learning Representations}, 2022.

\bibitem[Shokri et~al.(2017)Shokri, Stronati, Song, and Shmatikov]{mia}
Shokri, R., Stronati, M., Song, C., and Shmatikov, V.
\newblock Membership inference attacks against machine learning models.
\newblock In \emph{IEEE Symposium on Security and Privacy}, 2017.

\bibitem[Srinivas \& Babu(2015)Srinivas and Babu]{dfpruning}
Srinivas, S. and Babu, R.~V.
\newblock Data-free parameter pruning for deep neural networks.
\newblock In \emph{British Machine Vision Conference}, 2015.

\bibitem[Stutz et~al.(2021)Stutz, Hein, and Schiele]{flatminima}
Stutz, D., Hein, M., and Schiele, B.
\newblock Relating adversarially robust generalization to flat minima.
\newblock In \emph{Proceedings of the IEEE/CVF International Conference on Computer Vision}, 2021.

\bibitem[Szegedy et~al.(2014)Szegedy, Zaremba, Sutskever, Bruna, Erhan, Goodfellow, and Fergus]{szegedy2013intriguing}
Szegedy, C., Zaremba, W., Sutskever, I., Bruna, J., Erhan, D., Goodfellow, I., and Fergus, R.
\newblock Intriguing properties of neural networks.
\newblock In \emph{International Conference on Learning Representations}, 2014.

\bibitem[Szegedy et~al.(2016)Szegedy, Vanhoucke, Ioffe, Shlens, and Wojna]{smoothing}
Szegedy, C., Vanhoucke, V., Ioffe, S., Shlens, J., and Wojna, Z.
\newblock Rethinking the inception architecture for computer vision.
\newblock In \emph{Proceedings of the IEEE/CVF Conference on Computer Vision and Pattern Recognition}, 2016.

\bibitem[Tenison et~al.(2022)Tenison, Sreeramadas, Mugunthan, Oyallon, Belilovsky, and Rish]{tenison2022gradient}
Tenison, I., Sreeramadas, S.~A., Mugunthan, V., Oyallon, E., Belilovsky, E., and Rish, I.
\newblock Gradient masked averaging for federated learning.
\newblock \emph{arXiv preprint arXiv:2201.11986}, 2022.

\bibitem[Terj{\'e}k(2019)]{terjek2019adversarial}
Terj{\'e}k, D.
\newblock Adversarial lipschitz regularization.
\newblock In \emph{International Conference on Learning Representations}, 2019.

\bibitem[Truong et~al.(2021)Truong, Maini, Walls, and Papernot]{dfme}
Truong, J.-B., Maini, P., Walls, R.~J., and Papernot, N.
\newblock Data-free model extraction.
\newblock In \emph{Proceedings of the IEEE/CVF Conference on Computer Vision and Pattern Recognition}, 2021.

\bibitem[Uesato et~al.(2018)Uesato, O’donoghue, Kohli, and Oord]{spsa}
Uesato, J., O’donoghue, B., Kohli, P., and Oord, A.
\newblock Adversarial risk and the dangers of evaluating against weak attacks.
\newblock In \emph{Proceedings of International Conference on Machine Learning}, 2018.

\bibitem[Wang et~al.(2021)Wang, Li, Singh, Lu, and Vasconcelos]{imagine}
Wang, P., Li, Y., Singh, K.~K., Lu, J., and Vasconcelos, N.
\newblock Imagine: Image synthesis by image-guided model inversion.
\newblock In \emph{Proceedings of the IEEE/CVF Conference on Computer Vision and Pattern Recognition}, 2021.

\bibitem[Wang et~al.(2023{\natexlab{a}})Wang, Zhang, Lei, and Zhang]{wang2023sharpness}
Wang, P., Zhang, Z., Lei, Z., and Zhang, L.
\newblock Sharpness-aware gradient matching for domain generalization.
\newblock In \emph{Proceedings of the IEEE/CVF Conference on Computer Vision and Pattern Recognition}, 2023{\natexlab{a}}.

\bibitem[Wang et~al.(2019)Wang, Zou, Yi, Bailey, Ma, and Gu]{mart}
Wang, Y., Zou, D., Yi, J., Bailey, J., Ma, X., and Gu, Q.
\newblock Improving adversarial robustness requires revisiting misclassified examples.
\newblock In \emph{International Conference on Learning Representations}, 2019.

\bibitem[Wang et~al.(2023{\natexlab{b}})Wang, Chen, Yang, Guo, Jiang, Zhang, and Qi]{dfard}
Wang, Y., Chen, Z., Yang, D., Guo, P., Jiang, K., Zhang, W., and Qi, L.
\newblock Model robustness meets data privacy: Adversarial robustness distillation without original data.
\newblock \emph{arXiv preprint arXiv:2303.11611}, 2023{\natexlab{b}}.

\bibitem[Wu et~al.(2020)Wu, Xia, and Wang]{weightperturb}
Wu, D., Xia, S.-T., and Wang, Y.
\newblock Adversarial weight perturbation helps robust generalization.
\newblock In \emph{Advances in Neural Information Processing Systems}, 2020.

\bibitem[Xu et~al.(2020)Xu, Li, Zhuang, Liu, Cao, Liang, and Tan]{gdfq}
Xu, S., Li, H., Zhuang, B., Liu, J., Cao, J., Liang, C., and Tan, M.
\newblock Generative low-bitwidth data free quantization.
\newblock In \emph{European Conference on Computer Vision}, 2020.

\bibitem[Yang et~al.(2023)Yang, Shi, Wei, Liu, Zhao, Ke, Pfister, and Ni]{medmnistv2}
Yang, J., Shi, R., Wei, D., Liu, Z., Zhao, L., Ke, B., Pfister, H., and Ni, B.
\newblock Medmnist v2-a large-scale lightweight benchmark for 2d and 3d biomedical image classification.
\newblock \emph{Scientific Data}, 2023.

\bibitem[Yang et~al.(2020)Yang, Rashtchian, Zhang, Salakhutdinov, and Chaudhuri]{yang2020closer}
Yang, Y.-Y., Rashtchian, C., Zhang, H., Salakhutdinov, R.~R., and Chaudhuri, K.
\newblock A closer look at accuracy vs. robustness.
\newblock \emph{Advances in Neural Information Processing Systems}, 2020.

\bibitem[Yao et~al.(2021)Yao, Bielik, Tsankov, and Vechev]{automated}
Yao, C., Bielik, P., Tsankov, P., and Vechev, M.
\newblock Automated discovery of adaptive attacks on adversarial defenses.
\newblock \emph{Advances in Neural Information Processing Systems}, 2021.

\bibitem[Yin et~al.(2019)Yin, Kannan, and Bartlett]{yin2019rademacher}
Yin, D., Kannan, R., and Bartlett, P.
\newblock Rademacher complexity for adversarially robust generalization.
\newblock In \emph{International Conference on Machine Learning}, 2019.

\bibitem[Yin et~al.(2020)Yin, Molchanov, Alvarez, Li, Mallya, Hoiem, Jha, and Kautz]{deepinversion}
Yin, H., Molchanov, P., Alvarez, J.~M., Li, Z., Mallya, A., Hoiem, D., Jha, N.~K., and Kautz, J.
\newblock Dreaming to distill: Data-free knowledge transfer via deepinversion.
\newblock In \emph{Proceedings of the IEEE/CVF Conference on Computer Vision and Pattern Recognition}, 2020.

\bibitem[Yu et~al.(2020)Yu, Kumar, Gupta, Levine, Hausman, and Finn]{yu2020gradient}
Yu, T., Kumar, S., Gupta, A., Levine, S., Hausman, K., and Finn, C.
\newblock Gradient surgery for multi-task learning.
\newblock In \emph{Advances in Neural Information Processing Systems}, 2020.

\bibitem[Yun et~al.(2019)Yun, Han, Oh, Chun, Choe, and Yoo]{cutmix}
Yun, S., Han, D., Oh, S.~J., Chun, S., Choe, J., and Yoo, Y.
\newblock Cutmix: Regularization strategy to train strong classifiers with localizable features.
\newblock In \emph{Proceedings of the IEEE/CVF International Conference on Computer Vision}, 2019.

\bibitem[Zagoruyko \& Komodakis(2016)Zagoruyko and Komodakis]{wrn}
Zagoruyko, S. and Komodakis, N.
\newblock Wide residual networks.
\newblock In \emph{British Machine Vision Conference}, 2016.

\bibitem[Zhang et~al.(2019)Zhang, Yu, Jiao, Xing, El~Ghaoui, and Jordan]{trades}
Zhang, H., Yu, Y., Jiao, J., Xing, E., El~Ghaoui, L., and Jordan, M.
\newblock Theoretically principled trade-off between robustness and accuracy.
\newblock In \emph{International Conference on Machine Learning}, 2019.

\bibitem[Zhao et~al.(2022)Zhao, Zhang, and Hu]{zhao2022penalizing}
Zhao, Y., Zhang, H., and Hu, X.
\newblock Penalizing gradient norm for efficiently improving generalization in deep learning.
\newblock In \emph{International Conference on Machine Learning}, 2022.

\bibitem[Zhong et~al.(2022)Zhong, Lin, Nan, Liu, Zhang, Tian, and Ji]{intraq}
Zhong, Y., Lin, M., Nan, G., Liu, J., Zhang, B., Tian, Y., and Ji, R.
\newblock Intraq: Learning synthetic images with intra-class heterogeneity for zero-shot network quantization.
\newblock In \emph{Proceedings of the IEEE/CVF Conference on Computer Vision and Pattern Recognition}, 2022.

\bibitem[Zhou et~al.(2020)Zhou, Wu, Liu, Liu, and Zhu]{dast}
Zhou, M., Wu, J., Liu, Y., Liu, S., and Zhu, C.
\newblock Dast: Data-free substitute training for adversarial attacks.
\newblock In \emph{Proceedings of the IEEE/CVF Conference on Computer Vision and Pattern Recognition}, 2020.

\bibitem[Zhu et~al.(2021)Zhu, Hofstee, Peltenburg, Lee, and Alars]{autorecon}
Zhu, B., Hofstee, P., Peltenburg, J., Lee, J., and Alars, Z.
\newblock {AutoReCon: Neural architecture search-based reconstruction for data-free compression}.
\newblock In \emph{International Joint Conferences on Artificial Intelligence}, 2021.

\bibitem[Zhu et~al.(2022)Zhu, Yao, Han, Zhang, Liu, Niu, Zhou, Xu, and Yang]{iad}
Zhu, J., Yao, J., Han, B., Zhang, J., Liu, T., Niu, G., Zhou, J., Xu, J., and Yang, H.
\newblock Reliable adversarial distillation with unreliable teachers.
\newblock In \emph{International Conference on Learning Representations}, 2022.

\bibitem[Zi et~al.(2021)Zi, Zhao, Ma, and Jiang]{rslad}
Zi, B., Zhao, S., Ma, X., and Jiang, Y.-G.
\newblock Revisiting adversarial robustness distillation: Robust soft labels make student better.
\newblock In \emph{Proceedings of the IEEE International Conference on Computer Vision}, 2021.

\end{thebibliography}
\bibliographystyle{icml2024}

\newpage
\appendix
\onecolumn

\newpage

\section*{Appendix}

We provide a more extensive set of experimental results with some analyses that we could not include in the main body due to space constraints. 
The contents of this material are as below:
\begin{itemize} 
 \setlength\itemsep{0.em}
    \item \textbf{Detailed Experimental Settings (\cref{sec:supp:expsetting})}: We provide detailed information of our experiments.
    \item \textbf{Overall Procedure of \name (\cref{sec:supp:pseudocode})}: We provide pseudo-code of the overall procedure of \name.
    \item \textbf{Number of Synthetic Samples (\cref{sec:supp:num_samples})}: We study the effect of using different numbers of synthetic samples.
    \item \textbf{Extended Set of Experiments on medical dataset (\cref{sec:supp:medmnist_extend})}: Extended results on medical dataset are reported.
    \item \textbf{Extended Set of Experiments on $\epsilon$-bounds (\cref{sec:supp:eps})}: Extended results on diverse attack distance are presented.
    \item \textbf{Detailed Study on \name (\cref{sec:supp:detailed})}: Extended results of detailed study of \name on sample diversity and comparison against different training loss functions.
    \item \textbf{Evaluation under Modified Attacks (\cref{sec:supp:modified_attack}}: We evaluate robustness of \name under modified attacks.
    \item \textbf{Further Visualization of Loss Surface (\cref{sec:supp:trainloss})}: We provide further analysis on $\lossdfs$ and its effect on the loss surface.
    \item \textbf{Sensitivity Study on the Number of Aggregated Batches (\cref{sec:supp:batch_sensitivity})}: We conduct sensitivity study on the number of aggregated batches.
    \item \textbf{Sensitivity Study on $\tau$ (\cref{sec:supp:tau_sensitivity})}: We conduct sensitivity study on the threshold value $\tau$ used in \gradmask.
    \item \textbf{Sensitivity Study on $\lambda_1$ and $\lambda_2$ (\cref{sec:supp:lambda_sensitivity})}: We conduct sensitivity study on the hyperparameters of $\lossdfs$, $\lambda_1$ and $\lambda_2$.
    \item \textbf{Visualization of \mixabb using PCA (\cref{sec:supp:pca})}: We provide PCA visualizations of different synthetic datasets to study the effect of \mixabb on diversity.
    \item \textbf{Generated Synthetic Data (\cref{sec:supp:samples})}: Selected examples of synthetic data are presented.
\end{itemize}

\section{Detailed Experimental Settings}
\label{sec:supp:expsetting}
In this section, we provide details on experimental settings for both synthetic data generation and robust training.
For baseline implementation of DaST~\citep{dast}, DFME~\citep{dfme}, and AIT~\citep{ait}, we used the original code from the authors, except for the modifications we specified in \cref{sec:supp:expsetting:baseline}.
For DFARD~\citep{dfard} we followed the description in the publication since the original implementation is not available, and used ACGAN~\citep{acgan} due to missing details of a generator architecture in the original publication.
All experiments have been conducted using PyTorch 1.9.1 and Python 3.8.0 running on Ubuntu 20.04.3 LTS with CUDA version 11.1 using RTX3090 and A6000 GPUs.

\subsection{Code}
The code used for the experiment is included in a zip archive in the supplementary material, along with the script for reproduction. 
The code is under Nvidia Source Code License-NC and GNU General Public License v3.0. 

\subsection{Data Generation}
When optimizing gaussian noise, we use Adam optimizer with learning rate $=0.1$ with batch size of 200, which we optimize for 1000 iterations on \medmnist{} and 2000 on general domain datasets.
For \mix, we set the range [0,1] for the sampling distribution of coefficients, and use uniform distribution.
Code implementation for \mix builds upon a prior work~\citep{deepinversion}.
For \medmnist{} results, we generated 10,000 samples for training, and for the other datasets we used 60,000 samples.
To accelerate data generation, we use multiple GPUs in parallel where 10,000 samples are generated with each.
With batch size of 200, generating 10,000 samples of size 28$\times$28 using ResNet-18 takes 0.7 hours on RTX 3090. 
For 32$\times$32 sized samples, ResNet-20 takes 0.6 hours, ResNet-56 2.6 hours, and WRN-28-10 3.6 hours.

For \medmnist{}, we trained two network architectures ResNet-18 and ResNet-50 used in the original paper ~\cite{medmnistv2}.
We report the pretraining results in \cref{tab:supp:medmnist_teacher}.
\begin{table*}[h]
\centering
\caption{Performance of pretrained teachers on \medmnist{}.}
\begin{tabular}{ccc}
\hline
Dataset     & \multicolumn{1}{l}{ResNet-18} & \multicolumn{1}{l}{ResNet-50} \\ \hline
Tissue & 67.62                         & 68.29                         \\
Blood  & 95.53                         & 95.00                         \\
Derma  & 74.61                         & 73.92                         \\
Path   & 92.19                         & 91.41                         \\
OCT    & 80.60                         & 84.90                         \\
OrganA & 93.75                         & 94.04                         \\
OrganC & 90.74                         & 91.06                         \\
OrganS & 78.75                         & 78.37                         \\ 
\hline
\end{tabular}
\label{tab:supp:medmnist_teacher}
\end{table*}

\subsection{Adversarial Training}
For adversarial training, we used SGD optimizer with learning rate=1e-4, momentum=0.9, and batch size of 200 for 100 epochs, and 200 epochs for ResNet-20 and ResNet-18.
All adversarial perturbations were created using PGD-10~\citep{madry} with the specified $\epsilon$-bounds.
Following the convention, $l_2$-norm attacks are bounded by $\epsilon=128/255$ with step size of $15/255$.
$l_{\infty}$-norm attacks are evaluated under a diverse set of distances $\epsilon=\{8/255, 6/255, 4/255, 2/255\}$, which all use step size$=\epsilon/4$.
For $\lossdfs$, we simply use $\lambda_1=1$ and $\lambda_2=1$, which we found to best balance the learning from three different objective terms.
For \gradmask, we use $\mathcal{B}=\{10,20\}$ for all settings, which we found to perform generally well across different datasets and models.
When using \gradmask, we increment the learning rate linearly with $\mathcal{B}$ to take into consideration the increased effective batch size.
We use $\tau=0.5$ for all our experiments with \gradmask.

\subsection{Adaptation of the Baselines}
\label{sec:supp:expsetting:baseline}
In this section, we describe how we adapted the baselines (\cref{tab:baseline_losses}) to the problem of data-free adversarial robustness.
DaST~\citep{dast} is a black-box attack method with no access to the original data. 
DaST trains a substitute model using samples from a generative model~\citep{gan} to synthesize samples for querying the victim model.
To adapt DaST to our problem, we keep the overall framework but modify the training loss, substituting clean samples with perturbed ones.
This makes it possible to use the training algorithm, while the objective now is to robustly train a model with no data.

DFME~\citep{dfme} is a more recent work on data-free model extraction that also utilizes synthetic samples for model stealing. 
They leverage distillation methods~\citep{dfad} to synthesize samples that maximize student-teacher disagreement.
Similar to DaST, we substitute the student input to perturbed ones, while keeping other settings the same.

AIT~\citep{ait} utilizes the full precision model's feedback for training its generative model.
Unlike DaST and DFME which focus on student outputs when training the generator, AIT additionally utilizes the batch-normalization statistics stored in the teacher model for creating synthetic samples.
Since AIT is a model quantization method, its student model is of low-bit precision, and thus their training loss cannot be directly adopted to our task. 
We use TRADES~\citep{trades} loss function for training, a variation of STD~\citep{madry}. 

Lastly, DFARD~\citep{dfard} suggests data-free robust distillation.
Given a model already robustly trained, the goal is to distill its robustness to a lighter network.
They use adaptive distillation temperature to regulate learning difficulty. 
While this seems to align with the data-free adversarial robustness, the robust teacher is not available in our problem.
Therefore, we replace the robustly pretrained model with the given (non-robust) $T(x)$ so that student can correctly classify perturbed samples.

For implementations of test-time defense techniques including DAD~\citep{dad}, DiffPure~\citep{diffpure}, and TTE~\citep{TTE}, we used the official code provided by the authors.
For DAD, we follow their method and retrain a CIFAR-10 pretrained detector on \medmnist{} test set. 
Since their method assumes the usability of off-the-shelf detector, we used their pretrained detector trained on CIFAR-10 and fine-tuned it using each dataset from \medmnist.
Note that DAD is evaluated on the same test set that is used to finetune the detector module, which could be considered data leak.
For TTE, where we used +flip+4crops+4 flipped-crops as it is reported as the best setting in the original paper.
Lastly, for DiffPure, we used the same setting the authors used for evaluating on CIFAR-10 dataset, including the pretrained score SDE model.
To match the image size of the pretrained model, we resized the image samples originally sized as 28x28 to 32x32.
Also, note that DiffPure requires high computational cost for evaluating against adaptive attacks such as AutoAttack~\citep{autoattack}, and so the authors random sample 512 images for all evaluations. 
However, we used 2000 images randomly sampled from the test set for more accurate evaluation, except for the cases where the test set is smaller than 2000.

\section{Overall Procedure of \name}
\label{sec:supp:pseudocode}
The pseudo-code of the overall procedure of \name is depicted in \cref{alg:procedure}.
It comprises data generation using \emph{\mix} (line 4-10, \S\ref{sec:mix}), and adversarial training using a novel loss function (line 15, \S\ref{sec:Ltrain}) along with a gradient refinement technique (line 17-20, \S\ref{sec:gradmask}).

\newcommand{\Comment}[1]{{\hfill $\triangleright$ \textrm{#1}}}

\begin{algorithm}

\caption{Procedure of \name}
\label{alg:procedure}
\begin{algorithmic}[1]
\small
\STATE \textbf{Inputs:} {set of synthesis loss terms $\mathbb{S}$, number of batches for synthesis $N$, pretrained model's parameters $\theta_T$, target model for training $\theta_S$, synthesis iterations $Q$, train iterations $P$, number of aggregated batches $\mathcal{B}$, learning rate for synthesis $\eta_g$ and training $\eta_s$.} 
\STATE \textbf{Initialize:} {{$\theta_S$} $\gets$ $\theta_T$} \Comment{Initialize target model with pretrained model}
\STATE \textbf{Initialize:} $X=\{X_1,...,X_N\} \gets Z \sim \mathcal{N}(0,1)$ \Comment{Initialize batches with random noise}
\FOR {$i$=1 ,..., $N$}
    \STATE Sample $\{\alpha_1,...,\alpha_{|\mathbbm{S}|}\}$ from $\mathcal{U}(0,1)$ 
    \STATE$\loss_{Synth} = \sum_{\mathbbm{s}=1}^{|\mathbbm{S}|} \alpha_{\mathbbm{s}} * \loss_{\mathbbm{s}}$ \Comment{\MIX (\S\ref{sec:mix})}
    \FOR{$q$=1 ,..., $Q$} 
    \STATE${X_i} \gets X_i - \eta_g \nabla_{X_i}\loss_{Synth}(X_i;\theta_T)$
\ENDFOR
\ENDFOR
    \FOR{$p=1,...,P$}
    \STATE Sample $\mathcal{B}$ mini-batches $\{X_1,...,X_\mathcal{B}\}$ from $X$
    \FOR{$b$=1 ,..., $\mathcal{B}$}
    \STATE ${X_b'} \gets PGD(X_b;\theta_S)$                                   \Comment{\cref{eq:pgd}}
    \STATE $g^{(b)} \gets \nabla_{\theta_S}\lossdfs(X_b, X_b')$      \Comment{$\lossdfs$ (\S\ref{sec:Ltrain})}
    \ENDFOR
    \FOR{$k=1,...,\left|\theta_S\right|$ (in parallel)}
    \STATE ${A}_{k} = \frac{1}{\mathcal{B}}  \sum_{b=1}^{\mathcal{B}} sign(g_k^{(b)})$
    \Comment{\gradmask (\S\ref{sec:gradmask})}
    \STATE $g^*_k = \Phi({A}_k) \cdot  \sum^{\mathcal{B}}_{b=1} \mathbbm{1}_{\{{A}_{k} \cdot g^{(b)}_{k}>0\}} \cdot g^{(b)}_{k}$    \Comment{\cref{eq:gradmask3}}
\ENDFOR
    \STATE $\mathit{{\theta_S} \gets {\theta_S} - \eta_s g^*}$ \Comment{Update using refined gradient}

\ENDFOR

\end{algorithmic}

\end{algorithm}

\section{Number of Synthetic Samples}
\label{sec:supp:num_samples}

\begin{figure}
\includegraphics[width=0.5\columnwidth]{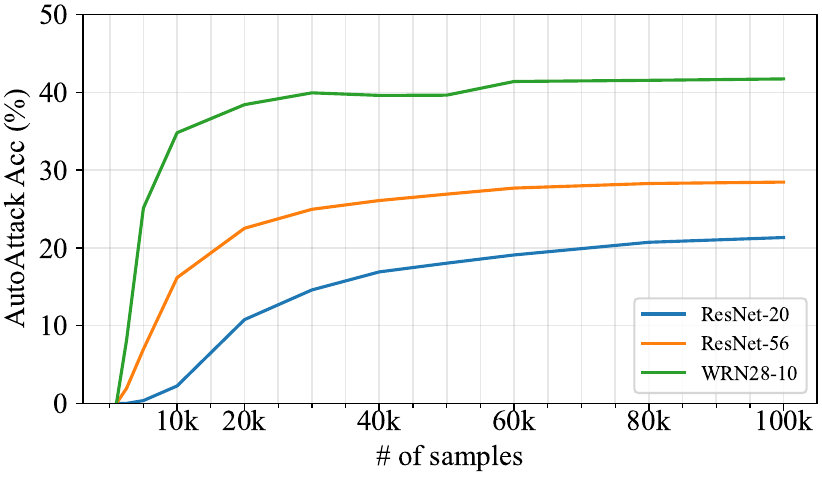}
\includegraphics[width=0.5\columnwidth]{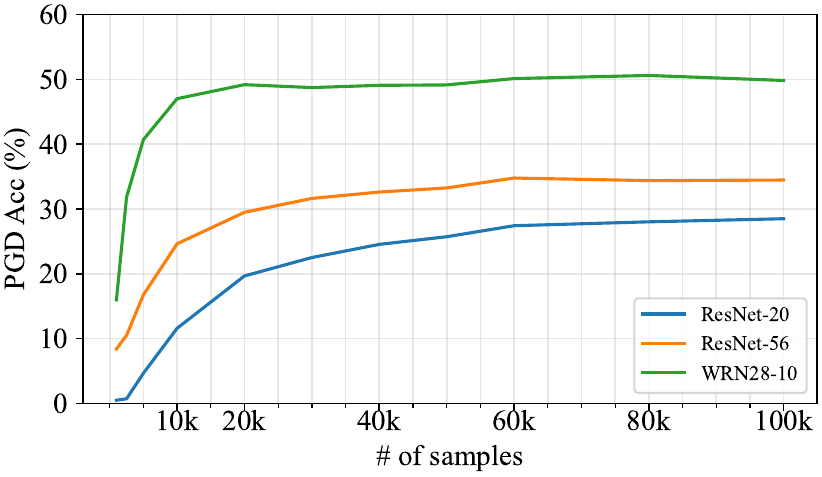}
\caption{Comparing performance using varying number of samples for training. Left denotes AutoAttack accuracy while the right denotes PGD-10 accuracy.}
\label{fig:supp:num_samples:AA}
\end{figure}

In this section, we show the performance gain from simply incrementing the number of synthetic samples.
\HY{ref fig here}\cref{fig:supp:num_samples:AA} plots the AutoAttack accuracy when trained using differing number of samples.
For all models, the trend is similar in that the performance increases linearly, and converge at some point around 50000-60000.
Although there exists marginal gain with further supplement of data, we settle for 60000 samples for the training efficiency.
One observation is that for smaller model (ResNet-20), it is much harder to obtain meaningful robustness for any set under 20000.
We posit this is due to the characteristic of data-free synthesis, where the only guidance is from a pretrained model and the quality of the data is bounded by the performance of the pretrained model.
Since larger models tend to learn better representation, it can be reasoned that the smaller models are less capable of synthesizing good quality data, along with the reason that smaller models are generally harder to train for adversarial robustness.

\section{Extended Set of Experiments on Medical Datasets}
\label{sec:supp:medmnist_extend}
In this section we report an extended version of our experiment on \medmnist, which includes Derma, OCT, OrganA, and OrganS datasets. 
\cref{tab:supp:medmnist_extend_main} compares \name against data-free baseline methods, and \cref{tab:supp:medmnist_extend_post_train} shows results on test-time defense techniques.
Aligned with the observation from \cref{tab:main_medmnist_post_train}, \name is the only method that provides consistently high robustness.
\HY{add explanation}

\begin{table*}[h]
\centering
    \caption{Performance on \medmnist{} with $l_\infty$ perturbation budget.}
    \def\arraystretch{0.88}%
\resizebox{.88\textwidth}{!}{

\begin{tabular}{clcccccccccccc}
\midrule
\multirow{2}{*}{Model} &
\multirow{2}{*}{\makecell{Method}}                                                                                                       
                         & \multicolumn{3}{c}{Derma} & \multicolumn{3}{c}{OCT} & \multicolumn{3}{c}{OrganA} & \multicolumn{3}{c}{OrganS}  \\
                         \cmidrule(lr){3-5}  \cmidrule(lr){6-8}  \cmidrule(lr){9-11}  \cmidrule(lr){12-14} 
                         & & $\mathcal{A}_{Clean}$ & $\bf\mathcal{A}_{PGD}$ & $\bf\mathcal{A}_{AA}$      &      $\mathcal{A}_{Clean}$ & $\bf\mathcal{A}_{PGD}$ & $\bf\mathcal{A}_{AA}$ &
                         $\mathcal{A}_{Clean}$ & $\bf\mathcal{A}_{PGD}$ & $\bf\mathcal{A}_{AA}$  &
                         $\mathcal{A}_{Clean}$ & $\bf\mathcal{A}_{PGD}$ & $\bf\mathcal{A}_{AA}$  \\ 
\midrule
 \multirowcell{6}{RN-18} 

& Public &	67.48&	50.47	&42.39	&	25.60	&	22.00	&	20.90	&86.79	&33.82	&30.28&	68.82	&	18.25	&	15.41	\\				
\cmidrule{2-14}																						
& DaST &	66.93	&	38.35	&	32.72	&	27.00	&	20.70	&	\textbf{20.50}	&	84.42	&	26.07	&	22.70	&	28.55	&	{\textcolor{white}0}9.40	&	{\textcolor{white}0}8.82	\\
&  DFME &	69.63	&	24.29	&	21.70	&	25.00	&	19.10	&	14.00	&	92.20	&	24.31	&	20.27	&	72.11	&	{\textcolor{white}0}9.49	&	{\textcolor{white}0}7.29	\\
&  AIT &	66.03	&	35.71	&	33.97	&	21.50	&	11.60	&	{\textcolor{white}0}9.50	&	70.50	&	11.31	&	{\textcolor{white}0}8.74	&	50.71	&	13.13	&	{\textcolor{white}0}8.29	\\
& DFARD &	74.91	&	{\textcolor{white}0}5.04	&	{\textcolor{white}0}3.89	&	69.10	&	{\textcolor{white}0}0.00	&	{\textcolor{white}0}0.00	&	90.81	&	31.71	&	28.83	&	63.46	&	10.86	&	{\textcolor{white}0}8.88	\\
\cmidrule{2-14}																									
& \textbf{\nameabb}   &	66.98	&	\textbf{63.09}	&	\textbf{61.85}	&	31.90	&	\textbf{23.80}	&	18.50	&	86.44	&	\textbf{49.06}	&	\textbf{44.90}	&		64.43&	\textbf{29.68}&	\textbf{24.31}\\
\midrule																									
\multirowcell{6}{RN-50}

& Public &	67.08&	44.59&	34.11	&	25.30	&	22.60	&	19.40	&78.77	&33.56	&28.78&	47.70	&	12.68	&	{\textcolor{white}0}4.28	\\
\cmidrule{2-14}	

& DaST &	72.47	&	{\textcolor{white}0}8.08	&	{\textcolor{white}0}4.99	&	25.00	&	{\textcolor{white}0}5.60	&	{\textcolor{white}0}0.00	&	64.05	&	17.45	&	15.61	&	53.49	&	{\textcolor{white}0}8.93	&	{\textcolor{white}0}5.29	\\
&  DFME &	67.08	&	{\textcolor{white}0}3.19	&	{\textcolor{white}0}1.10	&	25.00	&	{\textcolor{white}0}8.40	&	{\textcolor{white}0}0.00	&	22.42	&	{\textcolor{white}0}2.58	&	{\textcolor{white}0}1.82	&	74.50	&	{\textcolor{white}0}6.93	&	{\textcolor{white}0}4.21	\\
&  AIT &	23.59	&	{\textcolor{white}0}3.29	&	{\textcolor{white}0}1.75	&	24.50	&	{\textcolor{white}0}1.60	&	{\textcolor{white}0}0.70	&	46.77	&	{\textcolor{white}0}9.21	&	{\textcolor{white}0}7.08	&	27.48	&	{\textcolor{white}0}7.02	&	{\textcolor{white}0}6.13	\\
& DFARD &	54.02	&	10.97	&	{\textcolor{white}0}9.23	&	25.50	&	18.50	&	{\textcolor{white}0}1.30	&	31.37	&	22.60	&	19.82	&	76.09	&	{\textcolor{white}0}6.33	&	{\textcolor{white}0}4.07	\\
\cmidrule{2-14}																									
& \textbf{\nameabb}   &	67.78	&	\textbf{64.34}	&	\textbf{58.05}	&	28.90	&	\textbf{19.30}	&	\textbf{17.59}	&	90.80	& \textbf{42.58}	&\textbf{37.42}&		66.61	&\textbf{37.84}	&\textbf{33.63}	\\
\bottomrule

\end{tabular}
}
    \label{tab:supp:medmnist_extend_main}

\end{table*}
\begin{table}[h]
    \centering
    
    \caption{Performance on \medmnist{} with $l_\infty$ perturbation budget using test-time defense methods.}
            \vspace{1mm}
    \label{tab:supp:medmnist_extend_post_train}
    
    \def\arraystretch{0.8}%
    \resizebox{.6\columnwidth}{!}
    {
    \begin{tabular}{clcccccc}
    \toprule
     & \multicolumn{1}{r}{} &  \multicolumn{3}{c}{ResNet-18} & \multicolumn{3}{c}{ResNet-50} \\
    \cmidrule(lr){3-5}\cmidrule(lr){6-8}
      Dataset  & Method &   $\mathcal{A}_{Clean}$& $\bf\mathcal{A}_{PGD}$ & $\bf\mathcal{A}_{AA}$ & $\mathcal{A}_{Clean}$& $\bf\mathcal{A}_{PGD}$ & $\bf\mathcal{A}_{AA}$ \\
      \midrule

\multirowcell{4}{Derma} 

&   DAD & 	74.51	&	13.47	&	{\textcolor{white}0}4.24	&	72.02	&	23.79	&	{\textcolor{white}0}3.39	\\
& DiffPure &	65.04	&	55.66	&	28.72	&	68.36	&	59.38	&	45.18	\\
& TTE &	70.47	&	15.81	&	11.62	&	53.57	&	{\textcolor{white}0}2.29	&	{\textcolor{white}0}1.10	\\
\cmidrule{2-8}												
&  \textbf{\nameabb} &	66.98	&	\textbf{63.09}	&	\textbf{61.85}	&	67.78	&	\textbf{64.34}	&	\textbf{58.05}	\\
\midrule

\multirowcell{4}{OCT} 

&   DAD & 	80.20	&	22.90	&	{\textcolor{white}0}0.20	&	84.50	&	21.70	&	{\textcolor{white}0}0.00	\\
& DiffPure &	30.66	&	\textbf{26.76}	&	{\textcolor{white}0}9.10	&	28.91	&	29.10	&	16.60	\\
& TTE &	67.10	&	{\textcolor{white}0}0.00	&	{\textcolor{white}0}0.00	&	83.30	&	{\textcolor{white}0}0.00	&	{\textcolor{white}0}0.00	\\
\cmidrule{2-8}												
&  \textbf{\nameabb} &	31.90	&	23.80	&	\textbf{18.50}	&	28.90	&	19.30	&	\textbf{17.59}	\\
\midrule  																	
\multirowcell{4}{OrganA} 

&   DAD & 	85.07	&	37.78	&	19.46	&	81.48	&	34.14	&	13.83	\\
& DiffPure &	70.12	&	\textbf{63.48}	&	35.94	&	69.34	&	\textbf{59.96}	&	\textbf{42.79}	\\
& TTE &	72.38	&	31.88	&	25.78	&	68.38	&	24.77	&	17.06	\\
\cmidrule{2-8}												
&  \textbf{\nameabb} &	86.44	&	49.06	&	\textbf{44.90}	&	90.80	&42.58	&37.42\\
\midrule 

\multirowcell{4}{OrganS} 

&   DAD & 	47.86	&	29.45	&	{\textcolor{white}0}5.54	&	57.61	&	30.58	&	{\textcolor{white}0}5.84	\\
& DiffPure &	55.27	&	45.70	&	\textbf{24.53}	&	52.34	&	\textbf{46.88}	&	24.92	\\
& TTE &	61.16	&	{\textcolor{white}0}5.35	&	{\textcolor{white}0}2.07	&	53.49	&	{\textcolor{white}0}9.72	&	{\textcolor{white}0}3.79	\\
\cmidrule{2-8}												
&  \textbf{\nameabb} &64.43&	\textbf{29.68}	&24.31	&66.61	&37.84	&\textbf{33.63}\\
\bottomrule
    \end{tabular}
    }
\end{table}

\section{Extended Set of Experiments on $\epsilon$-bounds}
\label{sec:supp:eps}
\begin{table*}[h]
    \centering
    \setlength{\tabcolsep}{4pt}
    
    \caption{Performance on SVHN.}
            \vspace{1mm}
    \label{tab:supp:eps:svhn}
    
    \def\arraystretch{0.8}%
    \resizebox{.85\textwidth}{!}
    {
    \begin{tabular}{clccccccccc}
    \toprule
     & \multicolumn{1}{r}{} & \multicolumn{3}{c}{ResNet-20} & \multicolumn{3}{c}{ResNet-56} & \multicolumn{3}{c}{WRN-28-10} \\
    \cmidrule(lr){3-5}\cmidrule(lr){6-8}\cmidrule(lr){9-11}
      $\epsilon$  & Method &  $\mathcal{A}_{Clean}$ & $\bf\mathcal{A}_{PGD}$ & $\bf\mathcal{A}_{AA}$ & $\mathcal{A}_{Clean}$& $\bf\mathcal{A}_{PGD}$ & $\bf\mathcal{A}_{AA}$ & $\mathcal{A}_{Clean}$& $\bf\mathcal{A}_{PGD}$ & $\bf\mathcal{A}_{AA}$ \\
      \midrule

\multirowcell{7}{$2/255$} 				
& Original & 95.42 & 88.04 & 86.72& 96.00 & 88.86 & 87.85& 96.06& 89.23 & 88.16\\
\cmidrule(lr){2-11}	
& DaST &	93.80	&	34.30	&	12.33	&	91.00	&	46.29	&	31.77	&	96.45	&	35.21	&	{\textcolor{white}0}9.49	\\		
& DFME  & 	96.05	&	35.24	&	{\textcolor{white}0}8.39	&	97.30	&	38.79	&	10.98	&	97.21	&	24.67	&	{\textcolor{white}0}0.54	\\		
& AIT  & 	94.67	&	65.74	&	60.74	&	95.63	&	70.42	&	66.23	&	85.82	&	44.33	&	36.37	\\		
& DFARD &	96.58	&	32.64	&	{\textcolor{white}0}6.89	&	97.29	&	39.21	&	{\textcolor{white}0}8.94	&	97.11	&	26.38	&	{\textcolor{white}0}0.29	\\	\cmidrule(lr){2-11}	
&  \textbf{\name} 	&	94.22	&	\textbf{75.56}	&	\textbf{72.17}	&	94.16	&	\textbf{80.32}	&	\textbf{78.47}	&	95.94	&	\textbf{84.63}	&	\textbf{82.93}	\\
\midrule	\multirowcell{7}{$4/255$}	
&   Original & 93.19& 78.01& 74.59 & 94.67 &79.53 & 79.67 & 94.48& 79.53& 76.72 \\
\cmidrule(lr){2-11}	
 &  DaST & 20.66 & 13.90 & {\textcolor{white}0}7.06 & 20.20$^{\dagger}$ & 19.59 & 19.65 & 20.15 & 19.17 & 14.57	\\																
 &  DFME & 11.32$^{\dagger}$& {\textcolor{white}0}2.59& {\textcolor{white}0}0.84& 20.20$^{\dagger}$& 19.22& {\textcolor{white}0}4.27& {\textcolor{white}0}6.94$^{\dagger}$& {\textcolor{white}0}5.31&	{\textcolor{white}0}0.28\\
 & AIT &	91.45	& 37.87	&	24.74 &	86.65	&	45.45	&	38.96 & 83.89	&	40.45	&	33.06	\\
& DFARD &	20.11	&	15.94	&	19.68	&	19.58	&	15.43	&	{\textcolor{white}0}0.00	&	92.32	&	13.08	&	{\textcolor{white}0}0.01	\\
\cmidrule(lr){2-11}	
&  \textbf{\name} 	&	91.83	&	\textbf{54.82}	&	\textbf{47.55}	&	88.66	&	\textbf{62.05}	&	\textbf{57.54}	&	94.14	&	\textbf{69.60}	&	\textbf{62.66}	\\
\midrule	\multirowcell{7}{$6/255$}
& Original & 91.47 & 67.39 &60.56 & 91.59 & 71.10 & 57.95 & 93.62 & 75.03 & 57.36\\
\cmidrule(lr){2-11}	
& DaST &	{\textcolor{white}0}7.84	&	{\textcolor{white}0}1.64	&	{\textcolor{white}0}0.00	&	19.68	&	19.57	&	12.79	&	61.72	&	{\textcolor{white}0}8.82	&	{\textcolor{white}0}0.00	\\		
& DFME  & 	15.90$^{\dagger}$	&	15.94	&	14.81	&	97.34	&	{\textcolor{white}0}5.21	&	{\textcolor{white}0}0.00	&	97.11	&	{\textcolor{white}0}1.39	&	{\textcolor{white}0}0.00	\\		
& AIT  & 	83.70	&	23.20	&	{\textcolor{white}0}6.03	&	87.23	&	30.06	&	17.37	&	77.05	&	12.45	&	{\textcolor{white}0}3.61	\\		
& DFARD &	24.27	&	19.48	&	{\textcolor{white}0}0.44	&	97.17	&	{\textcolor{white}0}5.87	&	{\textcolor{white}0}0.00	&	54.24	&	19.58	&	{\textcolor{white}0}0.00	\\	\cmidrule(lr){2-11}	
&  \textbf{\name} 	&	89.00	&	\textbf{39.63}	&	\textbf{31.15}	&	81.90	&	\textbf{47.36}	&	\textbf{40.88}	&	92.18	&	\textbf{55.39}	&	\textbf{45.57}	\\
\midrule	\multirowcell{7}{$8/255$}	
& Original & 86.50 & 55.68 & 40.31 & 89.29 & 59.39 & 51.21 & 92.03 & 68.35 & 32.94\\
\cmidrule(lr){2-11}	
& DaST & 10.29 & {\textcolor{white}0}3.94 & {\textcolor{white}0}2.07 & 19.68$^{\dagger}$ & 19.59 & 19.68 & 20.39 & 16.69 & {\textcolor{white}0}1.35	\\													
& DFME & 20.15& {\textcolor{white}0}0.30& {\textcolor{white}0}0.00& 21.55& 16.60& {\textcolor{white}0}0.22& {\textcolor{white}0}6.84$^{\dagger}$& {\textcolor{white}0}6.70&	{\textcolor{white}0}2.29\\
& AIT 	&	47.47	&	15.21	&	{\textcolor{white}0}7.70 &	73.33	&	22.42	&	10.92 &	47.96	&	14.85	&	{\textcolor{white}0}7.24	\\
& DFARD &	20.03	&	13.46	&	{\textcolor{white}0}0.00	&	25.18	&	{\textcolor{white}0}5.46	&	{\textcolor{white}0}0.00	&	93.07	&	18.23	&	{\textcolor{white}0}0.02	\\	\cmidrule(lr){2-11}	
&  \textbf{\name} 	&	85.32	&	\textbf{29.96}	&	\textbf{20.84}	&	75.70	&	\textbf{37.32}	&	\textbf{29.04}	&	90.57	&	\textbf{43.80}	&	\textbf{31.77}	\\
     \bottomrule
     \multicolumn{11}{r}{$^\dagger$Did not converge}\\
    \end{tabular}
    }

\end{table*}

\begin{table*}[]
    \centering
        \setlength{\tabcolsep}{4pt}
    
    \caption{Performance on CIFAR-10.}
            \vspace{1mm}
    \label{tab:supp:eps:cifar10}
    
    \def\arraystretch{0.8}%
    \resizebox{.85\textwidth}{!}
    {
    \begin{tabular}{clccccccccc}
    \toprule
     & \multicolumn{1}{r}{} & \multicolumn{3}{c}{ResNet-20} & \multicolumn{3}{c}{ResNet-56} & \multicolumn{3}{c}{WRN-28-10} \\
    \cmidrule(lr){3-5}\cmidrule(lr){6-8}\cmidrule(lr){9-11}
      $\epsilon$  & Method &  $\mathcal{A}_{Clean}$ & $\bf\mathcal{A}_{PGD}$ & $\bf\mathcal{A}_{AA}$ & $\mathcal{A}_{Clean}$& $\bf\mathcal{A}_{PGD}$ & $\bf\mathcal{A}_{AA}$ & $\mathcal{A}_{Clean}$& $\bf\mathcal{A}_{PGD}$ & $\bf\mathcal{A}_{AA}$ \\
      \midrule

\multirowcell{7}{$2/255$} 
& Original & 78.82 & 66.59 & 65.53 & 81.01 & 68.34 & 67.42& 86.34 & 75.39 & 74.81\\
\cmidrule(lr){2-11}	
& DaST &	10.02$^{\dagger}$	&	{\textcolor{white}0}9.91	&	{\textcolor{white}0}9.77	&	17.46	&	{\textcolor{white}0}3.25	&	{\textcolor{white}0}0.50	&	12.72	&	{\textcolor{white}0}6.46	&	{\textcolor{white}0}2.37	\\		
& DFME  & 	32.05	&	{\textcolor{white}0}9.60	&	{\textcolor{white}0}4.19	&	91.29	&	16.28	&	{\textcolor{white}0}0.13	&	97.32	&	18.31	&	{\textcolor{white}0}0.00	\\		
& AIT & 	81.25	&	28.90	&	24.72	&	78.51	&	33.66	&	30.25	&	74.05	&	{\textcolor{white}0}6.66	&	{\textcolor{white}0}1.27	\\		
& DFARD &	91.89	&	{\textcolor{white}0}7.63	&	{\textcolor{white}0}0.08	&	95.34	&	16.97	&	{\textcolor{white}0}0.06	&	97.32	&	10.71	&	{\textcolor{white}0}0.00	\\	\cmidrule(lr){2-11}	
&  \textbf{\name} 	&	80.66	&	\textbf{50.09}	&	\textbf{46.73}	&	87.06	&	\textbf{57.99}	&	\textbf{55.44}	&	91.56	&	\textbf{70.48}	&	\textbf{68.12}	\\
\midrule	
\multirowcell{7}{$4/255$}
 &   Original & 75.60& 56.79& 54.37& 76.76 & 58.50& 56.54 & 82.89& 64.64& 62.84\\
\cmidrule(lr){2-11}	
&  DaST & 10.00$^{\dagger}$ & {\textcolor{white}0}9.89 & {\textcolor{white}0}8.62 & 12.06 &{\textcolor{white}0}7.68 & {\textcolor{white}0}5.32 & 10.00$^{\dagger}$ & {\textcolor{white}0}9.65 & {\textcolor{white}0}2.85	\\
   &DFME & 14.36& {\textcolor{white}0}5.23& {\textcolor{white}0}0.08& 13.81& {\textcolor{white}0}3.92& {\textcolor{white}0}0.03& 10.00$^{\dagger}$& {\textcolor{white}0}9.98&	{\textcolor{white}0}0.05\\
   & AIT &	32.89	&	11.93	&	10.67 &	38.47	&	12.29	&	11.36 &	34.92	&	10.90	&	{\textcolor{white}0}9.47	\\
& DFARD	&	12.28	&	{\textcolor{white}0}5.33	&	{\textcolor{white}0}0.00	&	10.84	&	{\textcolor{white}0}8.93	&	{\textcolor{white}0}0.00	&	{\textcolor{white}0}9.82	&	12.01	&	{\textcolor{white}0}0.02	\\
\cmidrule(lr){2-11}																			
&  \textbf{\name} 	&	74.79	&	\textbf{29.29}	&	\textbf{22.65}	&	81.30	&	\textbf{35.55}	&	\textbf{30.51}	&	86.74	&	\textbf{51.13}	&	\textbf{43.73}	\\

\midrule																	
\multirowcell{7}{$6/255$}
& Original  & 70.88 & 48.23 & 45.88 & 73.55 & 50.47 & 47.50& 77.89 & 54.56 & 52.23\\
\cmidrule(lr){2-11}	
& DaST &	10.00$^{\dagger}$	&	{\textcolor{white}0}9.86	&	{\textcolor{white}0}8.02	&	10.00$^{\dagger}$	&	{\textcolor{white}0}9.00	&	{\textcolor{white}0}2.21	&	10.17	&	{\textcolor{white}0}4.97	&	{\textcolor{white}0}0.07	\\		
& DFME  & 	10.00$^{\dagger}$	&	{\textcolor{white}0}0.82	&	{\textcolor{white}0}0.01	&	78.82	&	{\textcolor{white}0}3.35	&	{\textcolor{white}0}0.00	&	10.86	&	{\textcolor{white}0}9.26	&	{\textcolor{white}0}1.58	\\		
& AIT  & 	24.20	&	{\textcolor{white}0}7.71	&	{\textcolor{white}0}3.05	&	22.35	&	{\textcolor{white}0}9.46	&	{\textcolor{white}0}7.46	&	63.61	&	{\textcolor{white}0}3.87	&	{\textcolor{white}0}0.51	\\		
& DFARD &	11.23	&	{\textcolor{white}0}4.91	&	{\textcolor{white}0}0.00	&	95.27	&	{\textcolor{white}0}1.10	&	{\textcolor{white}0}0.00	&	92.46	&	{\textcolor{white}0}0.34	&	{\textcolor{white}0}0.00	\\	\cmidrule(lr){2-11}	
&  \textbf{\name} 	&	69.11	&	\textbf{17.94}	&	\textbf{11.03}	&	76.55	&	\textbf{21.55}	&	\textbf{16.11}	&	81.26	&	\textbf{37.26}	&	\textbf{26.07}	\\
\midrule																	
\multirowcell{7}{$8/255$} 
& Original & 69.19 & 41.69 & 37.30 & 70.79 & 43.89 & 39.97& 76.76 & 47.88 & 44.04\\
\cmidrule(lr){2-11}	
& DaST & {\textcolor{white}0}10.00$^{\dagger}$ & {\textcolor{white}0}9.99 & {\textcolor{white}0}6.81 & 10.60$^{\dagger}$ & {\textcolor{white}0}9.18 & {\textcolor{white}0}1.62 & 10.00$^{\dagger}$ & {\textcolor{white}0}9.88 & {\textcolor{white}0}0.56	\\		
 &  DFME & 13.17& {\textcolor{white}0}1.67& {\textcolor{white}0}0.00& 10.01$^{\dagger}$& {\textcolor{white}0}2.10& {\textcolor{white}0}0.00& 10.02$^{\dagger}$& {\textcolor{white}0}4.44&	{\textcolor{white}0}0.00\\
 &   AIT &	14.02	&	{\textcolor{white}0}3.49	&	{\textcolor{white}0}0.28 &	10.06$^{\dagger}$	&	{\textcolor{white}0}9.97	&{\textcolor{white}0}9.96 &	10.12$^{\dagger}$	&	{\textcolor{white}0}9.66	&	{\textcolor{white}0}8.16		\\
& DFARD &	11.23	&	{\textcolor{white}0}1.41	&	{\textcolor{white}0}0.00	&	13.04	&	{\textcolor{white}0}3.41	&	{\textcolor{white}0}0.00	&	10.11	&	{\textcolor{white}0}9.98	&	{\textcolor{white}0}0.00	\\	\cmidrule(lr){2-11}
&  \textbf{\name} 	&	63.69	&	\textbf{10.53}	&	\textbf{{\textcolor{white}0}4.71}	&	73.05	&	\textbf{13.27}	&	\textbf{{\textcolor{white}0}7.80}	&	76.63	&	\textbf{27.61}	&	\textbf{14.79}	\\

     \bottomrule
     \multicolumn{11}{r}{$^\dagger$Did not converge}\\
    \end{tabular}
    }
\end{table*}

In the field of empirical adversarial robustness~\citep{rebuffi2021data, schmidt2018adversarially, mart, weightperturb}, thorough evaluation under attacks of varying difficulties (number of iterations, size of $\epsilon$, etc) is needed to guarantee the model's robustness.
This is because a consistent trend across different attacks and resistance against strong attacks (AutoAttack) ensures the robustness is not from obfuscated gradients~\citep{obfuscated}.
In this regard, we provide further experiment results using diverse set of $\epsilon$-bounds using SVHN and CIFAR-10 in ~\cref{tab:supp:eps:svhn} and ~\cref{tab:supp:eps:cifar10}.
For each setting, we highlight the best results under $\mathcal{A}_{AA}$. 

In both datasets, baseline methods show poor performance regardless of the difficulty of the attack.
For example, in CIFAR-10, even at a relatively weaker attack of $\epsilon=2/255$, DaST, DFME, and DFARD do not exceed 10\% under AutoAttack evaluation, which is no better than random guessing.
Although AIT performs generally better than the other baselines, it suffers when training a larger model (WRN-28-10).
The overall trend of the baselines implies that these methods are unable to learn meaningful robustness, regardless of the size of the distortion.
On the other hand, \name shows consistent trend across all attacks.
While exceeding the baseline methods by a huge margin, the results are stable under both PGD and AutoAttack in all $\epsilon$'s.
This shows that \name is able to learn meaningful robustness from adversarial training of all presented distortion sizes.

\section{Detailed Study on \name}
\label{sec:supp:detailed}

We present extended version of detailed study presented in the main paper.
\cref{tab:supp:trainloss:svhn} and \cref{tab:supp:trainloss:cifar10} compare state-of-the-art AT loss functions against our proposed $\lossdfs$.
The results are consistent with what we have displayed in the main paper, where $\lossdfs$ performs the best in almost all settings.
Although the other loss functions perform generally well in WRN-28-10, they tend to fall into false sense of security with ResNet-20 and ResNet-56, where the seemingly robust models under weak attacks (PGD) easily break under stronger attacks (AutoAttack).
For example, in ResNet-56 of \cref{tab:supp:trainloss:cifar10}, STD~\citep{madry} achieves 46.38\% under PGD, but is easily circumvented by AutoAttack, which gives 0.12\%.
Similar phenomenon is observed across other loss functions.
However, $\lossdfs$ is consistent under both PGD and AutoAttack, and shows no sign of obfuscated gradients.
\begin{table*}[]
    \centering

    \caption{Comparison of $\loss_{Train}$ on SVHN. \HY{no change required}}
            \vspace{1mm}
    \label{tab:supp:trainloss:svhn}
    
    \def\arraystretch{0.8}%
    \resizebox{.88\textwidth}{!}
    {
    \begin{tabular}{lccccccccc}
    \toprule
     & \multicolumn{3}{c}{ResNet-20} & \multicolumn{3}{c}{ResNet-56} & \multicolumn{3}{c}{WRN-28-10} \\
    \cmidrule(lr){2-4}\cmidrule(lr){5-7}\cmidrule(lr){8-10}
       Method &  $\mathcal{A}_{Clean}$ & $\bf\mathcal{A}_{PGD}$ & $\bf\mathcal{A}_{AA}$ & $\mathcal{A}_{Clean}$& $\bf\mathcal{A}_{PGD}$ & $\bf\mathcal{A}_{AA}$ & $\mathcal{A}_{Clean}$& $\bf\mathcal{A}_{PGD}$ & $\bf\mathcal{A}_{AA}$ \\
      \midrule

STD 	&	23.34	&	16.73	&	13.83	&	95.12	&	42.66	&	{\textcolor{white}0}8.73	&	93.71	&	69.32	&	62.58	\\
TRADES  	&	92.99	&	51.13	&	36.71	&	95.73	&	\textbf{67.00}	&	20.87	&	94.12	&	69.10	&	61.75	\\
MART  	&	63.36	&	{\textcolor{white}0}6.48	&	{\textcolor{white}0}1.98	&	91.65	&	26.09	&	{\textcolor{white}0}4.74	&	35.94	&	{\textcolor{white}0}2.55	&	{\textcolor{white}0}1.09	\\
ARD 	&	94.78	&	43.10	&	30.38	&	96.02	&	47.37	&	37.16	&	96.29	&	61.11	&	52.56	\\
RSLAD 	&	93.75	&	44.06	&	29.81	&	94.25	&	56.60	&	48.40	&	96.03	&	64.59	&	57.04	\\
\cmidrule(lr){1-10}																			
\makecell[l]{$\bf{\lossdfs}$\\ \textbf{(Proposed)}}	&	91.78	&	\textbf{54.53}	&	\textbf{45.50 (+8.79)}	&	91.06	&	63.12	&	\textbf{56.54 (+8.14)}	&	94.87	&	\textbf{69.67}	&	\textbf{65.66 (+3.08)}	\\
     \bottomrule
    \end{tabular}
    }
\end{table*}

\begin{table*}[]
    \centering
    
    \caption{Comparison of $\loss_{Train}$ on CIFAR-10. \HY{no change required}}
            \vspace{1mm}
    \label{tab:supp:trainloss:cifar10}
    
    \def\arraystretch{0.8}%
    \resizebox{.88\textwidth}{!}
    {
    \begin{tabular}{lccccccccc}
    \toprule
     & \multicolumn{3}{c}{ResNet-20} & \multicolumn{3}{c}{ResNet-56} & \multicolumn{3}{c}{WRN-28-10} \\
    \cmidrule(lr){2-4}\cmidrule(lr){5-7}\cmidrule(lr){8-10}
       Method &  $\mathcal{A}_{Clean}$ & $\bf\mathcal{A}_{PGD}$ & $\bf\mathcal{A}_{AA}$ & $\mathcal{A}_{Clean}$& $\bf\mathcal{A}_{PGD}$ & $\bf\mathcal{A}_{AA}$ & $\mathcal{A}_{Clean}$& $\bf\mathcal{A}_{PGD}$ & $\bf\mathcal{A}_{AA}$ \\
      \midrule

STD 	&	23.51	&	{\textcolor{white}0}6.09	&	{\textcolor{white}0}1.66	&	92.49	&	\textbf{46.38}	&	{\textcolor{white}0}0.12	&	81.63	&	48.03	&	38.94	\\
TRADES  	&	86.34	&	26.81	&	{\textcolor{white}0}1.75	&	81.71	&	29.49	&	{\textcolor{white}0}9.36	&	79.61	&	45.86	&	37.08	\\
MART  	&	14.91	&	{\textcolor{white}0}2.67	&	{\textcolor{white}0}0.22	&	91.65	&	16.23	&	{\textcolor{white}0}0.00	&	13.69	&	{\textcolor{white}0}6.74	&	{\textcolor{white}0}0.09	\\
ARD 	&	90.13	&	{\textcolor{white}0}9.83	&	{\textcolor{white}0}0.17	&	92.21	&	{\textcolor{white}0}9.31	&	{\textcolor{white}0}2.51	&	90.95	&	36.61	&	31.16	\\
RSLAD 	&	77.85	&	11.66	&	{\textcolor{white}0}0.69	&	88.98	&	19.59	&	12.27	&	90.25	&	39.30	&	31.16	\\
\cmidrule(lr){1-10}																			
\makecell[l]{$\bf{\lossdfs}$\\ \textbf{(Proposed)}}	&	77.83	&	\textbf{27.42}	&	\textbf{19.09 (+17.34)}	&	83.67	&	34.78	&	\textbf{27.69 (+15.42)}	&	88.16	&	\textbf{50.13}	&	\textbf{41.40 (+2.46)}	\\

     \bottomrule
    \end{tabular}
    }
\end{table*}

For comparison, we present real-data training performance on the \medmnist{} dataset in \cref{tab:medmnist_only_original}.
The `original' data training uses the exact same domain for adversarial training, so that can be regarded as the upper bound of the data-free adversarial robustness. 
The experimental results show that even real data from another domain (CIFAR-10) significantly underperform compared to the original dataset. 
On the other hand, \name shows superior performance than the other-domain public dataset.
Remarkably, \name almost reached similar performance levels with the original dataset training in the Derma dataset.  
The experimental results show the advantages of \name, by reducing the gap towards real-data training.

\begin{table}[h]
\centering
    \caption{Real-data training performance of \medmnist with $l_\infty$ perturbation budget.}
    \vspace{1mm}
    \def\arraystretch{0.88}%
\resizebox{.75\columnwidth}{!}{

\begin{tabular}{cclcccccc}
\midrule
\multirow{3}{*}{Dataset} & \multirow{3}{*}{Data-free} &\multirow{3}{*}{Method}                                                                                                       
                         & \multicolumn{3}{c}{ResNet-18} & \multicolumn{3}{c}{ResNet-50}  \\
                         \cmidrule(lr){4-6}  \cmidrule(lr){7-9}
                         &&& $\mathcal{A}_{Clean}$ & $\bf\mathcal{A}_{PGD}$ & $\bf\mathcal{A}_{AA}$      &      $\mathcal{A}_{Clean}$ & $\bf\mathcal{A}_{PGD}$ & $\bf\mathcal{A}_{AA}$              \\ 
\midrule
 \multirowcell{3}{Tissue} 
 &  \multirowcell{2}{\xmark} & Original & 50.33	&37.53	&34.25	&48.94	&37.34&	35.06 \\
 &   & Public (CIFAR-10) &	22.04	&	{\textcolor{white}0}0.02	&	{\textcolor{white}0}0.00	&	27.84	&	10.11	&	{\textcolor{white}0}8.64	\\	
 \cmidrule{2-9}
& {\cmark} & \textbf{\name}   &32.07	&31.63	&31.57	&31.91	&27.15	&26.68	\\	
\midrule
\multirowcell{3}{Blood}  
 &  \multirowcell{2}{\xmark} & Original &86.73 &	71.94 &	70.36	 &85.41 &	71.79 &	70.51 \\
&  & Public (CIFAR-10)  &	{\textcolor{white}0}9.09	&	{\textcolor{white}0}9.09	&	{\textcolor{white}0}0.00	&	{\textcolor{white}0}9.09	&	{\textcolor{white}0}9.09	&	{\textcolor{white}0}0.00	\\ \cmidrule{2-9}
&{\cmark}& \textbf{\name}  &	59.89&	21.72	&19.29&	74.63	&36.07	&30.17	\\	
\midrule
\multirowcell{3}{Path} 
 &  \multirowcell{2}{\xmark} & Original & 74.37	&52.53	&49.35&	72.62	&51.73&	48.68\\
 &&Public (CIFAR-10) &		13.30&	{\textcolor{white}0}0.00	&{\textcolor{white}0}0.00	&{\textcolor{white}0}7.54&	{\textcolor{white}0}1.21&	{\textcolor{white}0}0.37\\	\cmidrule{2-9}
&{\cmark}& \textbf{\name}   &	33.06	&29.78	&25.38&	41.63&	15.35&	12.28\\		
\midrule
\multirowcell{3}{Derma} 
 &  \multirowcell{2}{\xmark} & Original & 66.90 & 63.90 & 63.01 & 67.58 & 61.99 &  60.14\\
 &&Public (CIFAR-10) &	67.48	&50.47	&42.39	&67.08	&44.59&	34.11\\	\cmidrule{2-9}
&{\cmark}& \textbf{\name}   &66.98	&63.09	&61.85	&67.78	&64.34	&58.05	\\		
\midrule
\multirowcell{3}{OrganA} 
 &  \multirowcell{2}{\xmark} & Original & 92.79&	81.20&	80.75&	92.67&	82.73	&82.25\\
 &&Public (CIFAR-10) &	86.79	&33.82&	30.28	&78.77	&33.56	&28.78	\\	\cmidrule{2-9}
&{\cmark}& \textbf{\name}   &	86.44	&49.06&	44.90&	90.80	&42.58&	37.42\\	
\midrule
\multirowcell{3}{OrganC}   
 &  \multirowcell{2}{\xmark} & Original &91.13	&81.06	&80.55	&90.52	&81.47	&80.94\\
&&Public (CIFAR-10) &	79.41	&	40.10	&	36.53	&	84.41	&	46.12	&	43.44	\\	\cmidrule{2-9}
&{\cmark}& \textbf{\name}  &	83.35 &	47.01	 &42.56	 &86.56 &	62.60 &	59.86\\	
\midrule
\multirowcell{3}{OrganS} 
 &  \multirowcell{2}{\xmark} & Original & 79.01	 &62.62 &	61.95 &	79.08 &	65.16	 &64.57
\\
 &&Public (CIFAR-10) &	68.82 &	18.25	 &15.41 &	47.70 &	12.68 &	{\textcolor{white}0}4.28\\	\cmidrule{2-9}
&{\cmark}& \textbf{\name}   &	64.43	 &29.68	 &24.31 &	66.61 &	37.84	 &33.63\\	
\bottomrule
\end{tabular}
}
    \label{tab:medmnist_only_original}

\end{table}

Similarly, for dataset diversification, we show an extended version in \cref{tab:supp:diversity:svhn} and \cref{tab:supp:diversity:cifar10}.
In all settings, \mix shows the best quantitative measure under Coverage and JSD.
Coverage is known to be a more accurate measure of diversity than Recall in the sense that it is more robust against outliers~\citep{gandiversitymetric}.
Also, JSD measures distributional distance, which is frequently used in evaluating GANs.
Thus, they show quantitative evidence to diversity gain of \mix.
This aligns with the robust training results, where \mix outperforms other diversifying methods in most settings.

    \begin{table}[h]    
    \centering
    \caption{Comparison of dataset diversification methods on SVHN.}
            \vspace{1mm}
    \label{tab:supp:diversity:svhn}
    
    \def\arraystretch{0.89}%
    \resizebox{.8\columnwidth}{!}
    {
    \begin{tabular}{clcclcccc}
    \toprule
     & & \multicolumn{3}{c}{Accuracy} & \multicolumn{4}{c}{Diversity Metric} \\
    \cmidrule(lr){3-5}\cmidrule(lr){6-9}
       Model & Method &  $\mathcal{A}_{Clean}$& $\bf\mathcal{A}_{PGD}$ & $\bf\mathcal{A}_{AA}$  &\multicolumn{1}{c}{Recall $\uparrow$} & \multicolumn{1}{c}{Coverage $\uparrow$} & \multicolumn{1}{c}{NDB $\downarrow$} & \multicolumn{1}{c}{JSD $\downarrow$}\\
      \midrule

 \multirowcell{5}{ResNet-20}
 &  $\mathcal{L}_{Synth}$ &	93.31	&	54.11	&	41.03	&	0.801	&	0.230	&	95	&	0.353	\\
& + Mixup& 	92.13	&	\textbf{57.71}	&	48.17 (+7.14)	&	0.882	&	0.241	&	\textbf{88}	&	0.368	\\
& + Cutout& 	91.34	&	56.01	&	\textbf{48.29 (+7.26)}	&	0.900	&	0.198	&	90	&	0.396	\\
& + CutMix & 	92.06	&	56.79	&	48.14 (+7.11)	&	0.887	&	0.225	&	91	&	0.387	\\
\cmidrule{2-9}														
& \textbf{+ \mixabb} & 	91.78	&	54.53	&	45.50 (+4.47)	&	\textbf{0.905}	&	\textbf{0.429}	&	90	&	\textbf{0.237}	\\

\midrule
 \multirowcell{5}{ResNet-56}
&  $\mathcal{L}_{Synth}$ &	93.17	&	61.40	&	54.38	&	0.821	&	0.218	&	\textbf{93}	&	0.342	\\
& + Mixup& 	92.23	&	62.26	&	55.11 (+0.73)	&	0.848	&	0.226	&	\textbf{93}	&	0.345	\\
& + Cutout  & 	93.92	&	60.54	&	53.80 (-0.58)	&	0.842	&	0.164	&	95	&	0.391	\\
& + CutMix & 	91.20	&	61.46	&	55.38 (-1.00)	&	0.871	&	0.189	&	95	&	0.369	\\
\cmidrule{2-9}														
& \textbf{+ \mixabb} & 	91.06	&	\textbf{63.12}	&	\textbf{56.54 (+2.16)}	&	\textbf{0.872}	&	\textbf{0.521}	&	\textbf{93}	&	\textbf{0.154}	\\
\midrule
 \multirowcell{5}{WRN-28-10}
&  $\mathcal{L}_{Synth}$ &	94.26	&	64.94	&	59.99	&	0.246	&	0.147	&	91	&	0.254	\\
& + Mixup& 	94.50	&	67.51	&	54.70 (-5.29)	&	0.252	&	0.120	&	94	&	0.277	\\
& + Cutout  & 	95.51	&	66.77	&	61.96 (+1.97)	&	0.305	&	0.060	&	91	&	0.332	\\
& + CutMix & 	95.67	&	66.71	&	61.16 (+1.17)	&	0.321	&	0.100	&	92	&	0.348	\\
\cmidrule{2-9}														
& \textbf{+ \mixabb} & 	94.87	&	\textbf{69.67}	&	\textbf{65.66 (+5.67)}	&	\textbf{0.548}	&	\textbf{0.232}	&	\textbf{88}	&	\textbf{0.190}	\\
     \bottomrule
    \end{tabular}
    }
\end{table}

    \begin{table}[h]    
    \centering
    \caption{Comparison of dataset diversification methods on CIFAR-10.}
            \vspace{1mm}
    \label{tab:supp:diversity:cifar10}
    
    \def\arraystretch{0.89}%
    \resizebox{.8\columnwidth}{!}
    {
    \begin{tabular}{clcclcccc}
    \toprule
     & & \multicolumn{3}{c}{Accuracy} & \multicolumn{4}{c}{Diversity Metric} \\
    \cmidrule(lr){3-5}\cmidrule(lr){6-9}
       Model & Method &  $\mathcal{A}_{Clean}$& $\bf\mathcal{A}_{PGD}$ & $\bf\mathcal{A}_{AA}$  &\multicolumn{1}{c}{Recall $\uparrow$} & \multicolumn{1}{c}{Coverage $\uparrow$} & \multicolumn{1}{c}{NDB $\downarrow$} & \multicolumn{1}{c}{JSD $\downarrow$}\\
      \midrule

 \multirowcell{5}{ResNet-20}
	&  $\mathcal{L}_{Synth}$  &	82.58	&	23.93	&	14.61	&	0.400	&	0.107	&	88	&	0.355	\\
	& + Mixup & 	84.26	&	16.91	&	{\color{white}0}5.95 (-8.66)	&	0.692	&	0.128	&	\textbf{87}	&	0.372	\\
	& + Cutout  & 	82.65	&	26.33	&	17.32 (+2.71)	&	0.747	&	0.137	&	95	&	0.369	\\
	& + CutMix& 	83.38	&	\textbf{28.66}	&	18.30 (+3.69)	&	\textbf{0.825}	&	0.175	&	89	&	0.347	\\
	\cmidrule{2-9}														
	& \textbf{+ \mixabb}& 	77.83	&	27.42	&	\textbf{19.09 (+4.48)}	&	0.724	&	\textbf{0.320}	&	90	&	\textbf{0.248}	\\
\midrule
 \multirowcell{5}{ResNet-56}
&  $\mathcal{L}_{Synth}$  &	83.72	&	30.91	&	27.42	&	0.658	&	0.136	&	93	&	0.310	\\
& + Mixup& 	83.55	&	32.87	&	\textbf{27.87 (+0.45)}	&	0.761	&	0.135	&	93	&	0.394	\\
& + Cutout  & 	82.96	&	31.39	&	26.83 (-0.59)	&	0.853	&	0.113	&	94	&	0.343	\\
& + CutMix & 	82.60	&	33.78	&	27.86 (+0.44)	&	\textbf{0.892}	&	0.150	&	93	&	0.364	\\
\cmidrule{2-9}														
& \textbf{+ \mixabb}& 	83.67	&	\textbf{34.78}	&	27.69 (+0.27)	&	0.678	&	\textbf{0.550}	&	\textbf{84}	&	\textbf{0.126}	\\
\midrule
 \multirowcell{5}{WRN-28-10}
&  $\mathcal{L}_{Synth}$  &	91.46	&	43.66	&	36.34	&	0.535	&	0.101	&	91	&	0.253	\\
& + Mixup& 	90.61	&	48.16	&	36.43 (+0.09)	&	0.641	&	0.084	&	94	&	0.322	\\
& + Cutout  & 	92.59	&	39.84	&	34.39 (-1.95)	&	0.535	&	0.034	&	95	&	0.443	\\
& + CutMix & 	91.90	&	42.79	&	34.79 (-1.55)	&	\textbf{0.845}	&	0.084	&	93	&	0.328	\\
\cmidrule{2-9}														
& \textbf{+ \mixabb}& 	88.16	&	\textbf{50.13}	&	\textbf{41.40 (+5.06)}	&	0.830	&	\textbf{0.163}	&	\textbf{88}	&	\textbf{0.211}	\\
     \bottomrule
    \end{tabular}
    }
\end{table}

\section{Evaluation under Modified Attacks}
\label{sec:supp:modified_attack}
\setlength{\tabcolsep}{4pt}
\begin{table*}[h]
    \centering
    
    \caption{Evaluation under modified attacks on SVHN and CIFAR-10.}
            \vspace{1mm}
    \label{tab:supp:modified_attack}
    
    \def\arraystretch{0.99}%
    \resizebox{.8\textwidth}{!}
    {
\begin{tabular}{ccccccccc}
\toprule
\multirowcell{2}{Dataset} & \multirowcell{2}{Model}
&&&&\multicolumn{4}{c}{Modified Attack} \\
\cmidrule{6-9}
&&$\mathcal{A}_{Clean}$ & $\mathcal{A}_{PGD_{CE}}$ & $\mathcal{A}_{AA}$ & $\mathcal{A}_{Latent}$ & $\mathcal{A}_{PGD_{(a)}}$ & $\mathcal{A}_{PGD_{(b)}}$ & $\mathcal{A}_{PGD_{(c)}}$ \\
\cmidrule{1-9}

\multirowcell{3}{SVHN}    & ResNet-20                     & 91.83                      & 54.82           & 47.55      & 74.38           & 71.95               & 55.84               & 55.24                                  \\ 
\multicolumn{1}{c}{}        & ResNet-56                     & 88.66                      & 62.05           & 57.54      & 77.99           & 77.04               & 62.95               & 62.58                                  \\ 
 & \multicolumn{1}{l}{WRN-28-10} & 94.14                      & 69.60           & 62.66      & 87.68           & 81.39               & 70.64               & 70.08                                  \\ 
 \cmidrule{1-9}
\multirowcell{3}{CIFAR-10}                    & ResNet-20                     & 74.79                      & 29.29           & 22.65      & 65.63           & 54.64               & 31.05               & 30.46                                  \\ 
                              & ResNet-56                     & 81.30                      & 35.55           & 30.51      & 67.73           & 60.61               & 36.94               & 36.43                                  \\ 
                              & \multicolumn{1}{l}{WRN-28-10} & 86.74                      & 51.13           & 43.73      & 79.38           & 70.40               & 51.82               & 51.21                                  \\ 
                                 \bottomrule

\end{tabular}
    }
\end{table*}

\begin{table*}[h]
    \centering
    
    \caption{Evaluation under modified attacks on \medmnist{}.}
            \vspace{1mm}
    \label{tab:supp:modified_attack_medmnist}
    
    \def\arraystretch{0.99}%
    \resizebox{.8\textwidth}{!}
    {
\begin{tabular}{ccccccccc}
\toprule
\multirowcell{2}{Dataset} 
& \multirowcell{2}{Model}
&&&&\multicolumn{4}{c}{Modified Attack} \\
\cmidrule{6-9}
&&$\mathcal{A}_{Clean}$ & $\mathcal{A}_{PGD_{CE}}$ & $\mathcal{A}_{AA}$ & $\mathcal{A}_{Latent}$ & $\mathcal{A}_{PGD_{(a)}}$ & $\mathcal{A}_{PGD_{(b)}}$ & $\mathcal{A}_{PGD_{(c)}}$ \\
\cmidrule{1-9}
Tissue   &  \multirowcell{4}{ResNet-18}          &  32.07 & 31.93           & 31.83      & 32.07       & 31.98                 & 31.95                 & 31.95                                     \\ 
Blood      &&  49.34 & 19.24           & 18.77      & 20.32       & 29.87                 & 24.70                 & 24.12                                      \\ 
Path && 33.06&	29.78&	25.38& 29.81& 32.10& 30.04& 29.99\\
OrganC &  &76.89& 46.92           &45.18      & 76.60       & 65.47                 & 48.40                 & 48.10                                   \\ 

 \bottomrule

\end{tabular}
    }
\end{table*}

Evaluating robust accuracy using PGD~\cite{madry} and AutoAttack~\cite{autoattack} are considered de facto standard to demonstrate the method’s robustness.
However, we extend our experiments and provide further evaluation under latent attack~\cite{latentattack} and using different combinations of our training loss $\lossdfs$ as the inner maximization of PGD.
We term these attacks \textit{modified attacks} because they are modified according to the training objective or takes advantage of model architecture to generate adversarial samples, similar to adaptive attacks~\cite{adaptiveauto, automated}.
For modified versions of PGD, each replaces the coventionally used cross entropy loss $CE(S(x'),y)$ with: (a) $KL(S(x')\|S(x))$,  (b) $KL(S(x')\|T(x)$, (c) $KL(S(x')\|T(x)) + KL(S(x')\|S(x))$.
For latent attack~\cite{latentattack}, we followed the original implementation and used output from the penultimate layer (before flattening), L-BFGS for attack optimization with $\eps$=10/255 for perturbation bound. 
The results are shown in \cref{tab:supp:modified_attack} and \cref{tab:supp:modified_attack_medmnist}, where our method \name is effective against modified attacks as well. 
In all datasets and models, none of the modified attack methods were stronger than cross entropy based PGD and AutoAttack.

\section{Further Visualization of Loss Surface}
\label{sec:supp:trainloss}
\begin{figure}
\captionsetup[subfigure]{justification=centering}
    \centering
    \begin{subfigure}{0.24\columnwidth}
    \includegraphics[width=\textwidth]{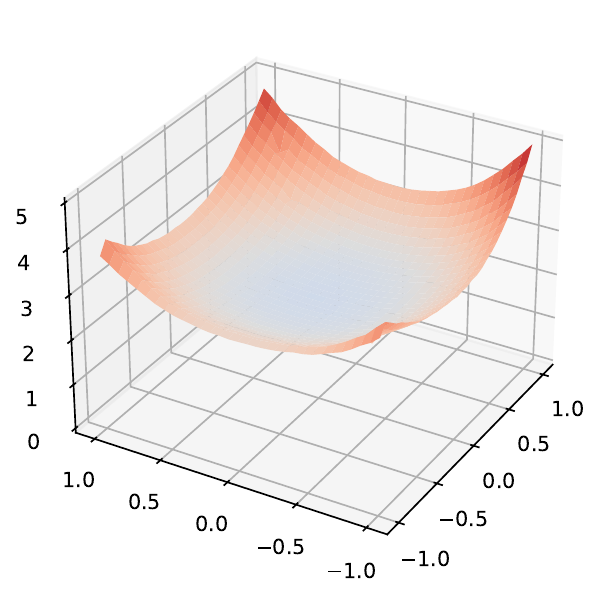}
    \caption{TRADES \\ w/o \gradmask}
    \label{fig:lossvis:trades_nomask}
    \end{subfigure} 
    \begin{subfigure}{0.24\columnwidth}
    \includegraphics[width=\textwidth]{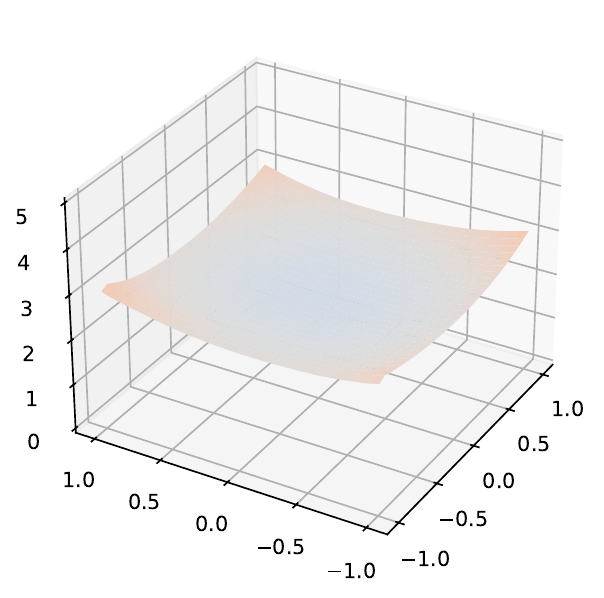}
    \caption{TRADES \\ w/ \gradmask}
    \label{fig:lossvis:trades_mask}
    \end{subfigure} 
    \begin{subfigure}{0.24\columnwidth}
    \includegraphics[width=\textwidth]{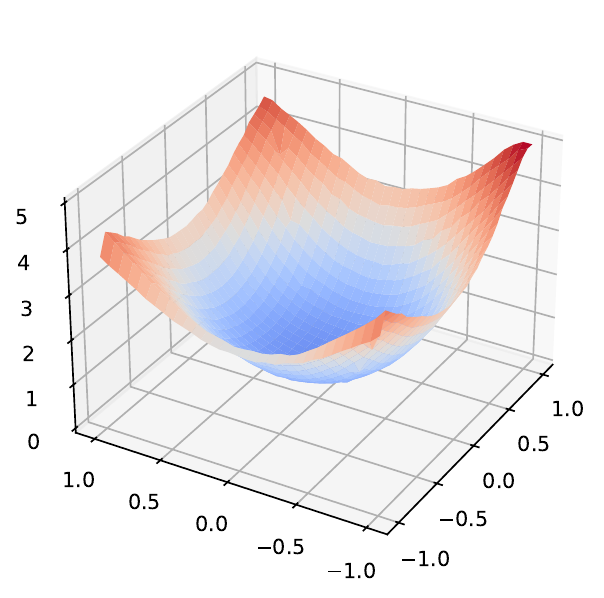}
    \caption{$\lossdfs$ \\ w/o \gradmask}
    \label{fig:lossvis:dfs_nomask}
    \end{subfigure} 
    \begin{subfigure}{0.24\columnwidth}
    \includegraphics[width=\textwidth]{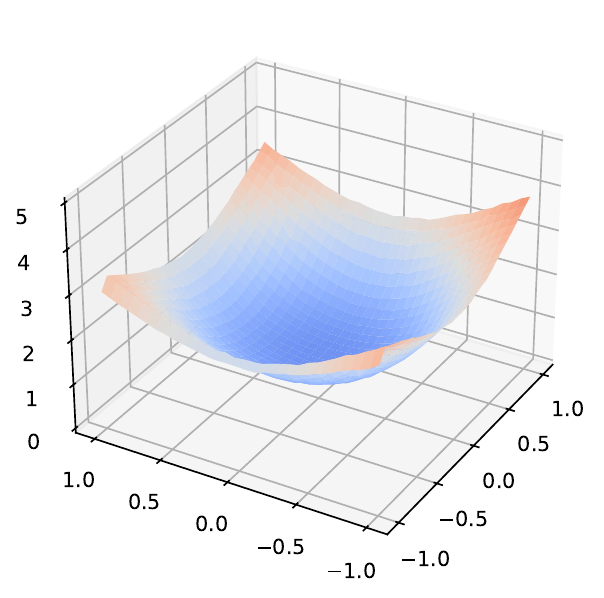}
    \caption{$\lossdfs$ \\ w/ \gradmask}
    \label{fig:lossvis:dfs_mask}
    \end{subfigure} 
    \caption{Loss surface visualization of ResNet56 model trained by data-free AT methods. Each figure represents different training losses with or without \gradmask. We use normalized random direction for $x$,$y$ axis, following \citet{li2018visualizing}. The figures demonstrate that \gradmask achieves flatter loss surfaces.}
    \label{fig:loss_surf_vis_r56}
\end{figure}

\begin{figure}
    \centering
    \begin{subfigure}{0.24\columnwidth}
    \includegraphics[width=\textwidth]{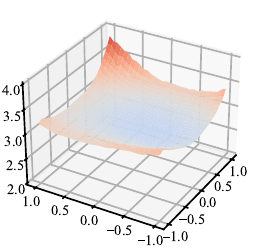}
    \caption{TRADES \\ w/o \gradmask}
    \label{fig:lossvis:trades_nomask}
    \end{subfigure} 
    \begin{subfigure}{0.24\columnwidth}
    \includegraphics[width=\textwidth]{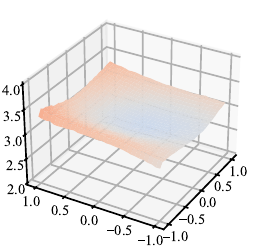}
    \caption{TRADES \\ w/ \gradmask}
    \label{fig:lossvis:trades_mask}
    \end{subfigure} 
    \begin{subfigure}{0.24\columnwidth}
    \includegraphics[width=\textwidth]{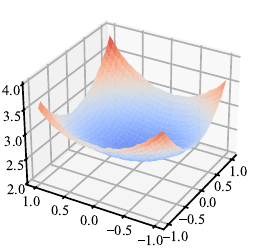}
    \caption{$\lossdfs$ \\ w/o \gradmask}
    \label{fig:lossvis:dfs_nomask}
    \end{subfigure} 
    \begin{subfigure}{0.24\columnwidth}
    \includegraphics[width=\textwidth]{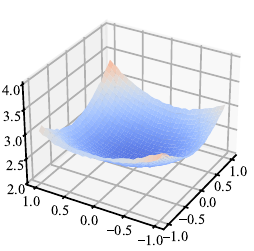}
    \caption{$\lossdfs$ \\ w/ \gradmask}
    \label{fig:lossvis:dfs_mask}
    \end{subfigure} 
    
    \caption{Loss surface visualization of WRN-28-10 model trained by data-free AT methods. Each figure represents different training losses with or without \gradmask. We use normalized random direction for $x$,$y$ axis, following \citet{li2018visualizing}. The figures demonstrate that \gradmask achieves flatter loss surfaces.}
    \label{fig:loss_surf_vis_wrn}
\end{figure}

We extend \cref{fig:loss_surf_vis} to different models, ResNet-56 and WRN-28-10.
The visualization results are shown in \cref{fig:loss_surf_vis_r56,fig:loss_surf_vis_wrn}.
In all visualization settings, applying \gradmask to data-free adversarial training achieves a flatter loss surface. 
This analysis further supports the experimental results that \gradmask contributes to better performance.

\section{Sensitivity Study on the Number of Aggregated Batches}
\label{sec:supp:batch_sensitivity}

    \begin{table*}[t]    

  \centering
    \caption{Sensitivity study on aggregated batch number using CIFAR-10 dataset.}
            \vspace{1mm}
    \label{tab:batch_sensitivity}
    \def\arraystretch{0.9}%
    \resizebox{0.5\textwidth}{!}
    {
    \begin{tabular}{ccccccc}
    \toprule
     \multirow{2}{*}{$\mathcal{B}$} &  \multicolumn{3}{c}{ResNet-20} & \multicolumn{3}{c}{WRN-28-10} \\
    \cmidrule(lr){2-4}\cmidrule(lr){5-7}
         & $\mathcal{A}_{Clean}$ & $\bf\mathcal{A}_{PGD}$ & $\bf\mathcal{A}_{AA}$ & $\mathcal{A}_{Clean}$& $\bf\mathcal{A}_{PGD}$ & $\bf\mathcal{A}_{AA}$ \\
      \midrule

1	&77.83	&	27.42	&	19.09	&	88.16	&	50.13	&	41.40	\\
2	&75.77	&	29.16    & 22.44&	88.07	&	50.50	&	41.96	\\
4	&	74.74	&	29.19 & 22.94	&	87.85	&	50.36	&	42.10	\\
8	&	75.01	&	29.47	&	23.09	&	87.67	&	50.53	&	41.80	\\
10	&		75.53	&	29.69	& 22.95	&	87.65	&	50.75	&	42.35	\\
20	&	74.63	&	29.28	& 22.63&	86.74	&	51.13	&	43.73	\\
40	&	28.87	&	13.72 & 10.35	&	85.48	&	50.39	&	44.43	\\

     \bottomrule
    \end{tabular}
    }

\end{table*}

In this section, we show a sensitivity study on the number of aggregated batches when applying \gradmask.
\cref{tab:batch_sensitivity} shows the performance under varying number of aggregated batches ($\mathcal{B}$) during training.
Aggregated batch being 1 means \gradmask was not applied.
For both models, we can observe that the performance is relatively stable for a wide range of $\mathcal{B}$.
Also, a smaller model displays slightly higher sensitivity towards $\mathcal{B}$, while a larger model is less affected by it.
We found $\mathcal{B}=\{10,20\}$ to work generally well across different datasets and models.

\section{Sensitivity Study on $\tau$}
\label{sec:supp:tau_sensitivity}
\begin{table}[h]
    \centering
    
    \caption{Sensitivity study on $\tau$ using WRN-28-10.}
            \vspace{1mm}
    \label{tab:supp:tau_sensitivity}
    
    \def\arraystretch{0.9}%
    \resizebox{.5\columnwidth}{!}
    {
    \begin{tabular}{ccccccc}
    \toprule
     &  \multicolumn{3}{c}{SVHN} & \multicolumn{3}{c}{CIFAR-10} \\
    \cmidrule(lr){2-4}\cmidrule(lr){5-7}
       $\tau$ &   $\mathcal{A}_{Clean}$& $\bf\mathcal{A}_{PGD}$ & $\bf\mathcal{A}_{AA}$ & $\mathcal{A}_{Clean}$& $\bf\mathcal{A}_{PGD}$ & $\bf\mathcal{A}_{AA}$ \\
      \midrule

0.0 &	93.69	&	69.05	&	61.99	&	88.16	&	50.13	&	41.40	\\
0.1 &	93.65	&	68.78	&	61.98	&	86.94	&	50.29	&	42.17	\\
0.2 &	93.66	&	68.83	&	62.96	&	86.95	&	50.12	&	42.06	\\
0.3 &	93.86	&	69.13	&	63.13	&	87.53	&	50.63	&	42.93	\\
0.4 &	94.05	&	69.12	&	63.17	&	87.13	&	50.99	&	43.34	\\
0.5 &	94.14	&	69.60	&	62.66	&	86.74	&	51.13	&	43.73	\\
0.6 &	94.38	&	70.09	&	63.52	&	87.17	&	51.58	&	43.62	\\
0.7 &	94.69	&	70.13	&	62.77	&	87.10	&	51.96	&	43.84	\\
0.8 &	95.00	&	69.90	&	61.85	&	86.83	&	51.87	&	43.10	\\
0.9 &	95.01	&	69.72	&	61.37	&	86.82	&	50.67	&	41.38	\\
1.0 &	95.42	&	69.10	&	59.45	&	88.10	&	49.11	&	39.12	\\

\bottomrule
    \end{tabular}
    }
\end{table}
In this section, we provide sensitivity study results on the threshold value $\tau$ used in \gradmask, which is displayed in \cref{tab:supp:tau_sensitivity}.
In general, the best range for $\tau$ lies in $[0.4,0.7]$, 
with more degradation when being close to 0.0 and 1.0. 
In all settings, we simply use $\tau=0.5$ for the difference within the best range is insignificant.
Note that in our method, $\tau=0.0$ means no masking is used (allow conflicting gradients), and $\tau=1.0$ means all gradients are masked unless they all have the same sign direction. 
Intuitively, values close to 0.0 diminish the effect of masking, and values close to 1.0 set an unrealistic high bar for gradients, where both should have a negative effect on the final performance.

\section{Sensitivity Study on $\lambda_1$ and $\lambda_2$}
\label{sec:supp:lambda_sensitivity}
\begin{table}[h]
    \centering
    
    \caption{Sensitivity study on $\lambda_1$ and $\lambda_2$ using SVHN dataset.}
            \vspace{1mm}
    \label{tab:supp:lambda_sensitivity_svhn}
    
    \def\arraystretch{0.95}%
    \resizebox{.9\columnwidth}{!}
    {
    \begin{tabular}{cccccccc}
    \toprule
     & \multicolumn{1}{r}{} &  \multicolumn{6}{c}{$\lambda_2$}  \\

           && 0 & 0.01 & 0.1&1 &10&100 \\
                \cmidrule(lr){3-3}      \cmidrule(lr){4-4} \cmidrule(lr){5-5} \cmidrule(lr){6-6} \cmidrule(lr){7-7} \cmidrule(lr){8-8}
            Model  & $\lambda_1$ &   $\mathcal{A}_{AA}$($\bf\mathcal{A}_{Clean}$) & $\mathcal{A}_{AA}$($\bf\mathcal{A}_{Clean}$)& $\mathcal{A}_{AA}$($\bf\mathcal{A}_{Clean}$) & $\mathcal{A}_{AA}$($\bf\mathcal{A}_{Clean}$)& $\mathcal{A}_{AA}$($\bf\mathcal{A}_{Clean}$)&  $\mathcal{A}_{AA}$($\bf\mathcal{A}_{Clean}$)\\
      \midrule

\multirowcell{6}{ResNet-20} 
&   0&	0.04 (96.59)	&	0.11 (96.55)	&	45.16 (93.10)	&	44.55 (72.21)	&	19.59 (19.59)	&	6.70 (6.70)	\\
& 0.01&	0.14 (96.64)	&	1.23 (96.70)	&	44.36 (93.52)	&	47.99 (78.43)	&	19.59 (19.59)	&	6.70 (6.70)	\\
& 0.1&	40.69 (93.12)	&	40.58 (93.63)	&	43.82 (93.20)	&	51.50 (88.24)	&	19.59 (19.59)	&	6.70 (6.70)	\\
& 1& 	42.95 (93.75)	&	42.80 (93.69)	&	43.25 (93.78)	&	47.55 (91.83)	&	19.56 (19.62)	&	6.70 (6.70)	\\
& 10& 	42.63 (87.56)	&	42.07 (86.39)	&	43.34 (88.56)	&	45.74 (86.45)	&	38.78 (80.30)	&	6.70 (6.70)	\\
& 100& 	6.70 (6.70)	&	6.70 (6.70)	&	6.70 (6.70)	&	6.70 (6.70)	&	6.70 (6.70)	&	6.70 (6.70)	\\
\midrule  

\multirowcell{6}{ResNet-56} 
&   0&	0.18 (97.22)	&	2.31 (97.31)	&	48.79 (93.58)	&	52.53 (73.34)	&	19.59 (19.59)	&	6.70 (6.70)	\\
& 0.01&	3.36 (97.28)	&	11.21 (97.08)	&	49.17 (93.71)	&	54.71 (76.90)	&	19.63 (19.69)	&	6.70 (6.70)	\\
& 0.1&	43.14 (95.04)	&	45.01 (94.22)	&	51.97 (93.72)	&	58.01 (84.51)	&	22.55 (25.50)	&	6.70 (6.70)	\\
& 1& 	55.56 (91.94)	&	55.59 (91.96)	&	56.01 (91.55)	&	57.54 (88.66)	&	50.97 (76.66)	&	6.70 (6.70)	\\
& 10& 	53.22 (85.85)	&	53.44 (86.15)	&	53.57 (85.57)	&	53.36 (84.65)	&	51.06 (84.01)	&	10.86 (14.79)	\\
& 100& 	37.96 (72.76)	&	33.70 (72.88)	&	36.11 (73.93)	&	37.28 (75.71)	&	38.67 (70.18)	&	6.70 (6.70)	\\

\midrule  
\multirowcell{6}{WRN-28-10} 
&   0&	0.22 (97.39)	&	0.61 (97.43)	&	60.42 (95.49)	&	56.30 (83.42)	&	21.51 (25.21)	&	19.59 (19.59)	\\
& 0.01&	0.69 (97.44)	&	1.38 (97.45)	&	60.56 (95.53)	&	59.14 (86.00)	&	22.12 (28.83)	&	19.59 (19.59)	\\
& 0.1&	27.62 (96.37)	&	41.93 (96.27)	&	61.03 (95.59)	&	63.53 (91.75)	&	40.85 (58.79)	&	19.59 (19.59)	\\
& 1& 	60.72 (95.41)	&	60.74 (95.39)	&	61.16 (95.21)	&	62.66 (94.14)	&	58.56 (86.35)	&	19.59 (19.59)	\\
& 10& 	60.15 (92.45)	&	60.05 (92.52)	&	60.13 (92.59)	&	60.00 (92.09)	&	57.75 (88.38)	&	38.88 (67.03)	\\
& 100& 	44.95 (81.76)	&	46.19 (83.62)	&	44.45 (81.28)	&	46.14 (82.94)	&	45.38 (82.58)	&	43.35 (76.67)	\\

\bottomrule
    \end{tabular}
    }
\end{table}
\begin{table}[h]
    \centering
    
    \caption{Sensitivity study on $\lambda_1$ and $\lambda_2$ using CIFAR-10 dataset.}
            \vspace{1mm}
    \label{tab:supp:lambda_sensitivity_cifar}
    
    \def\arraystretch{0.95}%
    \resizebox{.9\columnwidth}{!}
    {
    \begin{tabular}{cccccccc}
    \toprule
     & \multicolumn{1}{r}{} &  \multicolumn{6}{c}{$\lambda_2$}  \\

           && 0 & 0.01 & 0.1&1 &10&100 \\
                \cmidrule(lr){3-3}      \cmidrule(lr){4-4} \cmidrule(lr){5-5} \cmidrule(lr){6-6} \cmidrule(lr){7-7} \cmidrule(lr){8-8}
            Model  & $\lambda_1$ &   $\mathcal{A}_{AA}$($\bf\mathcal{A}_{Clean}$) & $\mathcal{A}_{AA}$($\bf\mathcal{A}_{Clean}$)& $\mathcal{A}_{AA}$($\bf\mathcal{A}_{Clean}$) & $\mathcal{A}_{AA}$($\bf\mathcal{A}_{Clean}$)& $\mathcal{A}_{AA}$($\bf\mathcal{A}_{Clean}$)&  $\mathcal{A}_{AA}$($\bf\mathcal{A}_{Clean}$)\\
      \midrule

\multirowcell{6}{ResNet-20} 
&   0&	0.00 (93.99)	&	0.00 (93.79)	&	18.06 (82.92)	&	20.31 (67.61)	&	9.26 (13.16)	&	10.00 (10.00)	\\
& 0.01&	0.00 (93.62)	&	0.00 (93.46)	&	18.84 (82.47)	&	21.18 (68.99)	&	10.43 (13.05)	&	10.00 (10.00)	\\
& 0.1&	10.36 (86.44)	&	12.32 (85.20)	&	20.72 (80.33)	&	24.20 (72.44)	&	11.45 (14.84)	&	10.00 (10.00)	\\
& 1& 	20.83 (78.49)	&	20.69 (78.59)	&	21.27 (78.24)	&	22.65 (74.79)	&	12.53 (15.75)	&	10.00 (10.00)	\\
& 10& 	15.74 (63.26)	&	15.82 (61.62)	&	14.59 (61.58)	&	14.20 (53.02)	&	12.31 (36.82)	&	10.00 (10.00)	\\
& 100& 	10.00 (10.00)	&	10.00 (10.00)	&	10.00 (10.00)	&	10.00 (10.00)	&	10.00 (10.00)	&	10.00 (10.00)	\\
\midrule  

\multirowcell{6}{ResNet-56} 
&   0&	0.00 (95.44)	&	0.00 (94.67)	&	24.81 (87.04)	&	32.09 (78.33)	&	11.44 (16.22)	&	10.00 (10.00)	\\
& 0.01&	0.00 (94.70)	&	0.00 (94.81)	&	25.31 (86.78)	&	32.62 (78.89)	&	11.04 (16.13)	&	9.65 (11.35)	\\
& 0.1&	19.83 (88.73)	&	21.18 (88.23)	&	25.65 (85.71)	&	33.08 (79.65)	&	9.93 (23.06)	&	9.00 (14.57)	\\
& 1& 	26.71 (84.31)	&	26.96 (84.43)	&	27.44 (83.92)	&	30.51 (81.30)	&	24.07 (58.25)	&	15.49 (16.49)	\\
& 10& 	26.01 (74.59)	&	26.22 (74.67)	&	25.50 (74.72)	&	25.87 (72.84)	&	22.91 (61.85)	&	10.00 (10.00)	\\
& 100& 	10.00 (10.00)	&	10.00 (10.00)	&	10.00 (10.00)	&	10.00 (10.00)	&	10.00 (10.00)	&	10.00 (10.00)	\\

\midrule  
\multirowcell{6}{WRN-28-10} 
&   0&	0.00 (97.48)	&	0.00 (97.34)	&	33.06 (92.61)	&	45.08 (82.11)	&	12.99 (16.15)	&	7.93 (11.38)	\\
& 0.01&	0.00 (97.35)	&	0.00 (97.27)	&	34.29 (92.13)	&	45.16 (82.75)	&	13.22 (17.05)	&	10.00 (10.00)	\\
& 0.1&	28.13 (92.89)	&	30.14 (92.52)	&	38.44 (91.27)	&	44.70 (84.01)	&	29.47 (54.82)	&	10.00 (10.00)	\\
& 1& 	40.46 (89.47)	&	40.43 (89.45)	&	41.05 (89.26)	&	43.73 (86.74)	&	39.47 (71.48)	&	14.91 (20.84)	\\
& 10& 	41.47 (86.28)	&	41.32 (85.52)	&	40.73 (85.40)	&	41.22 (84.81)	&	38.62 (78.42)	&	19.19 (42.60)	\\
& 100& 	15.75 (50.55)	&	16.64 (50.05)	&	18.02 (51.54)	&	18.29 (51.75)	&	18.07 (53.38)	&	17.68 (51.48)	\\

\bottomrule
    \end{tabular}
    }
\end{table}
In this section we show a sensitivity study on hyperparameters of $\lossdfs$ by varying the scale of $\lambda_1$ and $\lambda_2$ from 0 to 100.0.
The results for SVHN and CIFAR-10 are shown in \cref{tab:supp:lambda_sensitivity_svhn} and \cref{tab:supp:lambda_sensitivity_cifar}, respectively.
In general, setting both $\lambda_1$ and $\lambda_2$ as 1.0 performs well regardless of the model and dataset. Although there exist a few cases where increasing $\lambda_1$ or $\lambda_2$ yields better robust accuracy, they all come at the price of large degradation in clean accuracy. As noted in the main body of the paper, we found setting both values to 1.0 best balances such trade-off, while maintaining high robust accuracy.

\section{Visualization of \mixabb using PCA}
\label{sec:supp:pca}
\begin{figure*}[h]
\captionsetup[subfigure]{justification=centering}
\centering
     \begin{subfigure}{\textwidth}
    \includegraphics[width=\textwidth]{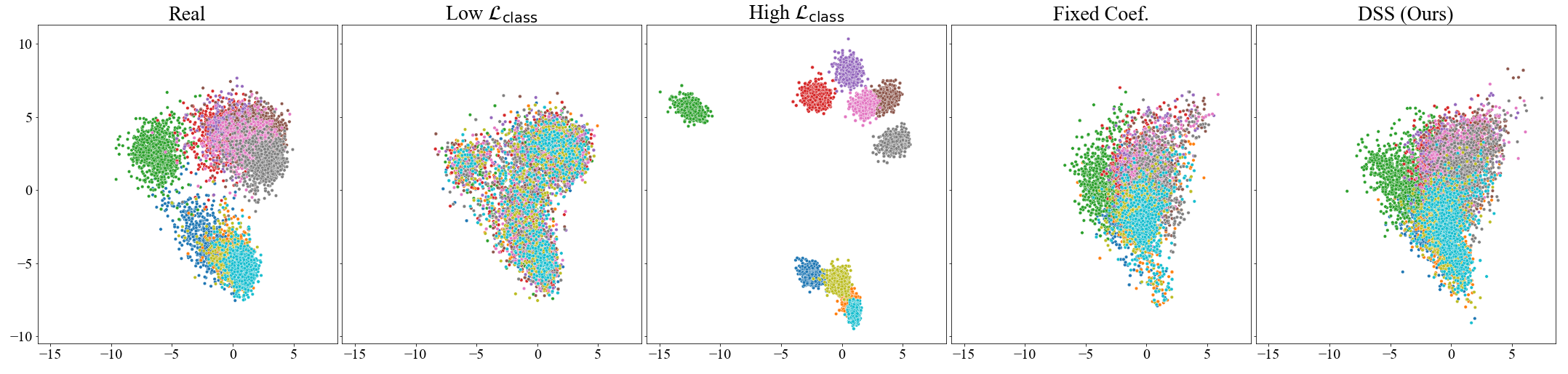}
    \caption{ResNet20 }
    \end{subfigure} 
     \begin{subfigure}{\textwidth}
    \includegraphics[width=\textwidth]{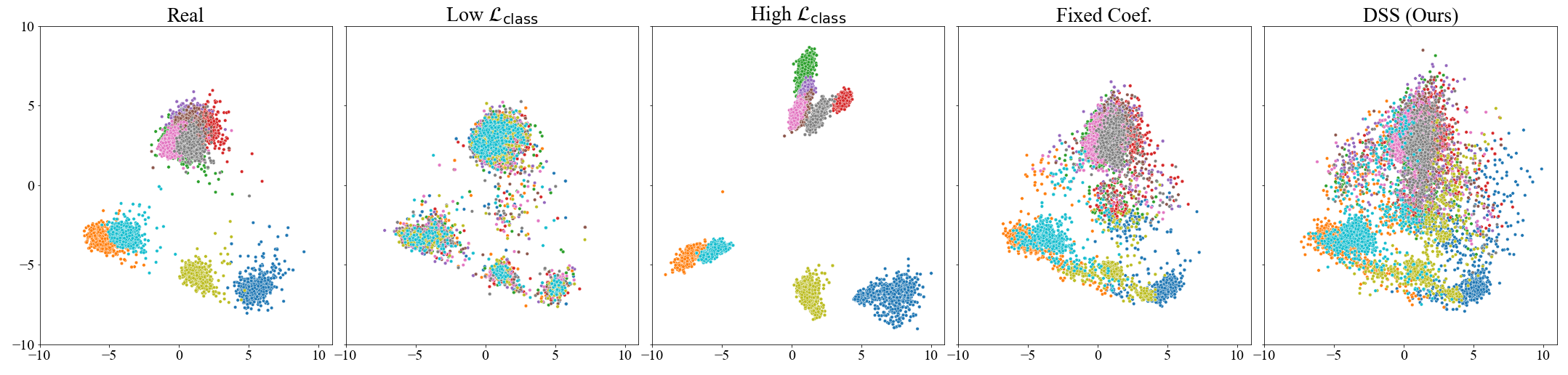}
    \caption{ResNet56}
    \end{subfigure} 
         \begin{subfigure}{\textwidth}
    \includegraphics[width=\textwidth]{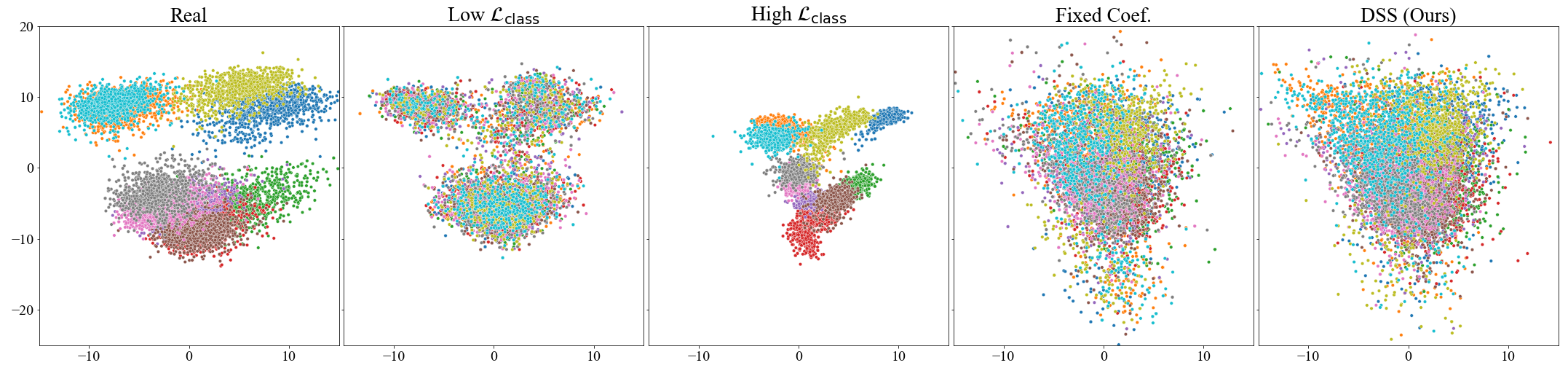}
    \caption{WRN-28-10}
    \end{subfigure} 
    
    \label{fig:supp:pca}
    \caption{PCA visualization using CIFAR-10. We compare \mixabb against 4 different sets of data: Real (CIFAR-10), samples synthesized with low $\mathcal{L}_{class}$, samples synthesized with high $\mathcal{L}_{class}$, and using fixed coefficients.}
\end{figure*}

In this section we provide feature visualization using PCA on CIFAR-10 dataset to conduct similar visualization as done in \cref{fig:toy_example} but on a high-dimensional image dataset.
Using all model architectures, \cref{fig:supp:pca} demonstrates the efficacy of \mixabb in increasing dataset diversity.
\mixabb shows larger coverage in terms of diversity than when using fixed coefficients, aligning with previous results on toy example (\cref{fig:toy_example}).
This shows that \mixabb can effectively enlarge diversity even in complex, real-world datasets.
We also visualize the case of high $\mathcal{L}_{class}$ and low $\mathcal{L}_{class}$ by setting each hyperparameter to 1.0 and 0.0.
While high $\mathcal{L}_{class}$ generates samples with more distinct class information, they are highly clustered to the class centers. 
However, absence of $\mathcal{L}_{class}$ loses class information essential to classification tasks. 
While fixed coefficients strive to find a good balance between these terms with negligence of diversity, \mixabb successfully achieves both.

\section{Generated Synthetic Samples}
\label{sec:supp:samples}
In this section, we display generated synthetic samples used in our experiments, including the baseline methods.
The resulting images are displayed in \cref{fig:supp:sample_tissuemnist_resnet18} to \cref{fig:supp:sample_cifar10_wrn28_10}.
The overall quality of the baseline samples are noticeably poor, with limited diversity and fidelity.
While these images are sufficient for specific tasks such as knowledge distillation or model compression, they are unable to give the necessary amount of information needed in robust training.
On the other hand, \mix is able to generate diverse samples that are also high in fidelity.
For example, in \cref{fig:supp:sample_bloodmnist_resnet18} and \cref{fig:supp:sample_dermamnist_resnet18}, \mix restores colors and shapes of the original data, while also generating non-overlapping, diversified set of examples.
Also, for SVHN, \mix is the only method that is able to generate readable numbers that are recognizable to human eyes.
Even in CIFAR-10, a dataset with more complex features, \mix generates samples that faithfully restore the knowledge learned from the original dataset.
For larger models with more capacity, the generated samples show recognizable objects such as dogs, airplanes, frogs, etc.
The difference in the quality of the generated samples, in addition to the experiment results show that fidelity and diversity of train data play crucial roles in robust training.

\newcommand{\samplewidth}{0.25\textwidth}

\begin{figure}[h]
    \centering
    
    \begin{subfigure}{\samplewidth}     
    \includegraphics[width=\textwidth]{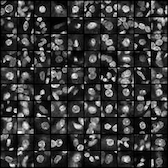}
    \caption{Real}
    \end{subfigure}
    \begin{subfigure}{\samplewidth}
    \includegraphics[width=\textwidth]{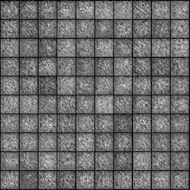}
    \caption{DaST~\citep{dast}}
    \end{subfigure}
    \begin{subfigure}{\samplewidth}
    \includegraphics[width=\textwidth]{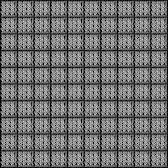}
    \caption{DFME~\citep{dfme}}
    \end{subfigure}
    \begin{subfigure}{\samplewidth}
    \includegraphics[width=\textwidth]{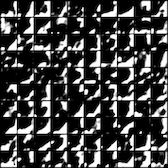}
    \caption{AIT~\citep{ait}}
    \end{subfigure}
    \begin{subfigure}{\samplewidth}
    \includegraphics[width=\textwidth]{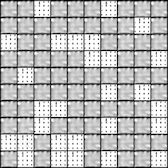}
    \caption{DFARD~\citep{dfard}}
    \end{subfigure}
    \begin{subfigure}{\samplewidth}
    \includegraphics[width=\textwidth]{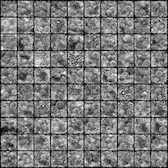}
    \caption{\name (Ours)}
    \end{subfigure}
    \caption{TissueMNIST, ResNet-18}
    \label{fig:supp:sample_tissuemnist_resnet18}

\end{figure}

\begin{figure}[h]
\ContinuedFloat

    \centering

    \begin{subfigure}{\samplewidth}     
    \includegraphics[width=\textwidth]{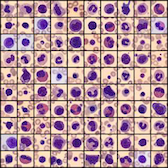}
    \caption{Real}
    \end{subfigure}
   \begin{subfigure}{\samplewidth}
    \includegraphics[width=\textwidth]{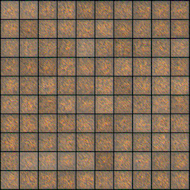}
    \caption{DaST~\citep{dast}}
    \end{subfigure}
   \begin{subfigure}{\samplewidth}
    \includegraphics[width=\textwidth]{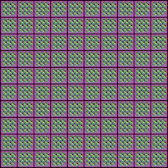}
    \caption{DFME~\citep{dfme}}
    \end{subfigure}
  \begin{subfigure}{\samplewidth}
    \includegraphics[width=\textwidth]{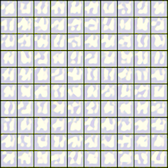}
    \caption{AIT~\citep{ait}}
    \end{subfigure}
   \begin{subfigure}{\samplewidth}
    \includegraphics[width=\textwidth]{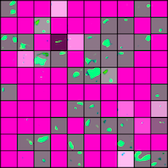}
    \caption{DFARD~\citep{dfard}}
    \end{subfigure}
    \begin{subfigure}{\samplewidth}
    \includegraphics[width=\textwidth]{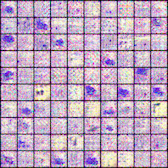}
    \caption{\name (Ours)}
    \end{subfigure}
    \caption{BloodMNIST, ResNet-18}
    \label{fig:supp:sample_bloodmnist_resnet18}

\end{figure}

\begin{figure}[h]
    \centering

    \begin{subfigure}{\samplewidth}     
    \includegraphics[width=\textwidth]{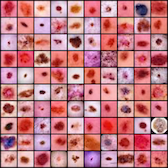}
    \caption{Real}
    \end{subfigure}
    \begin{subfigure}{\samplewidth}
    \includegraphics[width=\textwidth]{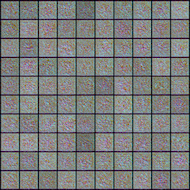}
    \caption{DaST~\citep{dast}}
    \end{subfigure}
    \begin{subfigure}{\samplewidth}     
    \includegraphics[width=\textwidth]{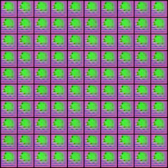}
    \caption{DFME~\citep{dfme}}
    \end{subfigure}
    \begin{subfigure}{\samplewidth}     
    \includegraphics[width=\textwidth]{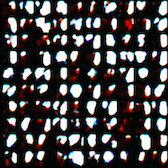}
    \caption{AIT~\citep{ait}}
    \end{subfigure}
    \begin{subfigure}{\samplewidth}
    \includegraphics[width=\textwidth]{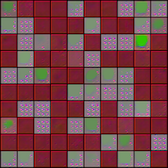}
    \caption{DFARD~\citep{dfard}}
    \end{subfigure}
    \begin{subfigure}{\samplewidth}
    \includegraphics[width=\textwidth]{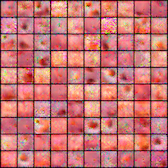}
    \caption{\name (Ours)}
    \end{subfigure}
    \caption{DermaMNIST, ResNet-18}
    \label{fig:supp:sample_dermamnist_resnet18}

\end{figure}

\begin{figure}[h]
    \centering

    \begin{subfigure}{\samplewidth}     
    \includegraphics[width=\textwidth]{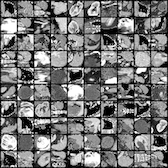}
    \caption{Real}
    \end{subfigure}
    \begin{subfigure}{\samplewidth}
    \includegraphics[width=\textwidth]{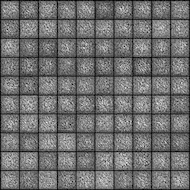}
    \caption{DaST~\citep{dast}}
    \end{subfigure}
       \begin{subfigure}{\samplewidth}     
    \includegraphics[width=\textwidth]{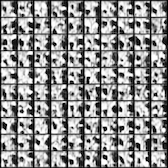}
    \caption{DFME~\citep{dfme}}
    \end{subfigure}
   \begin{subfigure}{\samplewidth}     
    \includegraphics[width=\textwidth]{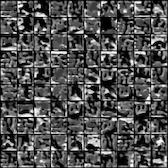}
    \caption{AIT~\citep{ait}}
    \end{subfigure}
    \begin{subfigure}{\samplewidth}
    \includegraphics[width=\textwidth]{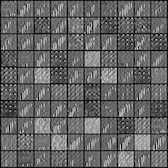}
    \caption{DFARD~\citep{dfard}}
    \end{subfigure}
    \begin{subfigure}{\samplewidth}
    \includegraphics[width=\textwidth]{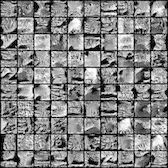}
    \caption{\name (Ours)}
    \end{subfigure}
    \caption{OrganCMNIST, ResNet-18}
    \label{fig:supp:sample_organcmnist_resnet18}

    \label{fig:supp:sample_medmnist}
\end{figure}

\begin{figure}[h]
    \centering

    \begin{subfigure}{\samplewidth}     
    \includegraphics[width=\textwidth]{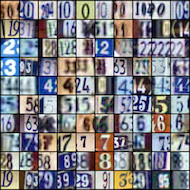}
    \caption{Real}
    \end{subfigure}
    \begin{subfigure}{\samplewidth}
    \includegraphics[width=\textwidth]{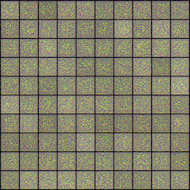}
    \caption{DaST~\citep{dast}}
    \end{subfigure}
       \begin{subfigure}{\samplewidth}     
    \includegraphics[width=\textwidth]{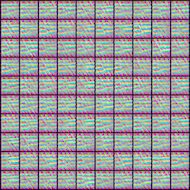}
    \caption{DFME~\citep{dfme}}
    \end{subfigure}
       \begin{subfigure}{\samplewidth}     
    \includegraphics[width=\textwidth]{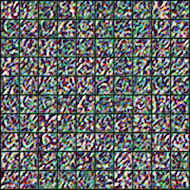}
    \caption{AIT~\citep{ait}}
    \end{subfigure}
    \begin{subfigure}{\samplewidth}
    \includegraphics[width=\textwidth]{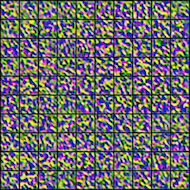}
    \caption{DFARD~\citep{dfard}}
    \end{subfigure}
    \begin{subfigure}{\samplewidth}
    \includegraphics[width=\textwidth]{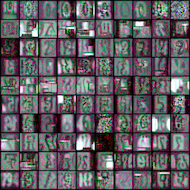}
    \caption{\name (Ours)}
    \end{subfigure}

    \caption{SVHN, WRN-28-10}
    \label{fig:supp:sample_svhn_wrn28_10}

\end{figure}

\begin{figure}[h]
    \centering

    \begin{subfigure}{\samplewidth}     
    \includegraphics[width=\textwidth]{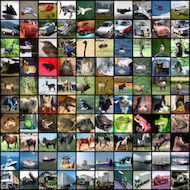}
    \caption{Real}
    \end{subfigure}
   \begin{subfigure}{\samplewidth}
    \includegraphics[width=\textwidth]{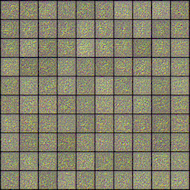}
    \caption{DaST~\citep{dast}}
    \end{subfigure}
       \begin{subfigure}{\samplewidth}     
    \includegraphics[width=\textwidth]{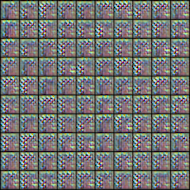}
    \caption{DFME~\citep{dfme}}
    \end{subfigure}
       \begin{subfigure}{\samplewidth}     
    \includegraphics[width=\textwidth]{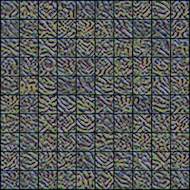}
    \caption{AIT~\citep{ait}}
    \end{subfigure}
    \begin{subfigure}{\samplewidth}
    \includegraphics[width=\textwidth]{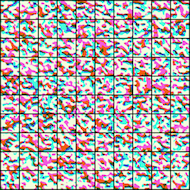}
    \caption{DFARD~\citep{dfard}}
    \end{subfigure}
    \begin{subfigure}{\samplewidth}
    \includegraphics[width=\textwidth]{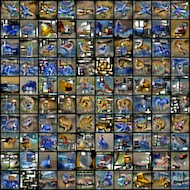}
    \caption{\name (Ours)}
    \end{subfigure}

    \caption{CIFAR-10, WRN-28-10}
    \label{fig:supp:sample_cifar10_wrn28_10}

\end{figure}


\end{document}